\documentclass{article}

\PassOptionsToPackage{numbers,sort&compress}{natbib}

\usepackage[eandd, final]{neurips_2026}
\makeatletter\renewcommand{\@notice}{}\makeatother

\usepackage[utf8]{inputenc} 
\usepackage[T1]{fontenc}    
\usepackage{hyperref}       
\usepackage{url}            
\usepackage{fontawesome5}   
\usepackage{booktabs}       
\usepackage{enumitem}       
\usepackage{amsfonts}       
\usepackage{nicefrac}       
\usepackage{microtype}      
\usepackage[dvipsnames]{xcolor}
\usepackage{tikz}
\usetikzlibrary{arrows.meta}
\usetikzlibrary{shapes.geometric}
\definecolor{refgreen}{HTML}{1F9D72}
\definecolor{candorange}{HTML}{F08A2C}
\DeclareRobustCommand{\refmark}{\tikz[baseline=-0.55ex]\node[fill=refgreen,inner sep=2.2pt]{};}
\DeclareRobustCommand{\candmark}{\tikz[baseline=-0.55ex]\node[star,star points=5,star point ratio=2.3,fill=candorange,inner sep=1.2pt]{};}
\definecolor{datagray}{HTML}{B7BDC8}
\DeclareRobustCommand{\datamark}{\tikz[baseline=-0.55ex]\node[circle,fill=datagray,inner sep=1.3pt]{};}
\definecolor{linkblue}{HTML}{2468C8}  
\usepackage{colortbl}                 
\colorlet{oursbg}{linkblue!10}        
\newcommand{\shadedrules}{\setlength{\aboverulesep}{0pt}\setlength{\belowrulesep}{0pt}}
\newcommand{\Tstrut}{\rule{0pt}{2.5ex}}
\newcommand{\Bstrut}{\rule[-1.2ex]{0pt}{0pt}}
\usepackage{times}

\usepackage{graphicx}

\newcommand{\ourmetric}{Chamfer distance}
\newcommand{\ourmetrictitle}{Chamfer Distance}
\newcommand{\ourmetricshort}{Chamfer}

\usepackage{amsmath}

\DeclareMathOperator*{\E}{\mathbb{E}}

\usepackage{subcaption}
\usepackage{calc}
\usepackage{array}
\usepackage{multirow}

\usepackage{siunitx}
\newlength{\qimgw}
\newlength{\arrowcol}
\newlength{\gridwidth}
\newlength{\qframe}
\newcommand{\framedimg}[2]{{\setlength{\fboxsep}{0pt}\setlength{\fboxrule}{\qframe}\setkeys{Gin}{width=\dimexpr\qimgw-2\qframe\relax,height=\dimexpr\qimgw-2\qframe\relax,keepaspectratio=false}\fcolorbox{#1}{white}{#2}}}
\newcommand{\qimg}[2][refgreen]{\framedimg{#1}{#2}}
\newcommand{\rimg}[2][candorange]{\framedimg{#1}{#2}}

\newlength{\noisew}
\newlength{\imnoisew}
\AtBeginDocument{%
  \setlength{\noisew}{0.157\textwidth}%
  \setlength{\imnoisew}{0.124\textwidth}%
}
\newcommand{\nimg}[1]{{\setkeys{Gin}{width=\noisew,height=\noisew,keepaspectratio=false}#1}}
\newcommand{\imnimg}[1]{{\setkeys{Gin}{width=\imnoisew,height=\imnoisew,keepaspectratio=false}#1}}
\newcommand{\rowlab}[1]{\raisebox{\dimexpr0.5\noisew-0.5\height\relax}[0pt][0pt]{\rotatebox{90}{\small #1}}}

\newcommand{\up}{$\uparrow$}
\newcommand{\dn}{$\downarrow$}

\title{The Fault in Our Metrics: Revisiting Generative Model Evaluation with \ourmetrictitle}

\author{%
  Chirag Vashist$^{1,\dagger}$, \quad
  Vaibhav Santhosh$^{2}$, \quad
  Tristan Engst$^{1}$, \quad
  Yanshu Zhang$^{1}$, \quad
  Ke Li$^{1,3,4}$ \\[2pt]
  $^{1}$Simon Fraser University \qquad
  $^{2}$NIT Karnataka \qquad
  $^{3}$Amii \qquad
  $^{4}$CIFAR \\[6pt]
  {\small\hypersetup{pdfborder={0 0 0}}%
   \href{https://serchirag.github.io/chamfer/}{\color{linkblue}\faGlobe~serchirag.github.io/chamfer}}
}

\begin{document}

\setlength{\gridwidth}{\dimexpr\textwidth - \arrowcol - 6pt\relax}
\setlength{\qimgw}{\dimexpr(\gridwidth - 36pt) / 12\relax}

\maketitle

\begingroup
\renewcommand{\thefootnote}{\fnsymbol{footnote}}
\footnotetext[2]{Corresponding author: \texttt{chirag\_vashist@sfu.ca}}
\endgroup

\begin{abstract}
  Evaluation metrics decide which image generative models are judged better, yet widely used metrics are either uninformative or unreliable. Generative performance should be evaluated across two separate axes: the realism of the generated samples and their coverage of the real distribution. Most popular metrics, like FID and $\text{FD}_{\text{DINOv2}}$, report a single symmetric score that conflates these two axes, which makes diagnosing and comparing models hard. A second family of metrics addresses this limitation by reporting along the two axes separately. We call such metrics disentangled, and they include Precision/Recall and Density/Coverage. However, we show that existing disentangled metrics approximate empirical distributions with oversized volumes, which can cause them to admit off-manifold samples. This causes these metrics to respond unreliably to simple distributional changes and image distortions. We propose Chamfer distance, an intuitive disentangled metric that directly measures cross-set nearest-neighbor distances between real and generated samples. We show that the resulting one-sided distances provide interpretable signals for fidelity and diversity, respond reliably to simple distributional changes, and remain computationally efficient.
\end{abstract}

\section{Introduction}

Image-based generative models have advanced rapidly, producing increasingly realistic and diverse samples \citep{vae,normalizing-flows,gan,dc-gan,ddpm,flow-matching}. As visual quality improves, evaluation metrics increasingly determine which methods are judged better, which trade-offs are visible, and which directions the field prioritizes. 
Because metrics carry this much weight, they need to be both informative enough to diagnose generative models and reliable enough to respond correctly to changes in the generated distribution.

The most widely used metrics, including FID \citep{fid,clean-fid}, $\text{FD}_{\text{DINOv2}}$ \citep{dgm-eval}, $\text{FDr}^{\text{K}}$ \citep{yang2026representationfrechetlossvisual}, KID \citep{kid}, and CMMD \citep{rethinking-fid}, summarize the difference between the real and generated distributions as a \emph{single symmetric score}. These scores are simple and stable to compute, but a single symmetric number has limited diagnostic value: it captures the size of the difference between two distributions, but not its direction. A generative model can be deficient along two distinct axes: \emph{fidelity}, whether its samples look realistic, and \emph{diversity}, whether it covers the full range of the real data. A single symmetric number conflates these two axes, making it hard to meaningfully analyze and compare the performance of different generative models.

A second line of work addresses this limitation by proposing \emph{disentangled} metrics that separately measure performance along the two axes. The most notable are Precision and Recall \citep{improved-pr}, which estimate the support of each empirical distribution as a union of isotropic balls centered on its own samples. The metrics are calculated by checking whether a sample from one distribution falls inside the other's estimated support. We show that these balls can have large volumes, which makes the estimated support extend well beyond the true manifold. Because membership is a binary test, off-manifold points that are far from any real data can still be counted as valid. This has practical consequences, and we demonstrate that such metrics are unreliable even in simple 2D examples. Moreover, we observe that they are insensitive to image degradation and remain stagnant even as images get visibly corrupted. 

In this work, we propose \ourmetric{}: a \emph{disentangled} metric that does not rely on inflated approximations of distributions or binary membership tests. Instead, we report fidelity and diversity separately by directly measuring \emph{cross-set} nearest-neighbor distances between real and generated samples. We show that \ourmetric{} provides sample-level interpretability, while retaining sensitivity to distributional distortions, reducing dependence on neighborhood parameters, and remaining computationally efficient. Our contributions are as follows:
\begin{itemize}
    \item We identify a gap in existing evaluation metrics: they either conflate fidelity/diversity or, when they do separate them, they produce unreliable scores. 
    \item Using a simple two-dimensional example, we show that popular metrics assign perfect scores to two clearly distinct distributions. Through controlled experiments, we show that existing ball-based metrics fail to respond to visible image degradation and do not improve as the candidate set grows denser.
    \item Finally, we propose \ourmetric{}, a disentangled metric that is intuitive, interpretable and computationally efficient. We benchmark modern generative models and show that \ourmetric{} remains informative in cases where existing metrics saturate or disagree with each other.
\end{itemize}

\section{Background}

\begin{figure}[t]
    \centering
    \begin{subfigure}[t]{0.30\textwidth}
        \centering
        \includegraphics[width=\linewidth]{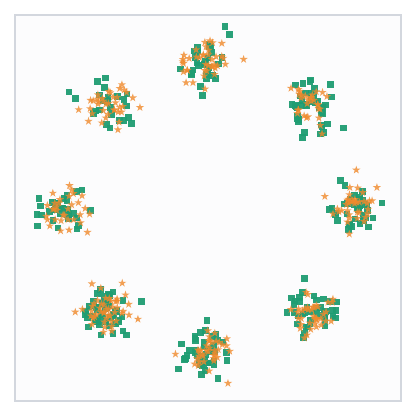}
        \caption{High fidelity and diversity}
        \label{fig:modela}
    \end{subfigure}%
    \hspace{0.02\textwidth}%
    \begin{subfigure}[t]{0.30\textwidth}
        \centering
        \includegraphics[width=\linewidth]{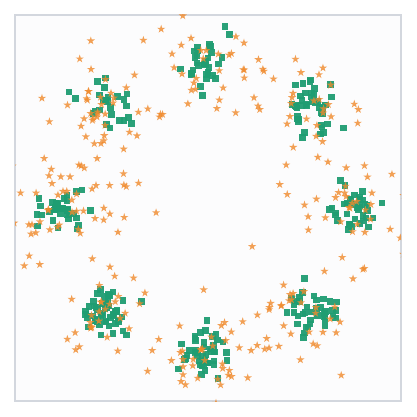}
        \caption{Low fidelity, high diversity}
        \label{fig:modelb}
    \end{subfigure}%
    \hspace{0.02\textwidth}%
    \begin{subfigure}[t]{0.30\textwidth}
        \centering
        \includegraphics[width=\linewidth]{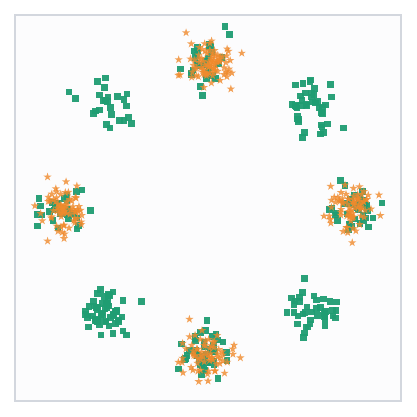}
        \caption{Low diversity, high fidelity}
        \label{fig:modelc}
    \end{subfigure}
    \caption{\textbf{A single symmetric score does not identify which axis a model failed on.} Green squares (\refmark) are the real distribution the models are trained on. Orange stars (\candmark) are samples from three different models with varying generative performance.}
    \label{fig:mmd-ring}
\end{figure}

In this section, we briefly review existing evaluation metrics. The metrics we discuss in this work compare a \emph{reference set} of real images against a \emph{candidate set}, which in practice contains samples from the model being evaluated. The two distributions are often compared in the embedding space of a pretrained encoder \citep{inception, clip, dinov2}.

\textbf{Distributional Metrics.} These metrics report the discrepancy between two distributions as a single symmetric score. Fréchet Inception Distance (FID), arguably the most widely used evaluation metric, is an example of this family. FID embeds real and candidate/generated images into the feature space of a pretrained Inception-v3 network \citep{inception}, fits a multivariate Gaussian to each set of embeddings, and computes the Fréchet distance between the two fitted Gaussian distributions. Despite its popularity, several works have exposed structural flaws of FID \citep{rethinking-fid,density-and-coverage,wacv-gaussian-mixtures}, and we discuss them in Appendix~\ref{appendix:fid-limit}. Other works compute the Fréchet distance with different encoders \citep{dgm-eval,yang2026representationfrechetlossvisual}, in part because Inception features align poorly with human judgments of image quality \citep{dgm-eval}. A different line of work, such as KID and CMMD, avoids the Gaussian assumption of FID by comparing the two sets through Maximum Mean Discrepancy (MMD) \citep{mmd-test}. 

Regardless of their formulation, all distributional metrics are symmetric: for two distributions $P$ and $Q$, they satisfy $d(P,Q) = d(Q,P)$. Hence, the reported score can indicate how large the discrepancy is, but not where it comes from. Consider Figure~\ref{fig:mmd-ring}, which shows candidate samples from three models evaluated against the same reference distribution. Model A (Figure~\ref{fig:modela}) is well fitted: its samples lie close to real data (high fidelity) and cover it well (high diversity). Model B (Figure~\ref{fig:modelb}) covers every real sample, but many of its own samples lie far from any real data (low fidelity, high diversity). Model C (Figure~\ref{fig:modelc}) places all of its samples close to real data, but leaves many real samples with no generated sample nearby (high fidelity, low diversity). Any reasonable evaluation metric would rank Model A above both Model B and Model C. A symmetric score would rank Model B and Model C below Model A. However, from Model B's score alone, one cannot tell whether it lacks fidelity or diversity, and the same holds for Model C. Hence, a symmetric score conflates fidelity and diversity, making it hard to analyze and compare different models.

\begin{figure}[t]
    \centering
    \newcommand{\dclabel}[1]{\raisebox{-.5\height}{\rotatebox{90}{\footnotesize #1}}}%
    \newlength{\dclabw}\settowidth{\dclabw}{\dclabel{jack-o'-lantern}}%
    \newlength{\dcgridw}\setlength{\dcgridw}{\dimexpr(\textwidth-\dclabw-10pt)/2\relax}%
    \newcommand{\dcgrid}[1]{\raisebox{-.5\height}{\setkeys{Gin}{width=\dcgridw}#1}}%
    \setlength{\tabcolsep}{0pt}%
    \begin{tabular}{@{}c@{\hspace{4pt}}c@{\hspace{6pt}}c@{}}
        & \small\textbf{BigGAN}: FID $6.67$ & \small\textbf{IMM (1 step)}: FID $8.11$ \\ \noalign{\vskip 3pt}
        \dclabel{jack-o'-lantern} & \dcgrid{\includegraphics{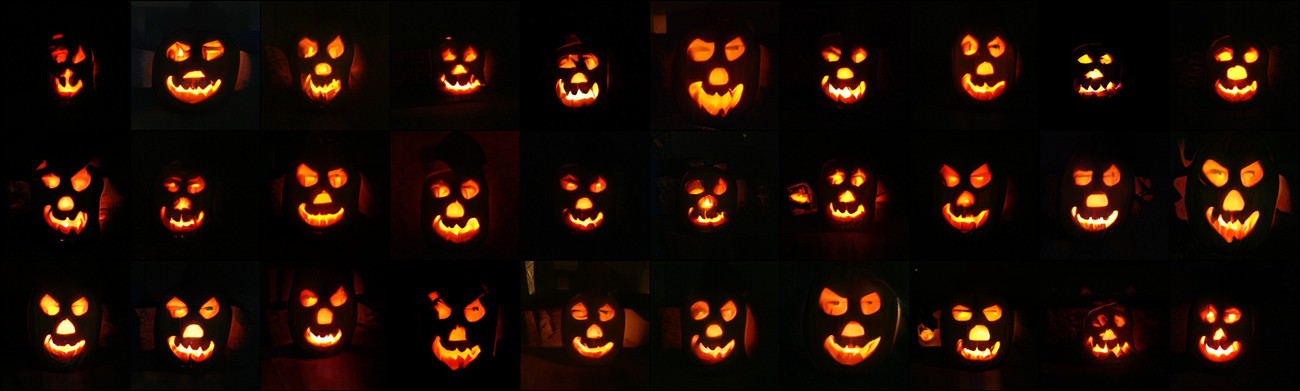}} & \dcgrid{\includegraphics{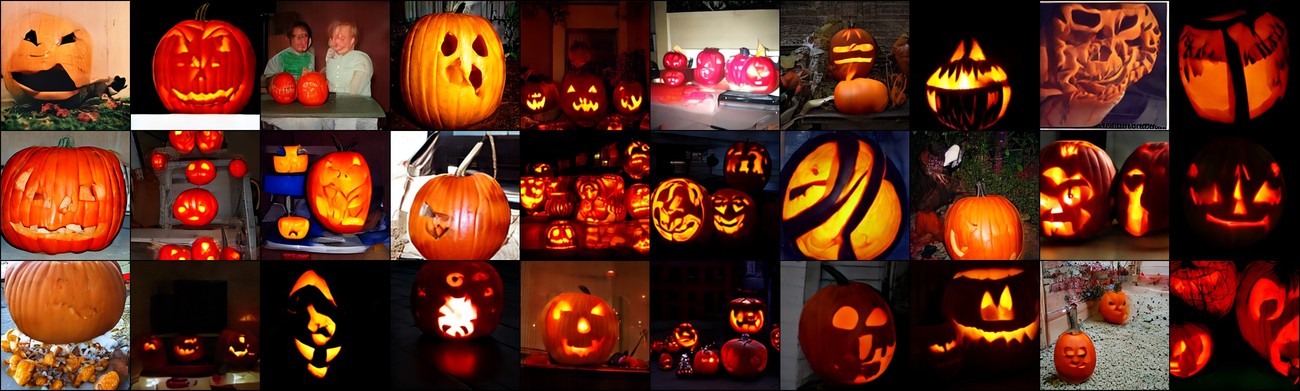}} \\ \noalign{\vskip 3pt}
        \dclabel{volcano}         & \dcgrid{\includegraphics{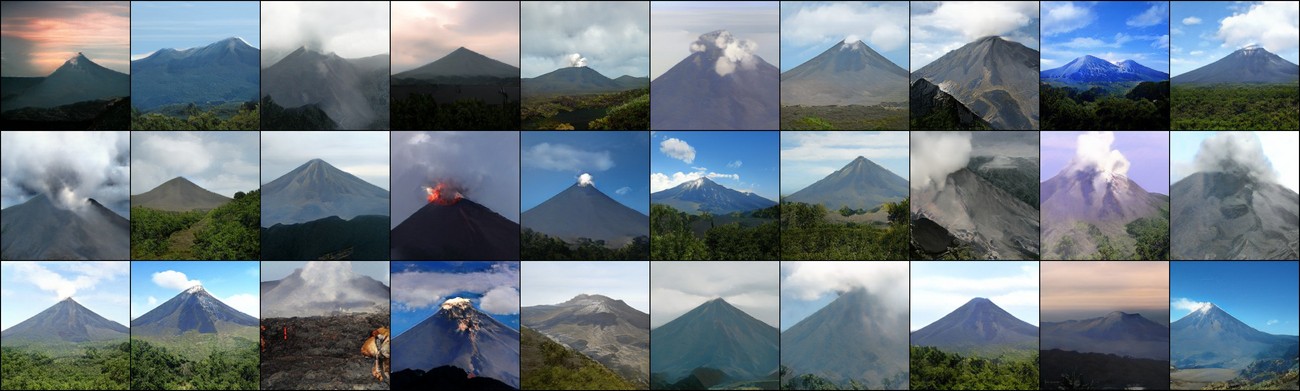}} & \dcgrid{\includegraphics{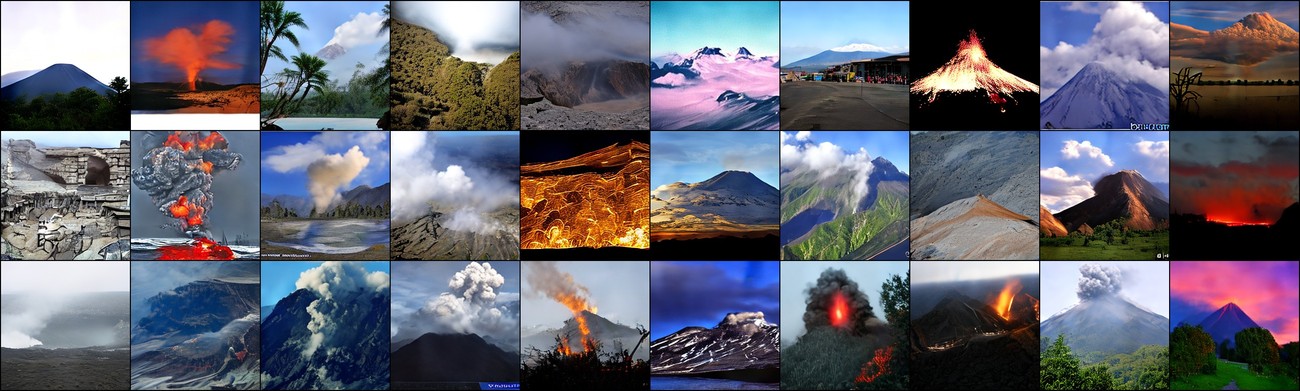}} \\ \noalign{\vskip 3pt}
        \dclabel{daisy}           & \dcgrid{\includegraphics{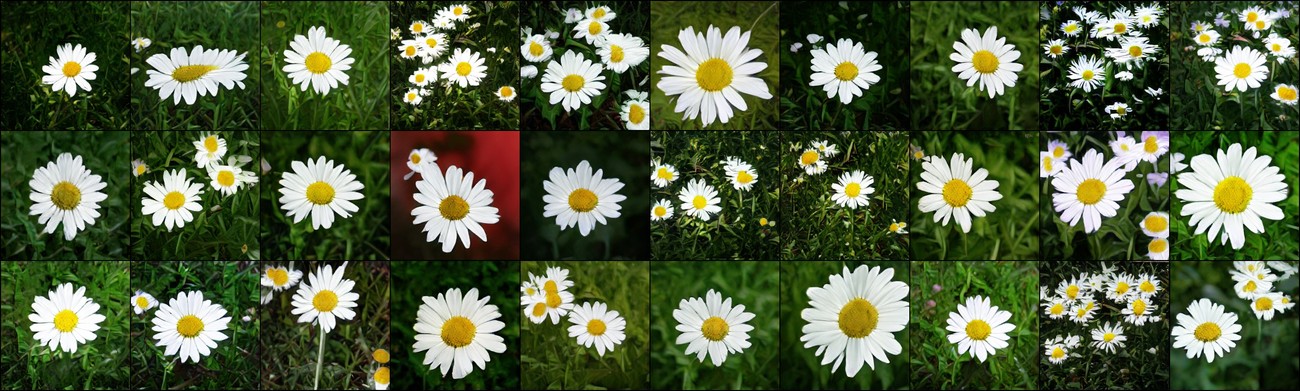}} & \dcgrid{\includegraphics{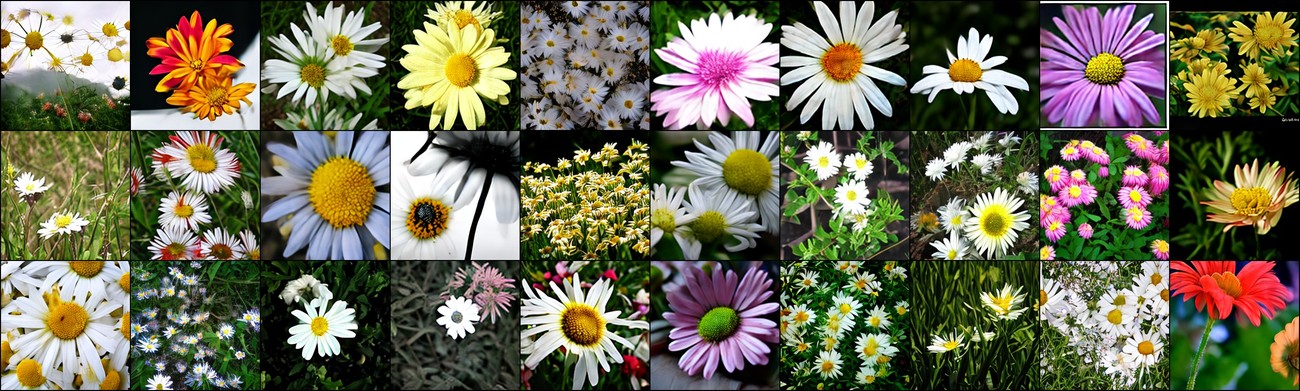}} \\
    \end{tabular}
    \caption{\textbf{Random samples for three ImageNet classes}: Samples from IMM look more visually diverse, yet FID ranks BigGAN \citep{brock2018large} above IMM \citep{zhou2025inductive}.}
    \label{fig:dist-confusion}
\end{figure}

This can be demonstrated by a practical example too. Consider Figure~\ref{fig:dist-confusion} and Figure~\ref{fig:dist-confusion-extra} which shows random samples from BigGAN and IMM (1 step) for different ImageNet classes. Looking at the samples, one can see that that IMM has more visually diverse samples. Yet, a distributional metric like FID just reports a single number, which does not capture this obvious visual difference. 

\textbf{Disentangled Metrics.}
In order to address this drawback, \emph{disentangled metrics} have been proposed that measure generative models along two separate axes. Most existing disentangled metrics follow the \emph{KNN-ball framework}, first proposed by \citet{improved-pr}. The support of each empirical distribution is approximated by placing a ball around every sample, with radii set by local nearest-neighbor distances. The union of all such balls gives the estimated support for that empirical distribution. These estimated supports make it possible to answer binary membership queries of the form: ``Does a sample $x$ lie in the support of distribution $P$?'' Concretely, let $P$ and $Q$ denote the real and candidate distributions, respectively, with samples $\{x_i\}_{i=1}^{N} \sim P$ and $\{y_j\}_{j=1}^{M} \sim Q$. The support of $P$ is approximated by the union of balls centered at each real sample:
$\widehat{\mathcal{M}}_X  := \bigcup_{i=1}^{N} B\!\left(f(x_i),\, \mathrm{NND}_k\left(f(x_i)\right)\right)$,
where $B(c, r)$ is the ball of radius $r$ around $c$ in feature space, $f(\cdot)$ is a feature extractor, and $\mathrm{NND}_k\left(f(x_i)\right)$ is the distance from $f(x_i)$ to its $k$-th nearest neighbor in $\{f(x_j)\}_{j=1,\, j\neq i}^{N}$. The support of $Q$ is defined analogously. Precision and Recall are then:
\[
\text{Precision} :=
\frac{1}{M} \sum_{j=1}^{M}
\mathbf{1}\!\left[f(y_j) \in \widehat{\mathcal{M}}_X\right],
\qquad
\text{Recall} :=
\frac{1}{N} \sum_{i=1}^{N}
\mathbf{1}\!\left[f(x_i) \in \widehat{\mathcal{M}}_Y\right].
\]
Several works refine this KNN-ball framework by modifying the scoring rule, including Density/Coverage \citep{density-and-coverage} and Symmetric Precision/Recall \citep{assym-pr}. We cover them briefly in Section~\ref{sec:ball-limitations}. Despite these refinements, the underlying principle of all these approaches remains the same: each method uses \emph{within-set} nearest-neighbor distances to construct local support estimates and evaluates samples by their membership in those estimated supports.

\section{Limitations of KNN-ball-based metrics}
\label{sec:ball-limitations}

\begin{figure}[h]
    \centering
    \begin{minipage}[t]{0.66\textwidth}
        \vspace{0pt}
        \centering
        \begin{subfigure}[t]{0.32\linewidth}
            \centering
            \includegraphics[width=\linewidth]{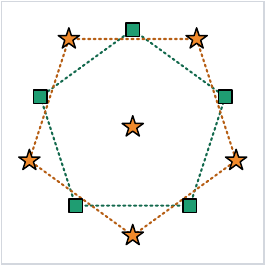}
            \caption{The two pentagons.}
            \label{fig:pentagon-sets}
        \end{subfigure}
        \hfill
        \begin{subfigure}[t]{0.32\linewidth}
            \centering
            \includegraphics[width=\linewidth]{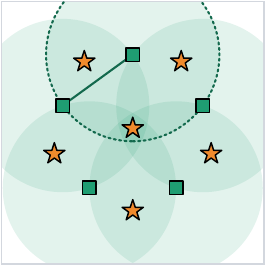}
            \caption{Reference support.}
            \label{fig:pentagon-real}
        \end{subfigure}
        \hfill
        \begin{subfigure}[t]{0.32\linewidth}
            \centering
            \includegraphics[width=\linewidth]{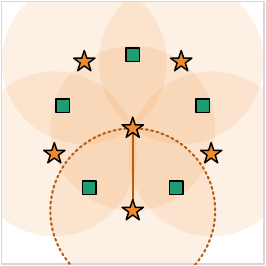}
            \caption{Candidate support.}
            \label{fig:pentagon-cand}
        \end{subfigure}
        \caption{\textbf{A counterexample for KNN-ball metrics.} Two pentagons: a reference set of green squares (\refmark) and a candidate set of orange stars (\candmark).}
        \label{fig:pentagon}
    \end{minipage}
    \hfill
    \begin{minipage}[t]{0.31\textwidth}
        \vspace{0pt}
        \centering
        \captionsetup{position=top}
        \captionof{table}{Metrics for the two pentagons ($k=1$).}
        \label{tab:pentagon}
        \footnotesize
        \renewcommand{\arraystretch}{1.1}
        \begin{tabular}{l S}
            \toprule
            Metric & {Value} \\
            \midrule
            Precision\up       & 1.00 \\
            Recall\up          & 1.00 \\
            \midrule
            Density\up         & 2.50 \\
            Coverage\up        & 1.00 \\
            \midrule
            Sym.\ Precision\up & 1.00 \\
            Sym.\ Recall\up    & 1.00 \\
            \midrule
            FD\dn              & 0.00 \\
            \bottomrule
        \end{tabular}
    \end{minipage}
\end{figure}

While the motivation behind disentangled metrics is valid, in this paper we argue that \emph{existing} disentangled metrics are unreliable, even in simple cases. These metrics estimate the manifold of each distribution with nearest-neighbor search: they place an isotropic ball around every sample, with its radius set by the distance to the $k$-th nearest neighbor within the same set.
In practice, isotropic balls cover not only regions along the true manifold, but also large off-manifold regions far from any real data.
Since these metrics rely only on binary membership in this estimate, samples that are far from the true manifold can still be counted as
belonging to it, as long as they fall inside some ball. 

Consider a simple conceptual example that illustrates this. Figure~\ref{fig:pentagon-sets} shows two distributions: one of green squares (\refmark) and another of orange stars (\candmark).
Clearly, the two distributions are distinct, and any bona fide metric should be able to tell them apart. Yet, as shown in Table~\ref{tab:pentagon}, existing ball-based metrics obtain perfect scores and wrongly conclude that the distributions are identical.

\textbf{Precision and Recall.} Precision measures the fraction of candidate samples that lie
within the estimated real support, while Recall measures the fraction of real samples that lie
within the estimated candidate support. As Figures~\ref{fig:pentagon-real}
and~\ref{fig:pentagon-cand} show, both metrics achieve a perfect score of 1.

\textbf{Density and Coverage} \citep{density-and-coverage}. Density counts how many real balls contain each candidate, while Coverage measures the fraction of real balls that contain at least one candidate. As Figure~\ref{fig:pentagon-real} shows, Coverage achieves a perfect score of 1, and
Density is even higher at 2.5.

\textbf{Symmetric Precision and Recall.} \citet{assym-pr} introduce complements of
Precision and Recall. For example, cPrecision measures the fraction of candidate samples whose ball contains at least one real sample. Symmetric Precision is defined by taking the minimum of Precision and cPrecision. Similarly, Symmetric Recall is the minimum of Recall and Coverage. In the case of Figure~\ref{fig:pentagon-sets}, both symmetric metrics achieve a perfect score of 1.

\textbf{Fréchet Distance (FD) under a Gaussian Assumption.} Even though it is not a disentangled ball-based metric, it is worthwhile to note that FD, which underlies popular metrics like FID, $\text{FD}_{\text{DINOv2}}$ \citep{dgm-eval} and $\text{FDr}^{\text{K}}$ \citep{yang2026representationfrechetlossvisual}, also fails here. Both sets have the same mean and covariance, so their fitted Gaussians coincide, leading to a perfect FD score of 0.

This example illustrates how existing metrics can be unreliable even in a simple 2D setting. In the following subsections, we explore how the same mechanism affects ball-based metrics on real image data using simple experiments.



\subsection{Insensitivity to Image Distortions}
\label{sec:limit-img}

\begin{figure}[h]
    \centering
    \begin{subfigure}[b]{0.47\textwidth}
        \centering
        \includegraphics[width=\linewidth]{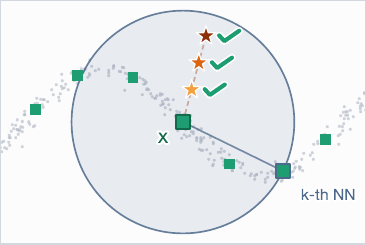}
        \caption{The inflated estimated support phenomenon.}
        \label{fig:ball-insensitivity}
    \end{subfigure}
    \hfill
    \begin{subfigure}[b]{0.51\textwidth}
        \centering
        \setlength{\tabcolsep}{1pt}
        \renewcommand{\arraystretch}{0}
        \begin{tabular}{@{}cccc@{}}
            \rowlab{Original}  & \nimg{\includegraphics{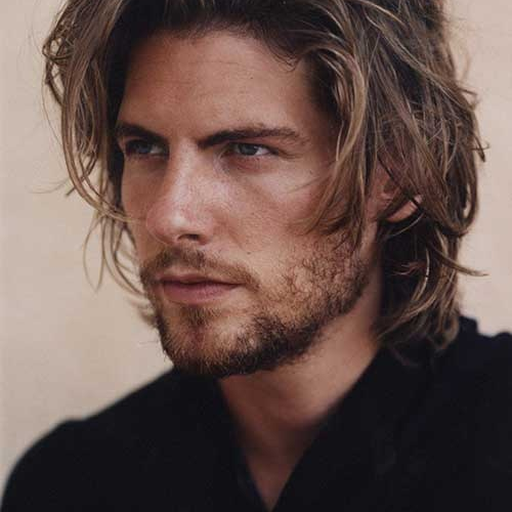}}       & \nimg{\includegraphics{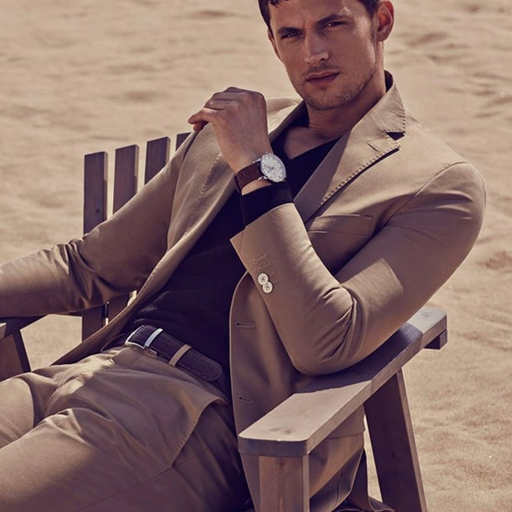}}       & \nimg{\includegraphics{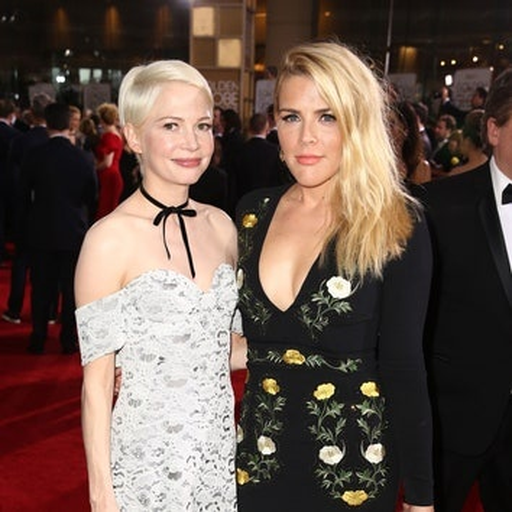}}       \\
            \rowlab{Distorted} & \nimg{\includegraphics{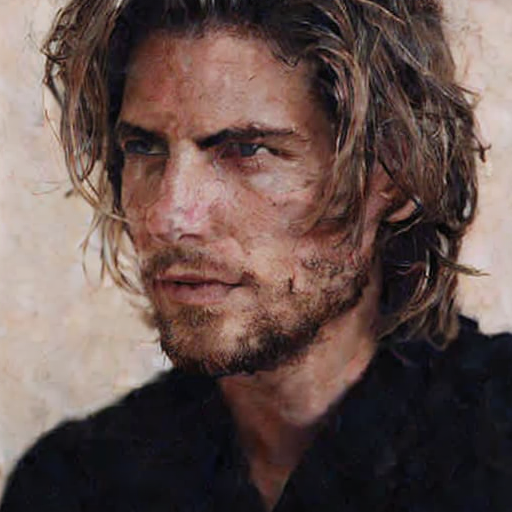}} & \nimg{\includegraphics{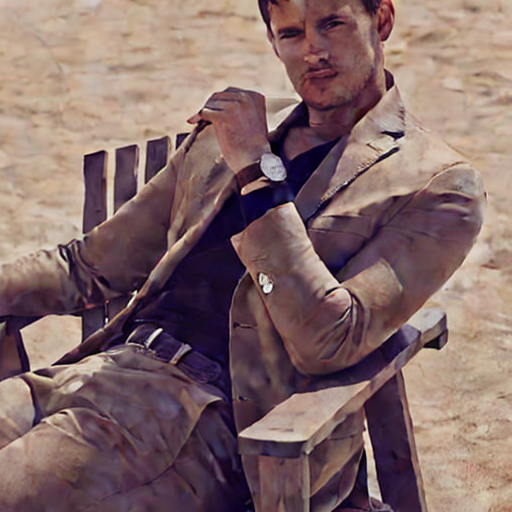}} & \nimg{\includegraphics{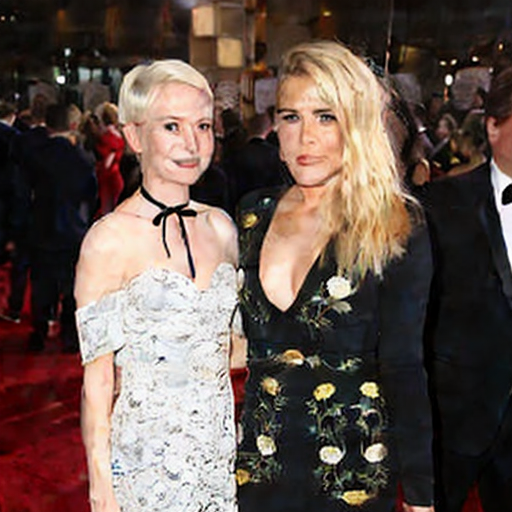}}
        \end{tabular}
        \caption{Real images and their distortions.}
        \label{fig:noise-grid}
    \end{subfigure}
    \caption{\textbf{The binary membership test and insensitivity to distortions.} \textbf{(a)} KNN-ball metrics estimate the support around each point \(x\) with a ball whose radius is equal to the \(k\)-th nearest neighbor distance. Gray dots (\datamark) show the data manifold and green squares (\refmark) the reference samples. Orange stars (\candmark) are increasingly distorted versions of \(x\), obtained by adding an increasing degree of noise. These distorted points can lie far from the manifold, but as long as they lie within the ball, they are counted as valid. \textbf{(b)} The original and distorted images from the LAION-5B dataset. Here, the distorted images correspond to a noise level of $\sigma = 0.225$.}
    \label{fig:ball-and-noise}
\end{figure}

In general, any useful evaluation metric should respond when samples are degraded by perceptible distortions. We test this property using 5K randomly sampled LAION-5B images \citep{laion5b}. For each image, we create a paired distorted sample by first encoding the image with a pretrained Stable Diffusion VAE \citep{stable-diffusion}. Then, Gaussian noise with mean zero and standard deviation $\sigma$ is added to the encoded latents. Varying $\sigma$ controls the degree of corruption. Finally, the corrupted encodings are decoded back to pixel space. For this experiment the metrics are evaluated on features extracted using DINOv2 (ViT-g/14) \citep{dinov2} with neighborhood size fixed at $k=3$. Clean images are used as the reference set, while distorted images form the candidate set. Figure~\ref{fig:noise-grid} shows representative original-distorted pairs at one noise level, illustrating the \emph{perceptual degradation} induced by the corruption process. 

As shown in Table~\ref{tab:noise-dino}, KNN-ball-based metrics remain nearly unchanged as the noise level increases. This happens because ball-based metrics only test for binary membership inside inflated supports. Figure~\ref{fig:ball-insensitivity} illustrates this in two dimensions: as long as the distorted images remain within the ball, the associated metrics do not register any corruption to the images. Additional results for the same experiment on the ImageNet dataset can be found in Appendix~\ref{app:distortion-imagenet}.

\begin{table}[h]
\centering
\small
\caption{\textbf{Sensitivity to image distortions.} Metrics are computed using DINOv2 (ViT-g/14) \citep{dinov2} features for 5K LAION images and their distorted counterparts across increasing latent noise levels. Rows correspond to the standard deviation \(\sigma\) of the Gaussian noise added.}
\label{tab:noise-dino}
\shadedrules
\begin{tabular}{l S S S S S S >{\columncolor{oursbg}}S >{\columncolor{oursbg}}S}
\toprule
\Tstrut\Bstrut & \multicolumn{2}{c}{\textbf{PR}~\citep{improved-pr}} & \multicolumn{2}{c}{\textbf{DC}~\citep{density-and-coverage}} & \multicolumn{2}{c}{\textbf{Symmetric}~\citep{assym-pr}} & \multicolumn{2}{>{\columncolor{oursbg}}c}{\textbf{Chamfer} \textit{(Ours)}} \\
\cmidrule(lr){2-3} \cmidrule(lr){4-5} \cmidrule(lr){6-7} \cmidrule(lr){8-9}
\(\sigma\)\Tstrut\Bstrut & {Prec.\up} & {Recall\up} & {Density\up} & {Coverage\up} & {Prec.\up} & {Recall\up} & {Fidelity\dn} & {Coverage\dn} \\
\midrule
0.10\Tstrut & 0.998 & 0.998 & 1.211 & 0.996 & 0.996 & 0.996 & 31.375 & 31.397 \\
0.125  & 0.998 & 0.996 & 1.217 & 0.996 & 0.996 & 0.996 & 31.939 & 31.977 \\
0.15   & 0.998 & 0.998 & 1.220 & 0.995 & 0.996 & 0.995 & 32.596 & 32.645 \\
0.175  & 0.997 & 0.998 & 1.209 & 0.994 & 0.996 & 0.994 & 33.354 & 33.403 \\
0.20   & 0.996 & 0.997 & 1.206 & 0.994 & 0.995 & 0.994 & 34.107 & 34.169 \\
0.225\Bstrut & 0.996 & 0.996 & 1.183 & 0.993 & 0.994 & 0.993 & 34.959 & 35.026 \\
\bottomrule
\end{tabular}
\end{table}

\subsection{Insensitivity to Sampling Density}
\label{sec:limit-density}

\begin{figure}[t]
    \centering
    \begin{tikzpicture}
        \draw[line width=1.1pt, -{Triangle[length=7pt,width=6pt]}] (0,0) -- (0.62\textwidth,0);
        \node[above, inner sep=3pt] at (0.31\textwidth,0) {\small Sampling density increases};
    \end{tikzpicture}\\[3pt]
    \begin{subfigure}[t]{0.32\textwidth}
        \centering
        \includegraphics[width=\linewidth]{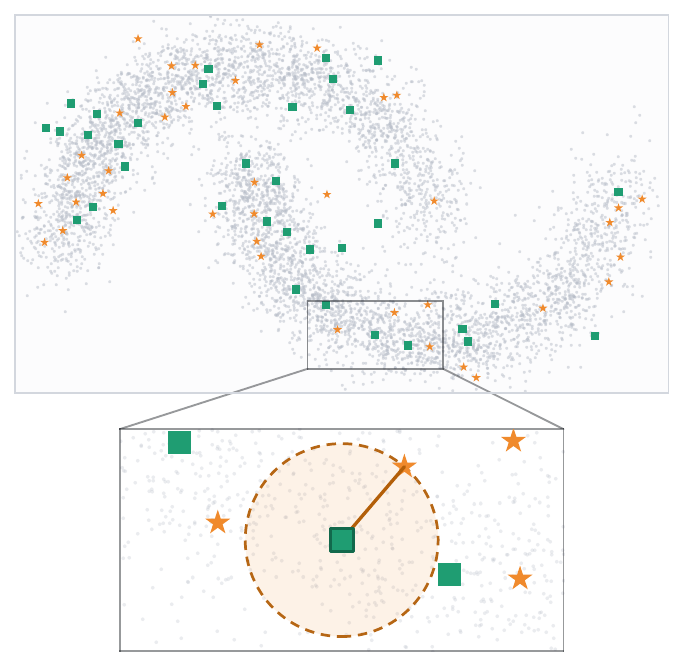}
        \caption{40 candidate samples.}
        \label{fig:sampling-density-sparse}
    \end{subfigure}
    \hfill
    \begin{subfigure}[t]{0.32\textwidth}
        \centering
        \includegraphics[width=\linewidth]{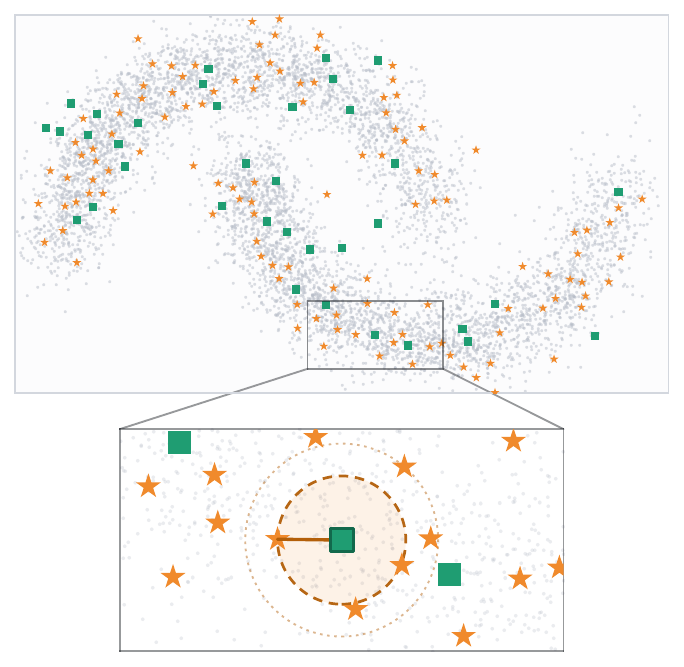}
        \caption{120 candidate samples.}
        \label{fig:sampling-density-mid}
    \end{subfigure}
    \hfill
    \begin{subfigure}[t]{0.32\textwidth}
        \centering
        \includegraphics[width=\linewidth]{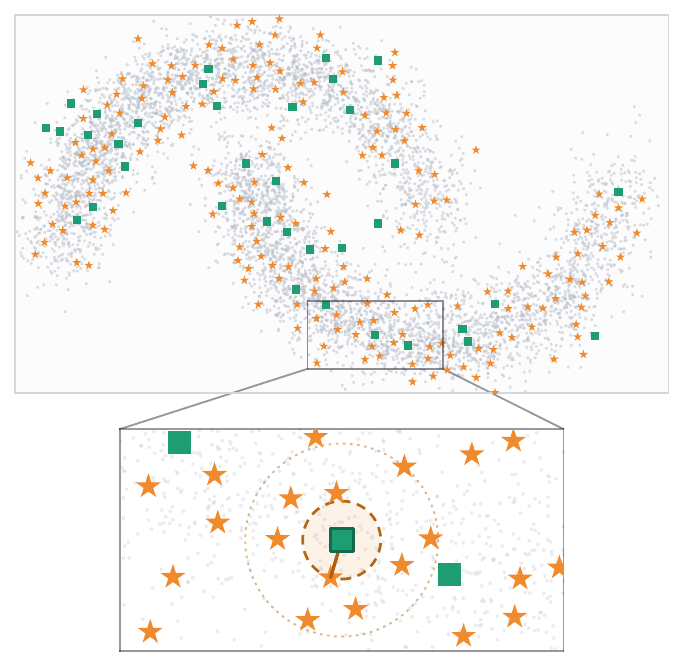}
        \caption{200 candidate samples.}
        \label{fig:sampling-density-dense}
    \end{subfigure}
    \caption{\textbf{Sensitivity to sampling density.} A denser candidate set should be recognized as covering the reference set better. Gray dots (\datamark) show the underlying two-dimensional distribution, green squares (\refmark) a fixed reference set, and orange stars (\candmark) the candidate set. Both sets are drawn from the same distribution.}
    \label{fig:sampling-density}
\end{figure}

Another limitation of KNN-ball-based metrics is that once these balls are large enough, the associated metrics can saturate even when the candidate set becomes substantially denser. We consider an \emph{oracle setting}, where the candidate/generated set consists of real images rather than samples from a trained generative model. Equivalently, this can be viewed as sampling from a perfect oracle generator that only produces real images. This removes image quality as a confounding factor from our analysis, since every candidate sample is an actual real image. For the following analysis, the candidate set is formed directly from real ImageNet \citep{imagenet-1k} training images, with randomly sampled subsets of sizes 10K, 30K, and 50K.

\textbf{Expected behavior.} Before turning to the results, it is useful to reason through what a diversity metric is intended to measure and how it should behave. Diversity asks whether the candidate set represents the different regions of the real data distribution. In the oracle setting, increasing the size of the candidate set gives a denser empirical sample from that same real distribution. Therefore, for any fixed reference image, as the candidate set size increases, it becomes more likely that some candidate image lies nearby. Figure~\ref{fig:sampling-density} illustrates it in two dimensions: as sampling density increases, each reference point (\refmark) is better covered by a candidate point (\candmark). Aggregated over the reference set, a well-calibrated diversity metric should therefore indicate increased coverage of the reference distribution as the candidate set size increases and hence should get better. We call this desideratum \emph{sensitivity to sampling density}. 

\begin{table}[h]
\centering
\caption{\textbf{Oracle real-subset evaluation on ImageNet-1K.} The standard Precision/Recall evaluation protocol uses a 10K-image real subset as the reference set \citep{dhariwal2021diffusionmodelsbeatgans}. We evaluate real ImageNet training subsets of size 10K, 30K, and 50K as the candidate/generated set. FID is computed against statistics estimated from the full ImageNet training set. The neighborhood hyperparameter is set to $k = 1$ and features are extracted using Inception-v3 \citep{inception}.
}
\label{tab:metrics}
\shadedrules
\begin{tabular}{l S[table-format=1.2] S[table-format=3.2] S[table-format=1.2] S[table-format=1.2] S[table-format=1.2] S[table-format=1.2] >{\columncolor{oursbg}}S[table-format=2.2] >{\columncolor{oursbg}}S[table-format=2.2]}
\toprule
\Tstrut\Bstrut & {\textbf{FID}\dn} & {\textbf{IS}\up} & \multicolumn{2}{c}{\textbf{PR}} & \multicolumn{2}{c}{\textbf{Symmetric}} & \multicolumn{2}{>{\columncolor{oursbg}}c}{\textbf{Chamfer} \textit{(Ours)}} \\
\cmidrule(lr){4-5} \cmidrule(lr){6-7} \cmidrule(lr){8-9}
\textbf{Gen.\ size}\Tstrut\Bstrut & & & {Prec.\up} & {Recall\up} & {Prec.\up} & {Recall\up} & {Fid.\dn} & {Cov.\dn} \\
\midrule
10K\Tstrut & 4.1276 & 330.90 & 0.5098 & 0.5153 & 0.5003 & 0.5036 & 10.5917 & 10.5742 \\
30K & 1.3720 & 331.75 & 0.5106 & 0.4937 & 0.2530 & 0.4937 & 10.6022 &  9.8980 \\
50K\Bstrut & 0.8478 & 331.71 & 0.5113 & 0.4960 & 0.1708 & 0.4960 & 10.6055 &  9.5553 \\
\bottomrule
\end{tabular}
\end{table}

\textbf{Results.} Table~\ref{tab:metrics} shows that KNN-ball diversity metrics do not respond correctly to changes in sampling density. Recall is essentially flat, and even decreases from \(0.52\) at 10K to \(0.50\) at 50K, although the larger candidate set is a denser sample from the same real distribution. Notably, the same trend holds for Symmetric Recall \citep{assym-pr}, which was proposed to mitigate the limitations of previous methods. This behavior is consistent with the inflated-ball phenomenon. At 10K candidates, the estimated KNN-ball support already covers a large region, so many reference samples are counted as covered by the binary membership test. Adding more real candidate samples then does not improve the score, even though coverage should improve in the oracle setting. Instead of tracking the sampling density of the candidate set, these metrics primarily reflect membership in an inflated support estimate.

\textbf{Note on performance of FID:} Since FID is the most widely used evaluation metric, it is worthwhile to take a short detour and observe how FID behaves under the oracle setting. For FID, each candidate set (10K, 30K, and 50K images) is compared against reference statistics estimated from the full ImageNet training set of 1.2 million images. Even the 10K subset obtains an FID of \(4.13\), despite containing less than \(1\%\) of the ImageNet training set. Increasing the candidate set size further drives FID down to \(1.37\) at 30K and \(0.85\) at 50K. These scores would normally be interpreted as very strong generative performance \citep{zhou2025inductive, stylegan-xl}. This result is significant because it shows that \emph{a low FID only requires matching the mean and covariance of the reference features, which a small subset of real images already does}. A method can therefore achieve a low FID while still having poor distributional coverage, especially if it memorizes, retrieves, or over-represents common modes. In this sense, the experiment demonstrates that low FID is not sufficient evidence that a model has learned the full data distribution.

Additionally, Appendix~\ref{appendix:dist-shift} reports how the above metrics behave under further distributional changes. 

\section{Proposed Metric}

\begin{figure}[h]
  \centering
  \setlength{\tabcolsep}{0pt}
  \renewcommand{\arraystretch}{0.6}
  \begin{tabular}{@{}c@{\hspace{4pt}}c@{}}
    \smash{\makebox[\arrowcol][c]{\rotatebox[origin=c]{90}{%
      \makebox[\gridwidth][c]{$\longleftarrow$\;\textbf{Increasing distance}\;$\longleftarrow$}%
    }}} &
    \begin{tabular}{@{}c@{\hspace{4pt}}ccc@{\hspace{12pt}}c@{\hspace{4pt}}ccc@{\hspace{12pt}}c@{\hspace{4pt}}ccc@{}}
      \tiny\textbf{\textcolor{candorange}{Fake}} & \multicolumn{3}{@{}c@{}}{\makebox[3\qimgw][c]{\tiny\textbf{\textcolor{refgreen}{Real}}}} &
      \tiny\textbf{\textcolor{candorange}{Fake}} & \multicolumn{3}{@{}c@{}}{\makebox[3\qimgw][c]{\tiny\textbf{\textcolor{refgreen}{Real}}}} &
      \tiny\textbf{\textcolor{candorange}{Fake}} & \multicolumn{3}{@{}c@{}}{\makebox[3\qimgw][c]{\tiny\textbf{\textcolor{refgreen}{Real}}}} \\[2pt]
      \qimg[candorange]{\includegraphics{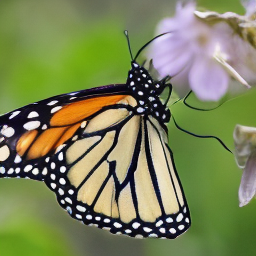}} & \qimg{\includegraphics{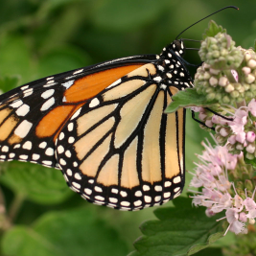}} & \qimg{\includegraphics{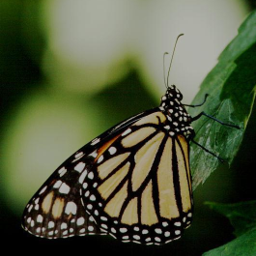}} & \qimg{\includegraphics{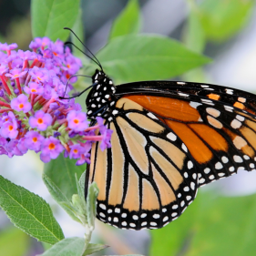}} &
      \qimg[candorange]{\includegraphics{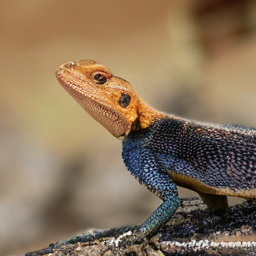}}    & \qimg{\includegraphics{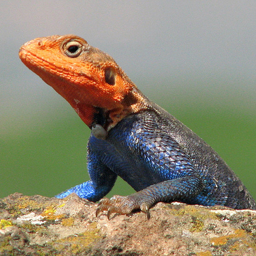}}    & \qimg{\includegraphics{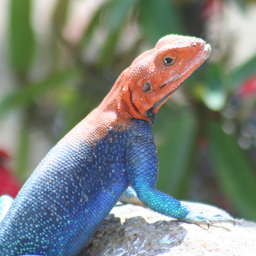}}    & \qimg{\includegraphics{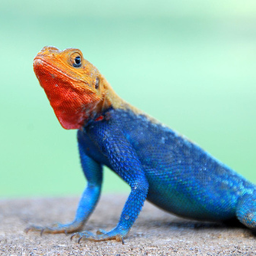}}    &
      \qimg[candorange]{\includegraphics{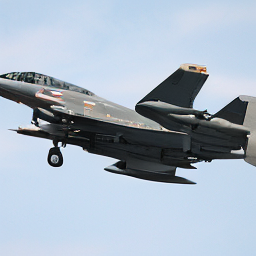}}    & \qimg{\includegraphics{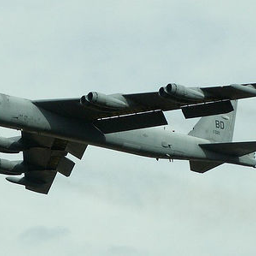}}    & \qimg{\includegraphics{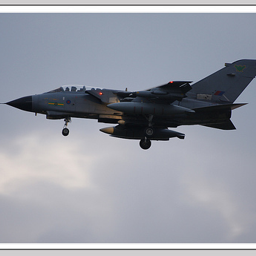}}    & \qimg{\includegraphics{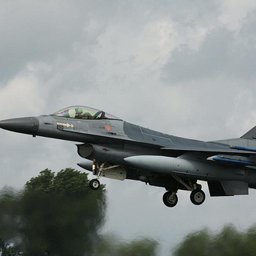}}    \\
      & \tiny$d\!=\!2.80$ & \tiny$d\!=\!3.05$ & \tiny$d\!=\!3.14$ &
      & \tiny$d\!=\!3.59$ & \tiny$d\!=\!3.92$ & \tiny$d\!=\!4.02$ &
      & \tiny$d\!=\!4.70$ & \tiny$d\!=\!4.76$ & \tiny$d\!=\!5.05$ \\[4pt]
      \qimg[candorange]{\includegraphics{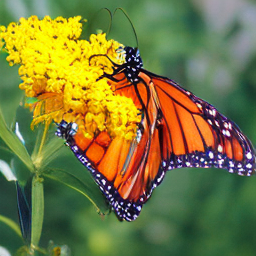}} & \qimg{\includegraphics{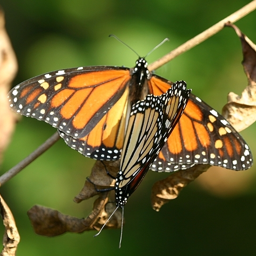}} & \qimg{\includegraphics{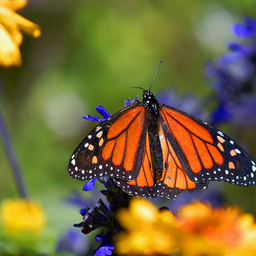}} & \qimg{\includegraphics{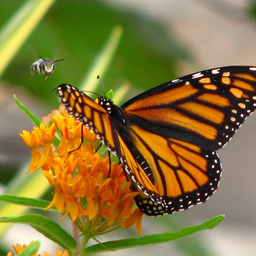}} &
      \qimg[candorange]{\includegraphics{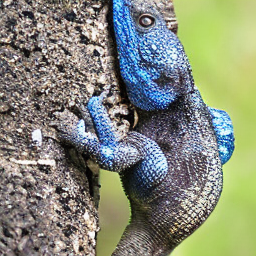}}    & \qimg{\includegraphics{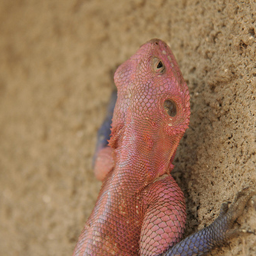}}    & \qimg{\includegraphics{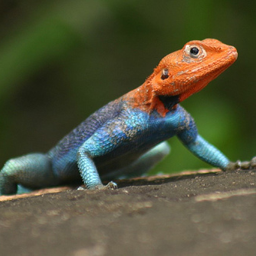}}    & \qimg{\includegraphics{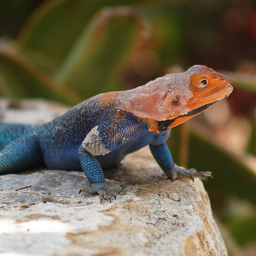}}    &
      \qimg[candorange]{\includegraphics{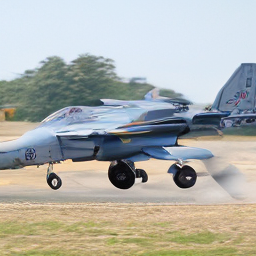}}    & \qimg{\includegraphics{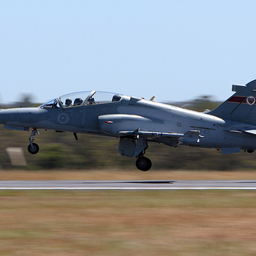}}    & \qimg{\includegraphics{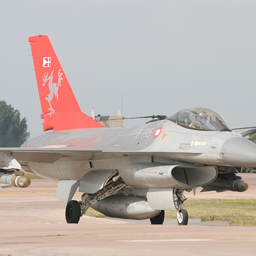}}    & \qimg{\includegraphics{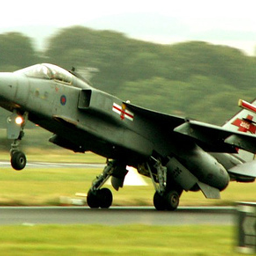}}    \\
      & \tiny$d\!=\!4.18$ & \tiny$d\!=\!4.33$ & \tiny$d\!=\!4.38$ &
      & \tiny$d\!=\!5.51$ & \tiny$d\!=\!5.54$ & \tiny$d\!=\!5.70$ &
      & \tiny$d\!=\!6.12$ & \tiny$d\!=\!6.26$ & \tiny$d\!=\!6.26$ \\[4pt]
      \qimg[candorange]{\includegraphics{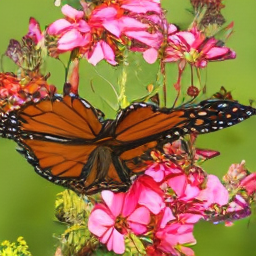}} & \qimg{\includegraphics{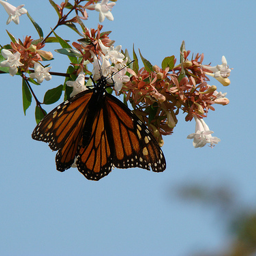}} & \qimg{\includegraphics{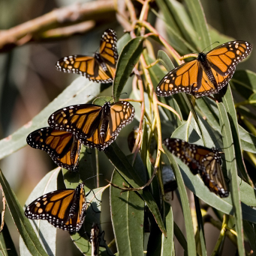}} & \qimg{\includegraphics{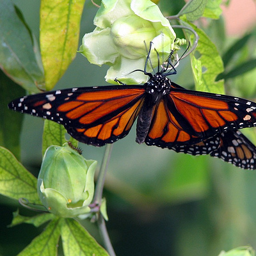}} &
      \qimg[candorange]{\includegraphics{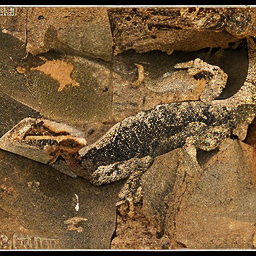}}    & \qimg{\includegraphics{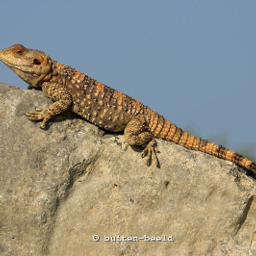}}    & \qimg{\includegraphics{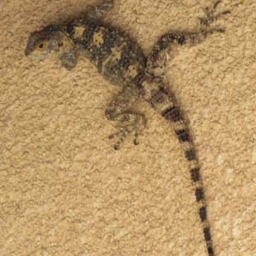}}    & \qimg{\includegraphics{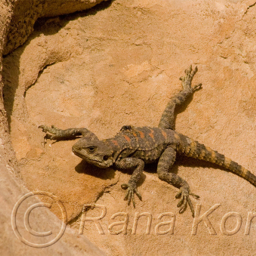}}    &
      \qimg[candorange]{\includegraphics{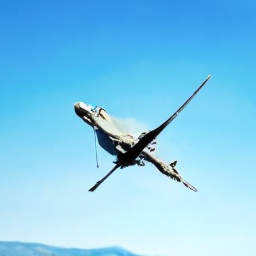}}    & \qimg{\includegraphics{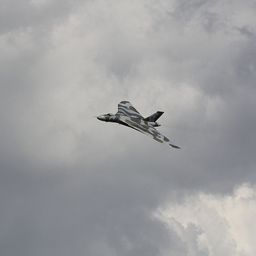}}    & \qimg{\includegraphics{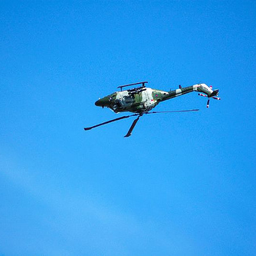}}    & \qimg{\includegraphics{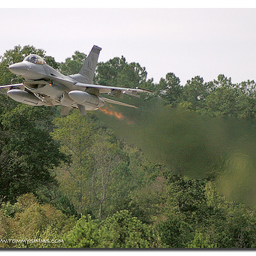}}    \\
      & \tiny$d\!=\!8.84$ & \tiny$d\!=\!9.27$ & \tiny$d\!=\!9.43$ &
      & \tiny$d\!=\!14.88$ & \tiny$d\!=\!15.04$ & \tiny$d\!=\!15.39$ &
      & \tiny$d\!=\!15.33$ & \tiny$d\!=\!15.57$ & \tiny$d\!=\!15.83$ \\
    \end{tabular}%
  \end{tabular}
  \caption{\textbf{Cross-set nearest-neighbor search distance as a fidelity signal.} Each column group corresponds to a distinct ImageNet class. In each group, the leftmost image is from the generated set, and the three images to its right are its nearest neighbors from the real set. Nearest neighbors are computed in Inception-v3 feature space, and \(d\) denotes the distance from each generated image to its nearest real neighbor. Rows are sorted by increasing nearest-neighbor distance. As the distance increases from top to bottom, the generated images become visibly less realistic.}
  \label{fig:visual-precision}
\end{figure}

Our aim in this paper is to propose a disentangled metric that reports fidelity and diversity separately, without performing a binary membership test on inflated estimated supports. Previous sections suggest that the problem is not nearest-neighbor search itself, but how existing metrics use it. KNN-ball metrics first build an estimated support from \emph{within-set} neighborhoods and then apply a binary membership test. We instead consider using nearest-neighbor search directly \emph{across} the real and generated sets, giving us a simple distance-based view of both fidelity and diversity.

\begin{figure}[t]
  \centering
  \setlength{\tabcolsep}{0pt}
  \renewcommand{\arraystretch}{0.6}
  \begin{tabular}{@{}c@{\hspace{4pt}}c@{}}
    \smash{\makebox[\arrowcol][c]{\rotatebox[origin=c]{90}{%
      \makebox[\gridwidth][c]{$\longleftarrow$\;\textbf{Increasing distance}\;$\longleftarrow$}%
    }}} &
    \begin{tabular}{@{}c@{\hspace{4pt}}ccc@{\hspace{12pt}}c@{\hspace{4pt}}ccc@{\hspace{12pt}}c@{\hspace{4pt}}ccc@{}}
      \tiny\textbf{\textcolor{refgreen}{Real}} & \multicolumn{3}{@{}c@{}}{\makebox[3\qimgw][c]{\tiny\textbf{\textcolor{candorange}{Fake}}}} &
      \tiny\textbf{\textcolor{refgreen}{Real}} & \multicolumn{3}{@{}c@{}}{\makebox[3\qimgw][c]{\tiny\textbf{\textcolor{candorange}{Fake}}}} &
      \tiny\textbf{\textcolor{refgreen}{Real}} & \multicolumn{3}{@{}c@{}}{\makebox[3\qimgw][c]{\tiny\textbf{\textcolor{candorange}{Fake}}}} \\[2pt]
      \rimg[refgreen]{\includegraphics{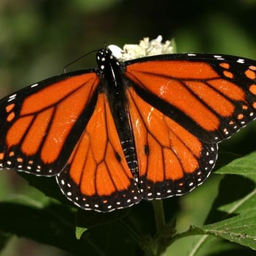}} & \rimg{\includegraphics{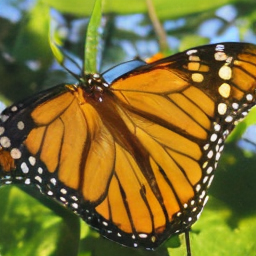}} & \rimg{\includegraphics{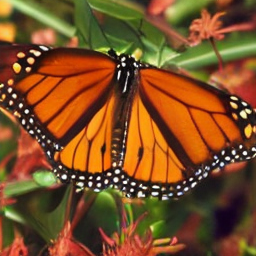}} & \rimg{\includegraphics{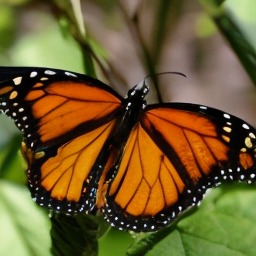}} &
      \rimg[refgreen]{\includegraphics{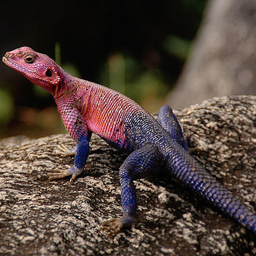}}    & \rimg{\includegraphics{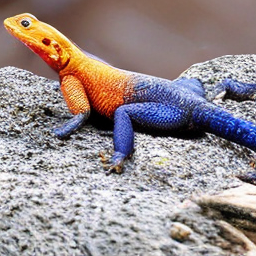}}    & \rimg{\includegraphics{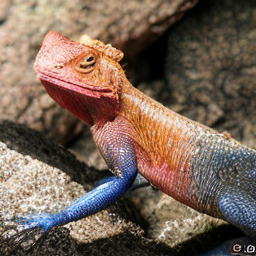}}    & \rimg{\includegraphics{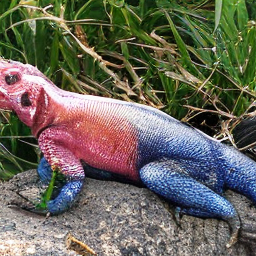}}    &
      \rimg[refgreen]{\includegraphics{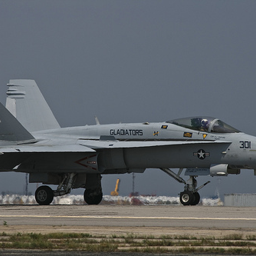}}    & \rimg{\includegraphics{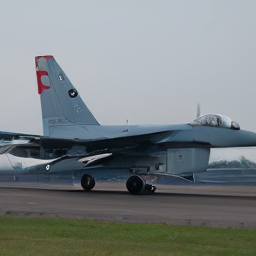}}    & \rimg{\includegraphics{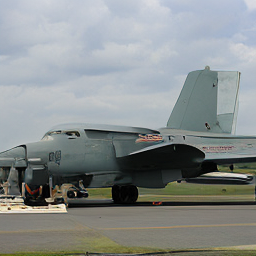}}    & \rimg{\includegraphics{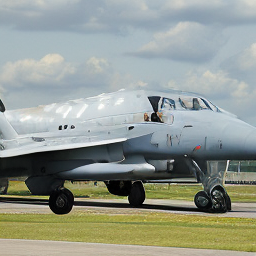}}    \\
      & \tiny$d\!=\!3.04$ & \tiny$d\!=\!3.06$ & \tiny$d\!=\!3.28$ &
      & \tiny$d\!=\!3.53$ & \tiny$d\!=\!3.70$ & \tiny$d\!=\!3.71$ &
      & \tiny$d\!=\!4.78$ & \tiny$d\!=\!4.84$ & \tiny$d\!=\!4.91$ \\[4pt]
      \rimg[refgreen]{\includegraphics{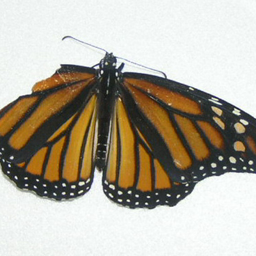}} & \rimg{\includegraphics{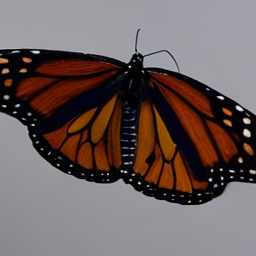}} & \rimg{\includegraphics{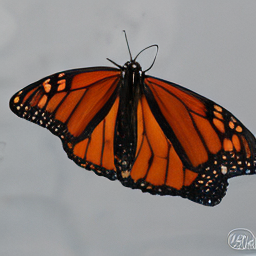}} & \rimg{\includegraphics{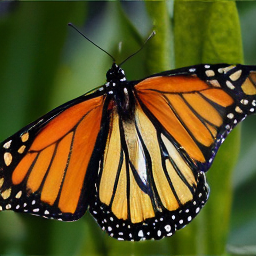}} &
      \rimg[refgreen]{\includegraphics{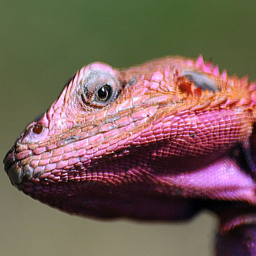}}    & \rimg{\includegraphics{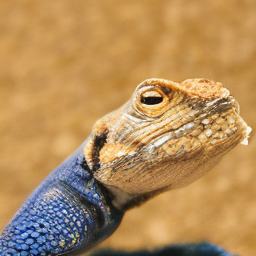}}    & \rimg{\includegraphics{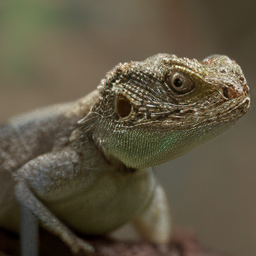}}    & \rimg{\includegraphics{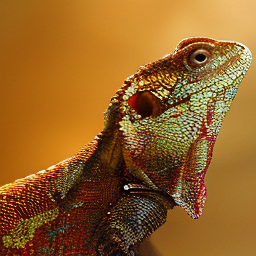}}    &
      \rimg[refgreen]{\includegraphics{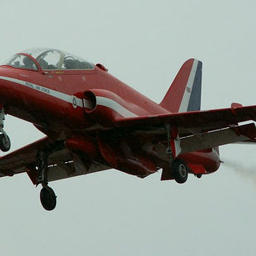}}    & \rimg{\includegraphics{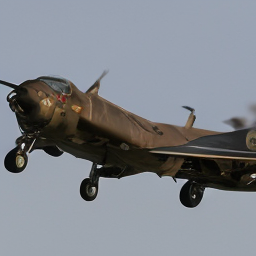}}    & \rimg{\includegraphics{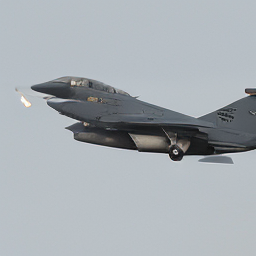}}    & \rimg{\includegraphics{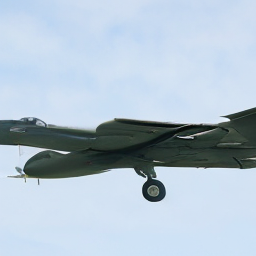}}    \\
      & \tiny$d\!=\!4.42$ & \tiny$d\!=\!4.98$ & \tiny$d\!=\!5.07$ &
      & \tiny$d\!=\!6.48$ & \tiny$d\!=\!6.49$ & \tiny$d\!=\!6.79$ &
      & \tiny$d\!=\!7.19$ & \tiny$d\!=\!7.31$ & \tiny$d\!=\!7.33$ \\[4pt]
      \rimg[refgreen]{\includegraphics{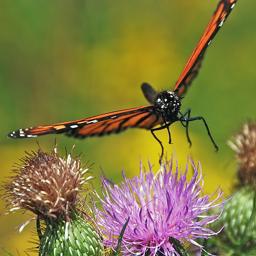}} & \rimg{\includegraphics{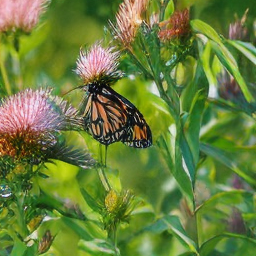}} & \rimg{\includegraphics{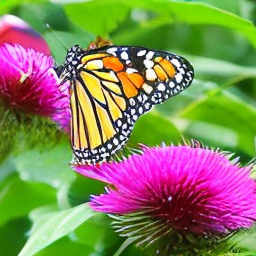}} & \rimg{\includegraphics{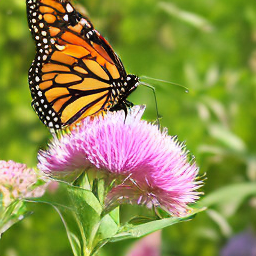}} &
      \rimg[refgreen]{\includegraphics{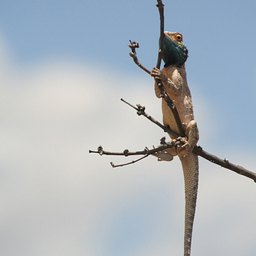}}    & \rimg{\includegraphics{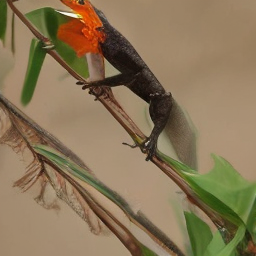}}    & \rimg{\includegraphics{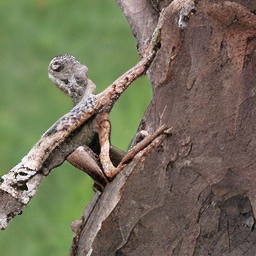}}    & \rimg{\includegraphics{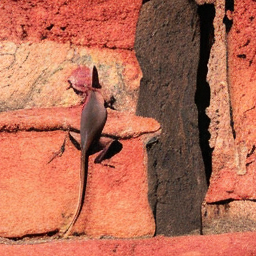}}    &
      \rimg[refgreen]{\includegraphics{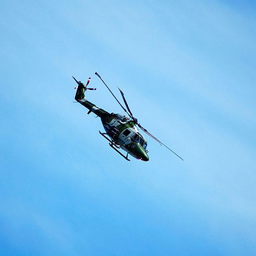}}    & \rimg{\includegraphics{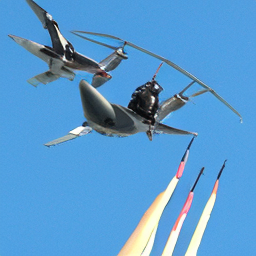}}    & \rimg{\includegraphics{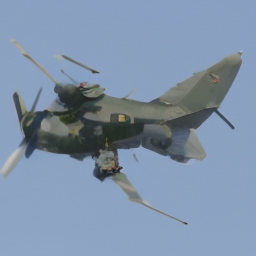}}    & \rimg{\includegraphics{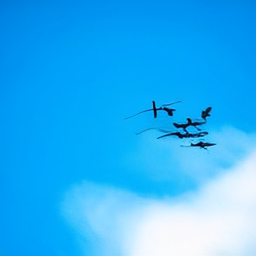}}    \\
      & \tiny$d\!=\!7.76$ & \tiny$d\!=\!7.91$ & \tiny$d\!=\!8.13$ &
      & \tiny$d\!=\!15.80$ & \tiny$d\!=\!16.42$ & \tiny$d\!=\!16.43$ &
      & \tiny$d\!=\!13.74$ & \tiny$d\!=\!13.76$ & \tiny$d\!=\!13.98$ \\
    \end{tabular}%
  \end{tabular}
  \caption{\textbf{Cross-set nearest-neighbor search distance as a diversity signal.} Each column group corresponds to a distinct ImageNet class. In each group, the leftmost image is from the real set, and the three images to its right are its nearest neighbors from the generated set. Nearest neighbors are computed in Inception-v3 feature space, and \(d\) denotes the distance from each real image to its nearest generated neighbor. Rows are sorted by increasing nearest-neighbor distance. As the distance increases from top to bottom, real images become less well represented by the generated set, indicating poorer coverage of the data.
}
  \label{fig:visual-recall}
\end{figure}

Figure~\ref{fig:visual-precision} illustrates the basic intuition for fidelity. For each generated query image, we retrieve its nearest neighbors in the real set. For this visualization, the real set consists of ImageNet-1K images, the generated set consists of samples from IMM-XL (8 steps)~\citep{zhou2025inductive}, and nearest-neighbor distances are computed in the Inception-v3 feature space.
If the distance is small, the retrieved real images are similar to the query, implying that the generated sample lies close to the real images. This indicates that this generated sample is realistic (top row in Figure~\ref{fig:visual-precision}).
If the distance is large, this implies that the generated sample is far from any real image, suggesting that it is unrealistic (bottom row in Figure~\ref{fig:visual-precision}). Figure~\ref{fig:visual-recall} illustrates the corresponding intuition for diversity. For each real image, we retrieve its nearest neighbors in the generated set. If the distance is small, the retrieved generated images are similar to the real query, implying that the generator has produced samples near this region of the real data distribution. This indicates that the corresponding real sample is well covered by the model (top row in Figure~\ref{fig:visual-recall}). If the distance is large, then the real query is far from any generated image, suggesting that the generator has failed to cover this region of the real distribution (bottom row in Figure~\ref{fig:visual-recall}).

We now turn this nearest-neighbor intuition into a statistical measure. Let \(P\) and \(Q\) be arbitrary distributions, and let \(P_n=\{x_1,\ldots,x_n\}\) and \(Q_m=\{y_1,\ldots,y_m\}\) be finite samples from them. For a query point \(x \sim P\), we measure its discrepancy from \(Q\) by averaging the distances to its \(K\) nearest neighbors in \(Q_m\). The total discrepancy from \(P\) to \(Q\) is then obtained by averaging this quantity over points drawn from \(P\):
\begin{equation}
    d(P_n,Q_m) = \E_{x \sim P_n}\!\left[\frac{1}{K}\sum_{j=1}^{K} \left \| f(x) - f(y_{j}^{*}) \right\|_{2} \right],
    \label{eq:kchamfer-one-sided}
\end{equation}
where \(y_{1}^{*}, \ldots, y_{K}^{*}\) are the \(K\) nearest neighbors of \(x\) in \(Q_m\), ordered by distance in feature space.

This construction is a direct generalization of the one-sided Chamfer distance~\citep{chamfer-og,chamfer-2,chamfer-3,chamfer-4}, a standard measure for comparing point clouds in 3D computer vision. When \(K=1\) and \(P_n\) and \(Q_m\) are finite 3D point sets, the symmetrized version recovers the classical Chamfer distance:
\[
d_{\text{Sym-Chamfer}}(P_n,Q_m)
=
\frac{1}{|P_n|}\sum_{x \in P_n}\min_{y \in Q_m}\|x-y\|_2
+
\frac{1}{|Q_m|}\sum_{y \in Q_m}\min_{x \in P_n}\|y-x\|_2.
\]
For this reason, we refer to our metric as \ourmetricshort: it preserves the nearest-neighbor interpretation of Chamfer distance while extending it to \(K\)-nearest-neighbor distances in high-dimensional feature spaces. For generative model evaluation, we take \(P_n\) to be samples from the generated distribution \(P\) and \(Q_m\) to be samples from the real distribution \(Q\). In this direction, \(d(P_n,Q_m)\) measures how close generated samples are to real samples, so we call it \textbf{\ourmetricshort-Fidelity}. Reversing the direction measures how well the generated distribution covers the real distribution, so \(d(Q_m,P_n)\) is \textbf{\ourmetricshort-Coverage}. Both metrics are distances, so lower values indicate better performance. 

Before moving to experiments, we summarize how our proposed metric avoids the issues with the popular metrics discussed above. Unlike distributional metrics, it reports fidelity and diversity as two separate metrics. Unlike KNN-ball metrics, it does not estimate a support or rely on binary membership tests. As a result, off-manifold samples are penalized by their actual distance to the real set rather than being accepted simply because they fall inside an inflated support estimate. For example, our metrics are non-zero for the two pentagons in Figure~\ref{fig:pentagon-sets}, correctly indicating that the two distributions are not identical, whereas other metrics in Table~\ref{tab:pentagon} report perfect scores. Additionally, because the final score is an average over per-sample distances, the metric is naturally diagnostic: each term corresponds to an individual generated or real sample, allowing practitioners to identify low-fidelity generations and under-covered regions of the data distribution.

\section{Experiments}

\textbf{Sensitivity to Image Distortions.}
We return to the image distortion experiment presented in Section~\ref{sec:limit-img}: despite visible degradation in the distorted images, KNN-ball-based metrics remained stagnant. As shown in Table~\ref{tab:noise-dino}, our proposed metrics \ourmetricshort{}-Fidelity and \ourmetricshort{}-Coverage increase monotonically with noise level, correctly reflecting the degradation in image quality.

\textbf{Sensitivity to Sampling Density.}
\ourmetric{} correctly responds to sampling density. Under the oracle setting of Table~\ref{tab:metrics}, where the generated set consists of real ImageNet images at three sizes (10K, 30K, 50K), \ourmetricshort{}-Coverage decreases monotonically from $10.57$ to $9.56$, reflecting that a denser generated set places its samples closer to the reference on average. 

\begin{figure}[]
\centering

\begin{minipage}{0.5\textwidth}
\centering
\small
\captionof{table}{\textbf{Sensitivity to number of neighbors $k$.} We evaluate 10K samples from iMF-XL/2~\citep{geng2025improved} (NFE=1) against 10K ImageNet reference images using Inception features.}
\label{tab:ablation_k}
\setlength{\tabcolsep}{3pt}\shadedrules
\resizebox{\textwidth}{!}{
\begin{tabular}{c S S S S >{\columncolor{oursbg}}S >{\columncolor{oursbg}}S}
\toprule
$k$\Tstrut\Bstrut & {Prec.} & {Recall} & {Density} & {Cov.} & {Chamfer-Fid.} & {Chamfer-Cov.}\\
\midrule
1\Tstrut   & 0.5207 & 0.4385 & 1.0620 & 0.5023 & 10.50 & 10.65\\
3   & 0.7811 & 0.6990 & 1.0779 & 0.8761  & 11.04 & 11.17\\
5   & 0.8729 & 0.8024 & 1.0745 & 0.9722 & 11.47 & 11.56\\
10  & 0.9468 & 0.9054 & 1.0742 & 0.9998 & 12.32 & 12.37\\
20  & 0.9753 & 0.9507 & 1.0841 & 1.0000 & 13.37 & 13.44\\
50  & 0.9889 & 0.9722 & 1.0935 & 1.0000 & 14.53 & 14.63\\
100\Bstrut & 0.9933 & 0.9817 & 1.0952 & 1.0000 & 15.12 & 15.24\\
\bottomrule
\end{tabular}
}
\end{minipage}
\hfill
\begin{minipage}{0.43\textwidth}
\centering
\includegraphics[width=\textwidth]{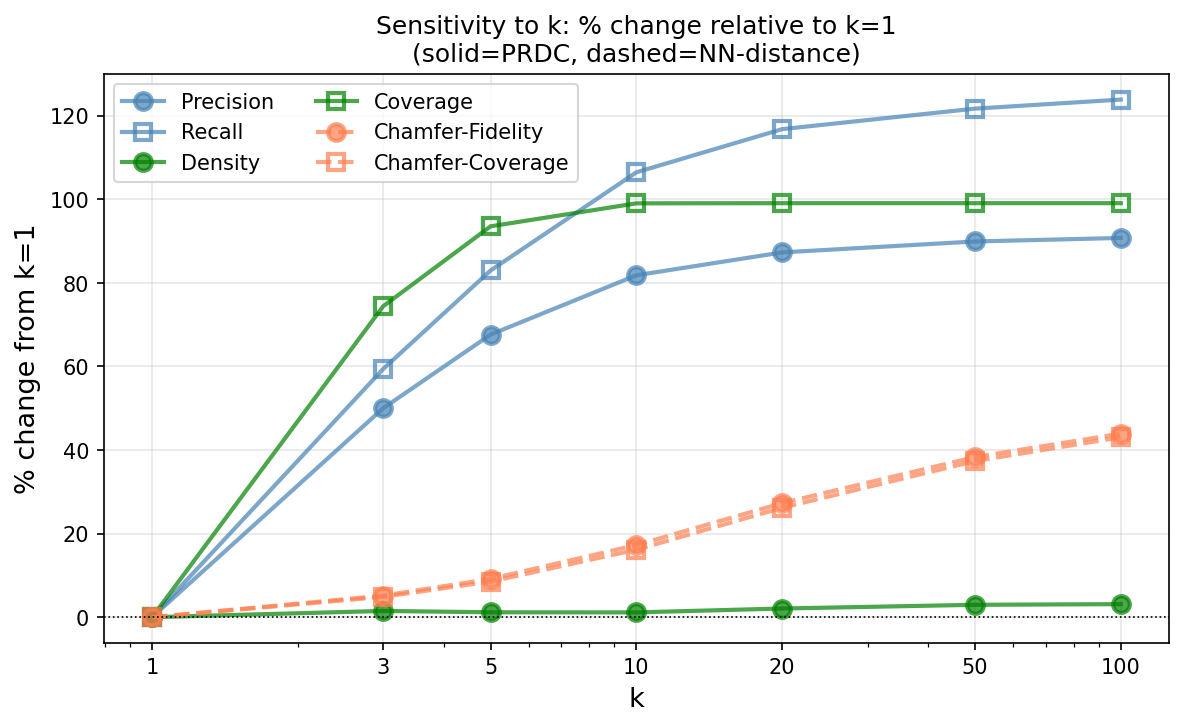}
\caption{\textbf{Sensitivity to number of neighbors.} While commonly used ball-based evaluation metrics exhibit high sensitivity to $k$, \textcolor{orange}{our metrics} show gradual change.}
\label{fig:ablation_k}
\end{minipage}

\end{figure}

\textbf{Sensitivity to Number of Neighbors.}
KNN-ball metrics are highly sensitive to the neighborhood parameter \(k\), since increasing \(k\) directly enlarges the estimated support. We quantify this dependence by evaluating \(10\)K iMeanFlow~\citep{geng2025improved} samples (NFE\(=1\)) against \(10\)K ImageNet references using Inception features, while sweeping \(k \in \{1,3,5,10,20,50,100\}\).
As shown in Table~\ref{tab:ablation_k} and Figure~\ref{fig:ablation_k}, changing \(k\) from \(1\) to \(3\) raises Precision from \(0.52\) to \(0.78\), Recall from \(0.43\) to \(0.69\), and Coverage from \(0.50\) to \(0.87\). As \(k\) grows, these binary membership scores approach saturation; Coverage reaches \(1.0\) by \(k=20\), after which it no longer distinguishes larger neighborhoods. In practice for ball-based metrics, \(k\) is a \emph{critical hyperparameter} that must be tuned for each evaluation setting. In contrast, our metrics change smoothly across the sweep, indicating lower sensitivity to \(k\).

\begin{table}[]
\centering
\small
\caption{\textbf{Wall-clock computation time.} We report post-feature-extraction runtime (in seconds) for FID, Precision/Recall, and \ourmetric{} across increasing reference and candidate set sizes.}
\label{tab:time}
\shadedrules
\begin{tabular}{cc cc >{\columncolor{oursbg}}c}
\toprule
Reference Size\Tstrut\Bstrut & Candidate Size& FID & Precision/Recall & \ourmetricshort \\
\midrule
\multirow{2}{*}{10K}
& 10K\Tstrut  &  8.36 &   1.71 & \textbf{1.00} \\
&  50K\Bstrut  &  9.67 &   6.84 & \textbf{1.15} \\
\midrule
\multirow{1}{*}{100K}
     & 100K\Tstrut\Bstrut & 25.95 &  52.12 &  \textbf{4.20} \\
\bottomrule
\end{tabular}
\end{table}

\textbf{Computational Efficiency.} We benchmark post-feature-extraction runtime for FID, Precision/Recall, and \ourmetric{} in Table~\ref{tab:time} using a single NVIDIA L40S. All metrics are computed from the same 2048-dimensional Inception-v3 features, so the reported times exclude feature extraction. \ourmetric{} is considerably faster than FID and Precision/Recall, allowing it to scale to much larger sets of reference and generated images. This efficiency arises from the fact that \ourmetric{} relies on nearest-neighbor lookups, which can be efficiently implemented using GPU-accelerated libraries like FAISS~\citep{faiss-2} (see Appendix~\ref{appendix:comp-analysis} for complexity analysis).

\section{Benchmarking Modern Generative Models}
We now turn to evaluating trained generative models. Table~\ref{tab:100k-combined-cos1-k1} reports evaluation results for class-conditional ImageNet generation, across popular models using 100K generated samples. We compute our metric with the $\ell_2$ distance and $k=1$ on normalized DINOv2-Giant (ViT-g/14) features~\citep{dinov2}, and report its values multiplied by 100 for readability. $\text{FD}_{\text{DINOv2}}$ uses unnormalized features, as originally proposed~\citep{dgm-eval}. Evaluation protocols are provided in Appendix~\ref{appendix:gen-zoo}. Additional results with unnormalized features and different values of $k$ are given in Appendix~\ref{appendix:additional-table}. 
Figure~\ref{fig:benchmark-bubble} plots \ourmetricshort{}-Fidelity against \ourmetricshort{}-Coverage for most of these models, showing at a glance how they differ along each axis. The marker area proportional to model size.

We observe that our proposed metric agrees with many accepted notions in the literature. For example, GANs, which are prone to mode collapse, have poor Chamfer-Coverage. For IMM, both Chamfer-Fidelity and Chamfer-Coverage decrease monotonically as NFE increases, consistent with the expectation that additional sampling steps refine generated samples. Beyond these expected trends, the benchmark also reveals the limitations of the metrics the field relies on, and what these limits have hidden about recent progress. 

\subsection{FID has become too saturated to be a reliable benchmark}

In the last decade, FID has arguably been the most popular metric for evaluating generative models. As models have improved, FID has become saturated, and we find that a lower FID is no longer clear evidence of a better model. 
\begin{itemize}[leftmargin=*,nosep,topsep=0pt]
    \item Drifting-L achieves a lower FID than iMF-XL/2 ($1.24$ against $1.48$), yet it is worse on every other metric, including $\text{FD}_{\text{DINOv2}}$ ($213.14$ against $138.65$), \ourmetricshort{}-Fidelity ($68.96$ against $61.67$) and \ourmetricshort{}-Coverage ($57.05$ against $54.47$). For further qualitative analysis, Section~\ref{sec:interpret-class} compares the two models on \ourmetricshort-Fidelity across different classes.
    \item StyleGAN-XL and DiT-XL/2 (250 steps) have similar FIDs ($2.09$ against $1.92$), yet DiT-XL/2 is far better on both \ourmetricshort{}-Fidelity ($58.00$ against $76.17$) and \ourmetricshort{}-Coverage ($53.91$ against $62.56$). Section~\ref{sec:interpret-fidelity} gives a detailed breakdown of these two models in terms of \ourmetricshort-Fidelity.
    \item JiT-H/16 and IMM (8 steps) have nearly the same FID ($1.65$ against $1.69$), yet JiT-H/16 is clearly better on both \ourmetricshort{}-Fidelity ($56.58$ against $63.62$) and \ourmetricshort{}-Coverage ($52.65$ against $54.89$).
\end{itemize}

Small differences in FID are therefore no longer a reliable sign of progress. In contrast, \ourmetric{} separates each of these pairs clearly, and shows whether the models differ in fidelity, coverage, or both.

\subsection{Few-step models still trail behind multi-step models}

In the last few years, several few-step generative models have been proposed that claim to match or surpass multi-step diffusion models such as DiT-XL/2. Examples of such models include IMM \cite{zhou2025inductive}, iMF \cite{geng2025improved}, Drifting \cite{deng2026generative} and Priming \cite{priming-models}. These claims rest mainly on FID, which, as shown above, has saturated as an evaluation metric. Our metric reveals that while their performance is remarkable, they still lag behind DiT-XL/2.
\begin{itemize}[leftmargin=*,nosep,topsep=0pt]
    \item Priming-L has one of the lowest FIDs among the models reported, less than half that of DiT-XL/2 (250 steps) ($0.93$ against $1.92$). Yet it is worse on both of our metrics.
    \item iMF-XL/2 also has a lower FID than DiT-XL/2 (250 steps). Yet it is worse on both \ourmetricshort{}-Fidelity ($61.67$ against $58.00$) and \ourmetricshort{}-Coverage ($54.47$ against $53.91$). Section~\ref{sec:interpret-fidelity} compares the two models sample by sample.
    \item IMM (8 steps) has a lower FID than DiT-XL/2 (250 steps). Yet it is worse on both \ourmetricshort{}-Fidelity ($63.62$ against $58.00$) and \ourmetricshort{}-Coverage ($54.89$ against $53.91$).
\end{itemize}

\subsection{\texorpdfstring{$\text{FD}_{\text{DINOv2}}$}{FD-DINOv2} is a bit more reliable than FID, but still deficient}

Recent works \citep{dgm-eval,yang2026representationfrechetlossvisual} have tried to move away from FID as a primary metric, and propose using encoders other than Inception. One example of such a metric is $\text{FD}_{\text{DINOv2}}$. However, like FID, this metrics only compare Gaussians fitted to the two distributions, and hence can suffer from the same problems we identified for FID in Figure~\ref{fig:pentagon}, Section~\ref{sec:limit-density} and Appendix~\ref{appendix:fid-limit}. Moreover, as single score, it conflates fidelity and coverage. Due to these limitations, $\text{FD}_{\text{DINOv2}}$ can favor a model that \ourmetric{} finds worse on both fidelity and coverage:

\begin{itemize}[leftmargin=*,nosep,topsep=0pt]
    \item $\text{FD}_{\text{DINOv2}}$ favors Priming-L over DiT-XL/2 (250 steps), yet DiT-XL/2 is better on both \ourmetricshort{}-Fidelity ($58.00$ against $60.66$) and \ourmetricshort{}-Coverage ($53.91$ against $54.09$).
    \item $\text{FD}_{\text{DINOv2}}$ again favors IMM (4 steps) over DiT-XL/2 (50 steps), yet DiT-XL/2 is better on both \ourmetricshort{}-Fidelity ($62.20$ against $67.01$) and \ourmetricshort{}-Coverage ($54.82$ against $55.63$).
\end{itemize}

Replacing the encoder is therefore not enough: as long as the metric is summarized by Gaussians and reports a single score, it cannot show where a model falls short.

\subsection{Precision and Recall aim to disentangle, but produce inconsistent rankings}

Although existing disentangled metrics like Precision and Recall aim to separate fidelity from coverage, they are unreliable due to their formulation. This unreliability stems from the issues discussed in Section \ref{sec:ball-limitations}.On the other hand, \ourmetric{} disentangles fidelity and coverage while producing reliable and meaningful rankings.

\begin{itemize}[leftmargin=*,nosep,topsep=0pt]
    \item Precision rates BigGAN on par with Drifting-L ($0.12$ against $0.13$) and above Drifting-B ($0.12$ against $0.11$), although BigGAN is considerably worse than both on every other metric.
    \item Recall rates IMM (1 step) above DiT-XL/2 (250 steps) ($0.38$ against $0.35$), although the former is worse on every other metric, including \ourmetricshort{}-Coverage ($59.99$ against $53.91$).
    \item For IMM, Recall peaks at 2 steps, where it is the highest of all models, and then decreases as the number of steps grows. This trend is counterintuitive, since performance generally improves with the number of steps.
    \ourmetricshort{}-Coverage, on the other hand, improves monotonically with the number of steps.
\end{itemize}

Such inconsistencies make Precision and Recall unsuitable for ranking models, even though they aim to report fidelity and coverage separately. On the other hand, \ourmetric{} disentangles fidelity and coverage, while producing reliable and meaningful rankings.

\begin{table}[ht]
\centering
\caption{\textbf{Evaluation of modern generative models on ImageNet.} Precision, Recall, \ourmetricshort{}-Fidelity and \ourmetricshort{}-Coverage use $\ell_2$-normalized DINOv2-Giant (ViT-g/14) features and $k=1$ nearest neighbor. Values are mean $\pm$ std over 3 runs, each drawing a new set of 100K generated samples. A std shown as $0.00$ is below 0.005.}
\label{tab:100k-combined-cos1-k1}
\resizebox{\ifdim\width>\textwidth\textwidth\else\width\fi}{!}{%
\sisetup{round-mode=places, round-precision=2}%
\shadedrules%
\newcommand{\famrule}{\arrayrulecolor{black!20}\specialrule{0.4pt}{0pt}{0pt}\arrayrulecolor{black}}
\newcommand{\meanstd}[2]{\mbox{\tablenum{#1}}\ensuremath{{}_{\pm\num{#2}}}}%
\newcommand{\meanph}[1]{\mbox{\tablenum{#1}}\ensuremath{{}_{\pm\textcolor{red}{?}}}}
\begin{tabular}{l
  S[table-format=4.0, round-mode=none]
  S[table-format=3.0, round-mode=none]
  >{\sisetup{table-format=3.2}}c
  >{\sisetup{table-format=1.2}}c
  >{\sisetup{table-format=1.2}}c
  >{\columncolor{oursbg}\sisetup{table-format=2.2}}c
  >{\columncolor{oursbg}\sisetup{table-format=2.2}}c
  >{\sisetup{table-format=1.2}}c}
\toprule
\Tstrut\Bstrut & & & \multicolumn{5}{c}{\textbf{DINOv2-Giant (ViT-g/14)}} & \multicolumn{1}{c}{\textbf{Inception-v3}} \\
\cmidrule(lr){4-8} \cmidrule(lr){9-9}
\textbf{Model}\Tstrut\Bstrut
  & {\textbf{Params (M)}}
  & {\textbf{NFE} \dn}
  & {\textbf{$\text{FD}_{\text{DINOv2}}$} \dn}
  & {\textbf{Precision} \up}
  & {\textbf{Recall} \up}
  & {\textbf{Chamfer-Fid.} \dn}
  & {\textbf{Chamfer-Cov.} \dn}
  & {\textbf{FID} \dn} \\
\midrule
\multicolumn{3}{l}{\textit{GANs}}\Tstrut & & & & & & \\
BigGAN~\citep{brock2018large}                & 112  & 1   & \meanstd{743.29}{0.85} & \meanstd{0.1246}{0.0007} & \meanstd{0.0605}{0.0032} & \meanstd{81.821}{0.090} & \meanstd{70.670}{0.140} & \meanstd{6.665}{0.047} \\
StyleGAN-XL~\citep{sauer2022stylegan}\Bstrut 
                                             & 166  & 1   & \meanstd{317.998}{0.81} & \meanstd{0.0845}{0.0008} & \meanstd{0.1786}{0.0023} & \meanstd{76.173}{0.095} & \meanstd{62.563}{0.120} & \meanstd{2.087}{0.056} \\
\midrule
\multicolumn{3}{l}{\textit{Diffusion and flow matching}}\Tstrut & & & & & & \\
DiT-XL/2 (50 steps)~\citep{peebles2023scalable} & 675  & 50  & \meanstd{168.70}{1.03} & \meanstd{0.2091}{0.0009} & \meanstd{0.3113}{0.0023} & \meanstd{62.201}{0.055} & \meanstd{54.822}{0.038} & \meanstd{3.363}{0.057} \\
DiT-XL/2 (125 steps)                         & 675  & 125 & \meanstd{120.10}{0.38} & \meanstd{0.2452}{0.0017} & \meanstd{0.3380}{0.0021} & \meanstd{58.875}{0.066} & \meanstd{54.097}{0.012} & \meanstd{2.068}{0.045} \\
DiT-XL/2 (250 steps)\Bstrut                  & 675  & 250 & \meanstd{108.55}{0.86} & \meanstd{0.2564}{0.0010} & \meanstd{0.3506}{0.0019} & \meanstd{57.995}{0.024} & \meanstd{53.906}{0.003} & \meanstd{1.918}{0.011} \\
\famrule
SiT-XL/2~\citep{ma2024sit}\Tstrut            & 675  & 250 & \meanstd{151.16}{0.72} & \meanstd{0.1991}{0.0005} & \meanstd{0.3584}{0.0022} & \meanstd{63.101}{0.137} & \meanstd{54.794}{0.044} & \meanstd{1.742}{0.012} \\
SiT-XL/2 + REPA \citep{repa} 
                                             & 675  & 250 & \meanstd{75.59}{0.50} & \meanstd{0.2916}{0.0009} & \meanstd{0.4109}{0.0030} & \meanstd{55.843}{0.038} & \meanstd{52.249}{0.026} & \meanstd{1.026}{0.009} \\
JiT-H/16 \citep{jit-diffusion} \Bstrut 
                                             & 953  & 99  & \meanstd{79.93}{0.21} & \meanstd{0.2735}{0.0010} & \meanstd{0.3962}{0.0014} & \meanstd{56.577}{0.044} & \meanstd{52.653}{0.015} & \meanstd{1.653}{0.010} \\
\midrule
\multicolumn{3}{l}{\textit{Few-step}}\Tstrut & & & & & & \\
IMM (1 step)~\citep{zhou2025inductive}       & 675  & 1   & \meanstd{384.37}{1.77} & \meanstd{0.0874}{0.0013} & \meanstd{0.3802}{0.0021} & \meanstd{80.503}{0.090} & \meanstd{59.989}{0.057} & \meanstd{8.107}{0.079} \\
IMM (2 steps)                                & 675  & 2   & \meanstd{230.07}{0.61} & \meanstd{0.1282}{0.0017} & \meanstd{0.4498}{0.0009} & \meanstd{72.097}{0.063} & \meanstd{56.951}{0.036} & \meanstd{3.674}{0.052} \\
IMM (4 steps)                                & 675  & 4   & \meanstd{163.24}{0.86} & \meanstd{0.1634}{0.0008} & \meanstd{0.4465}{0.0030} & \meanstd{67.013}{0.120} & \meanstd{55.628}{0.043} & \meanstd{2.190}{0.014} \\
IMM (8 steps)\Bstrut                         & 675  & 8   & \meanstd{128.80}{0.31} & \meanstd{0.1910}{0.0015} & \meanstd{0.4394}{0.0029} & \meanstd{63.617}{0.029} & \meanstd{54.886}{0.053} & \meanstd{1.693}{0.010} \\
\famrule
iMF-B/2~\citep{geng2025improved}\Tstrut      & 89   & 1   & \meanstd{295.71}{0.79} & \meanstd{0.1190}{0.0005} & \meanstd{0.1628}{0.0014} & \meanstd{69.625}{0.074} & \meanstd{59.038}{0.049} & \meanstd{3.169}{0.019} \\
iMF-M/2                                      & 174  & 1   & \meanstd{207.64}{1.13} & \meanstd{0.1533}{0.0010} & \meanstd{0.2411}{0.0028} & \meanstd{65.837}{0.113} & \meanstd{56.645}{0.056} & \meanstd{2.094}{0.018} \\
iMF-L/2                                      & 409  & 1   & \meanstd{153.50}{0.18} & \meanstd{0.1863}{0.0007} & \meanstd{0.3058}{0.0014} & \meanstd{62.687}{0.045} & \meanstd{55.062}{0.031} & \meanstd{1.582}{0.019} \\
iMF-XL/2\Bstrut                              & 610  & 1   & \meanstd{138.65}{0.41} & \meanstd{0.2005}{0.0014} & \meanstd{0.3304}{0.0036} & \meanstd{61.670}{0.095} & \meanstd{54.465}{0.028} & \meanstd{1.477}{0.011} \\
\famrule
pMF-H/16~\citep{pixel-mean-flows}\Tstrut\Bstrut
                                             & 956  & 1   & \meanstd{111.19}{0.57} & \meanstd{0.2380}{0.0018} & \meanstd{0.3505}{0.0011} & \meanstd{59.890}{0.052} & \meanstd{53.763}{0.040} & \meanstd{1.969}{0.015} \\
\famrule
Drifting-B~\citep{deng2026generative}\Tstrut    & 133  & 1   & \meanstd{248.38}{0.32} & \meanstd{0.1124}{0.0010} & \meanstd{0.3056}{0.0018} & \meanstd{70.973}{0.019} & \meanstd{58.114}{0.027} & \meanstd{1.436}{0.004} \\
Drifting-L\Bstrut                               & 463  & 1   & \meanstd{213.14}{1.14} & \meanstd{0.1288}{0.0003} & \meanstd{0.3105}{0.0043} & \meanstd{68.955}{0.116} & \meanstd{57.048}{0.030} & \meanstd{1.238}{0.005} \\
\famrule
Priming-L~\citep{priming-models}\Bstrut\Tstrut & 405  & 1   & \meanstd{88.12}{0.18} & \meanstd{0.2005}{0.0012} & \meanstd{0.4102}{0.0034} & \meanstd{60.662}{0.031} & \meanstd{54.092}{0.005} & \meanstd{0.925}{0.006} \\
\midrule
\multicolumn{3}{l}{\textit{Autoregressive and masked}}\Tstrut & & & & & & \\
VAR-d20~\citep{tian2024visual}               & 600  & 10  & \meanstd{147.80}{0.22} & \meanstd{0.2086}{0.0003} & \meanstd{0.2978}{0.0026} & \meanstd{60.885}{0.067} & \meanstd{55.041}{0.031} & \meanstd{2.389}{0.002} \\
VAR-d24                                      & 1000 & 10  & \meanstd{105.44}{0.14} & \meanstd{0.2580}{0.0008} & \meanstd{0.3540}{0.0018} & \meanstd{57.828}{0.040} & \meanstd{53.364}{0.011} & \meanstd{1.848}{0.004} \\
VAR-d30\Bstrut                               & 2000 & 10  & \meanstd{104.07}{0.20} & \meanstd{0.2644}{0.0003} & \meanstd{0.3834}{0.0025} & \meanstd{58.290}{0.053} & \meanstd{52.770}{0.004} & \meanstd{1.631}{0.011} \\
\famrule
LlamaGen-L~\citep{sun2024autoregressive}\Tstrut & 343  & 256 & \meanstd{269.79}{0.63} & \meanstd{0.1790}{0.0010} & \meanstd{0.2266}{0.0009} & \meanstd{64.218}{0.110} & \meanstd{58.059}{0.017} & \meanstd{4.131}{0.008} \\
Open-MAGVIT2-XL~\citep{open-magvit2}         & 1500 & 256 & \meanstd{145.70}{0.25} & \meanstd{0.2029}{0.0012} & \meanstd{0.3515}{0.0005} & \meanstd{62.118}{0.010} & \meanstd{54.939}{0.029} & \meanstd{2.093}{0.014} \\
RAR-XL~\citep{rar-xl}                        & 955  & 256 & \meanstd{263.36}{0.11} & \meanstd{0.1487}{0.0002} & \meanstd{0.2521}{0.0011} & \meanstd{67.593}{0.004} & \meanstd{56.908}{0.017} & \meanstd{1.194}{0.002} \\
RandAR-XL~\citep{randar-xl}                  & 775  & 88  & \meanstd{243.24}{0.59} & \meanstd{0.1867}{0.0001} & \meanstd{0.2796}{0.0020} & \meanstd{63.803}{0.053} & \meanstd{56.419}{0.027} & \meanstd{1.900}{0.013} \\
TiTok-L32~\citep{tiktok-32-tokens}\Bstrut    & 177  & 8   & \meanstd{377.30}{0.67} & \meanstd{0.1579}{0.0009} & \meanstd{0.1947}{0.0018} & \meanstd{75.845}{0.041} & \meanstd{57.942}{0.050} & \meanstd{2.497}{0.029} \\
\midrule
\multicolumn{3}{l}{\textit{FD-loss variants \citep{yang2026representationfrechetlossvisual}}}\Tstrut & & & & & & \\
pMF-H + FD-loss (Inception)                  & 956  & 1   & \meanstd{68.44}{0.13} & \meanstd{0.2567}{0.0019} & \meanstd{0.4229}{0.0012} & \meanstd{57.521}{0.035} & \meanstd{52.943}{0.015} & \meanstd{0.393}{0.004} \\
pMF-H + FD-loss (SIM)                        & 956  & 1   & \meanstd{35.88}{0.05} & \meanstd{0.3525}{0.0005} & \meanstd{0.3793}{0.0010} & \meanstd{52.017}{0.097} & \meanstd{51.831}{0.001} & \meanstd{0.433}{0.001} \\
JiT-H + FD-loss (Inception)                  & 953  & 1   & \meanstd{124.22}{0.17} & \meanstd{0.2061}{0.0027} & \meanstd{0.3282}{0.0011} & \meanstd{61.714}{0.036} & \meanstd{54.454}{0.012} & \meanstd{0.394}{0.003} \\
JiT-H + FD-loss (SIM)\Bstrut                 & 953  & 1   & \meanstd{51.90}{0.19} & \meanstd{0.3263}{0.0019} & \meanstd{0.2879}{0.0002} & \meanstd{52.753}{0.063} & \meanstd{52.511}{0.013} & \meanstd{0.433}{0.001} \\
\bottomrule
\end{tabular}%
        }
\end{table}

\begin{figure}[h]
\centering
\includegraphics[width=\linewidth]{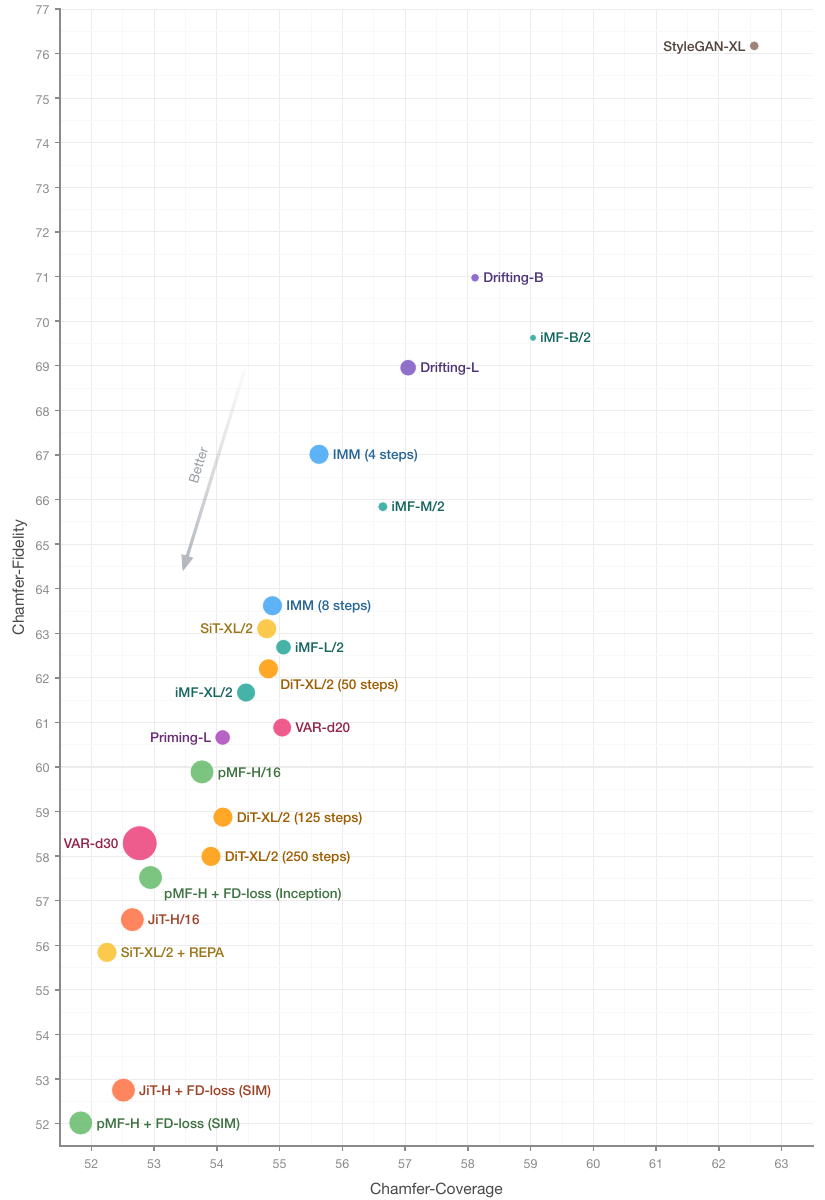}
\caption{\textbf{Fidelity and coverage of modern generative models.} \ourmetricshort{}-Fidelity and \ourmetricshort{}-Coverage on ImageNet 256 for a subset of the models in Table~\ref{tab:100k-combined-cos1-k1}, computed with the $\ell_2$ distance on normalized DINOv2-Giant features and multiplied by 100. Lower is better on both axes. Marker area is proportional to the number of parameters.}
\label{fig:benchmark-bubble}
\end{figure}
\clearpage

\section{Qualitative Analysis}
\label{sec:interpret}

\ourmetric{} is an average of per-sample distances, each measuring how far an individual image is from the other set. Hence, the reported score can be traced back to individual images. This property of \ourmetric{} enables practitioners to interpret, analyze and diagnose the performance of generative models at a granular level. As a result, the reported score is not an abstract number: it can be explained by the images that produce it. In this section, we use these per-sample distances in three ways: to break down the performance of a model by class (Section~\ref{sec:interpret-class}), to compare models using the distribution of per-sample distances (Sections~\ref{sec:interpret-fidelity} and~\ref{sec:interpret-coverage}), and to find the real images that existing models fail to cover on average (Section~\ref{sec:interpret-coverage-across}).


\subsection{Per-sample \ourmetricshort{}-Fidelity for a class}
\label{sec:interpret-class}

\begin{figure}[h]
    \centering
    \definecolor{imfblue}{HTML}{1F5FAF}
    \definecolor{driftorange}{HTML}{C4501F}
    \ifdefined\spreadnat\else\newlength{\spreadnat}\fi
    \settowidth{\spreadnat}{\includegraphics{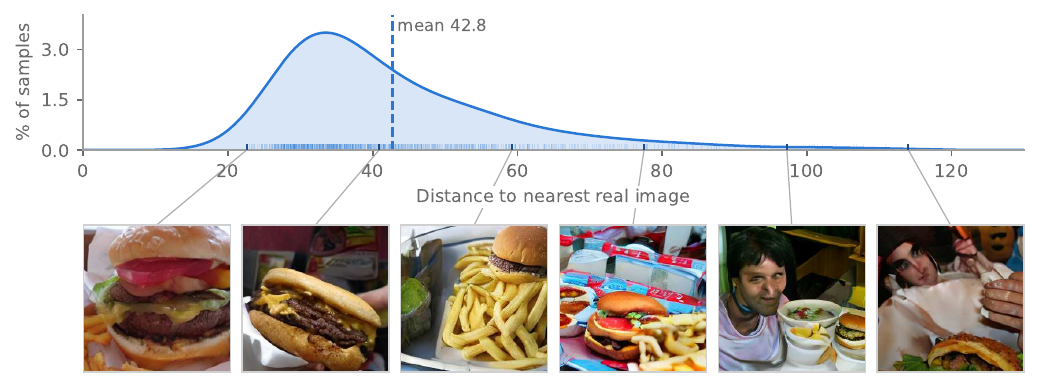}}%
    \makeatletter
    \edef\spreadscale{\strip@pt\dimexpr 1pt*\dimexpr 0.47\linewidth\relax/\spreadnat\relax}%
    \makeatother
    \newcommand{\spreadpanel}[1]{\raisebox{-.5\height}{#1}}%
    \newcommand{\spreadrow}[3]{%
        \raisebox{-.5\height}{\rotatebox{90}{\small #1}} &
        \spreadpanel{#2} &
        \spreadpanel{#3} \\}%
    \newcommand{\classsep}{\arrayrulecolor{black!18}\midrule}
    \setlength{\tabcolsep}{3pt}%
    \setlength{\aboverulesep}{5pt}%
    \setlength{\belowrulesep}{5pt}%
    \setlength{\lightrulewidth}{0.4pt}%
    \begin{tabular}{@{}c r r@{}}
        & \multicolumn{1}{c}{\small\color{imfblue} iMeanFlow} & \multicolumn{1}{c@{}}{\small\color{driftorange} Drifting-L} \\[3pt]
        \spreadrow{macaw}
            {\includegraphics[scale=\spreadscale]{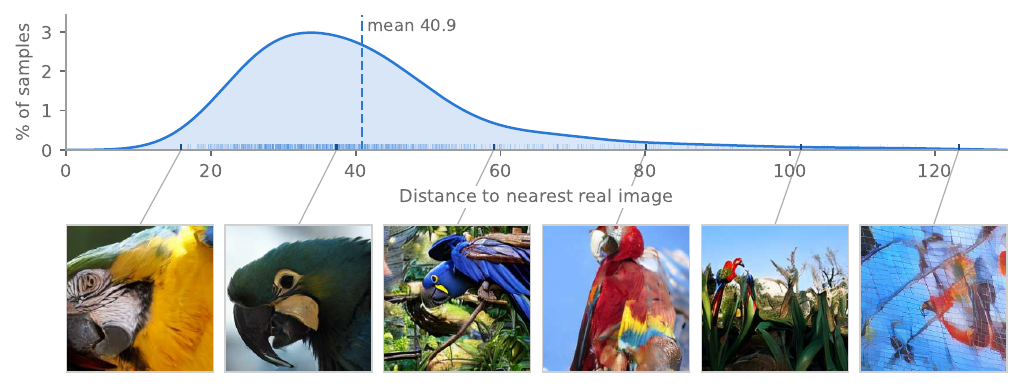}}
            {\includegraphics[scale=\spreadscale]{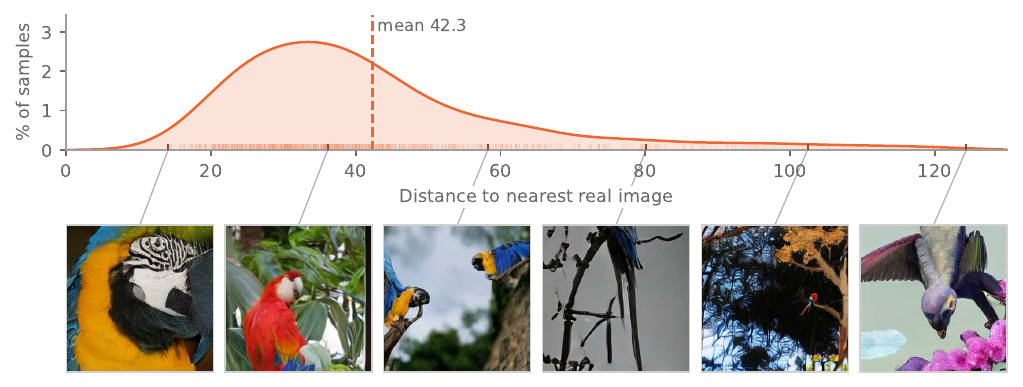}}
        \classsep
        \spreadrow{cheeseburger}
            {\includegraphics[scale=\spreadscale]{figures/interpret/class_spread_1k/cheeseburger_imeanflow.pdf}}
            {\includegraphics[scale=\spreadscale]{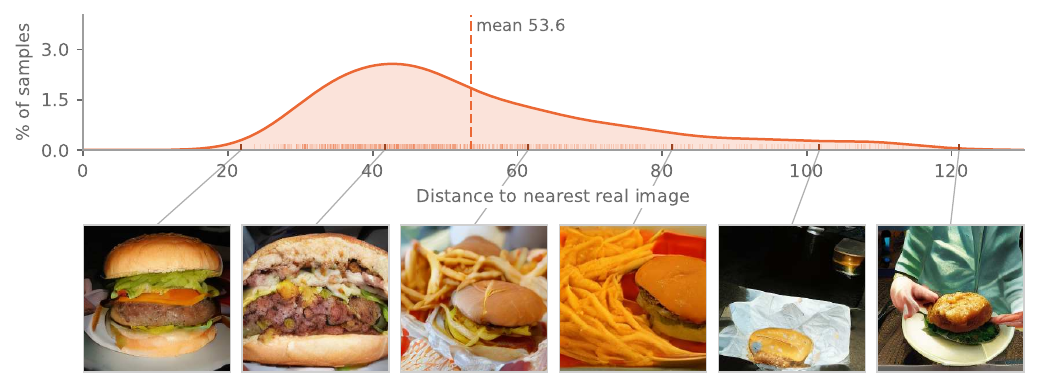}}
        \classsep
        \spreadrow{hot-air balloon}
            {\includegraphics[scale=\spreadscale]{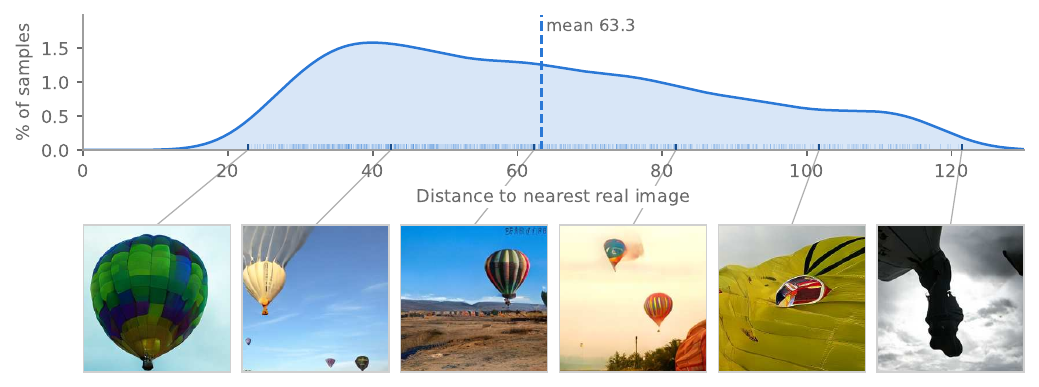}}
            {\includegraphics[scale=\spreadscale]{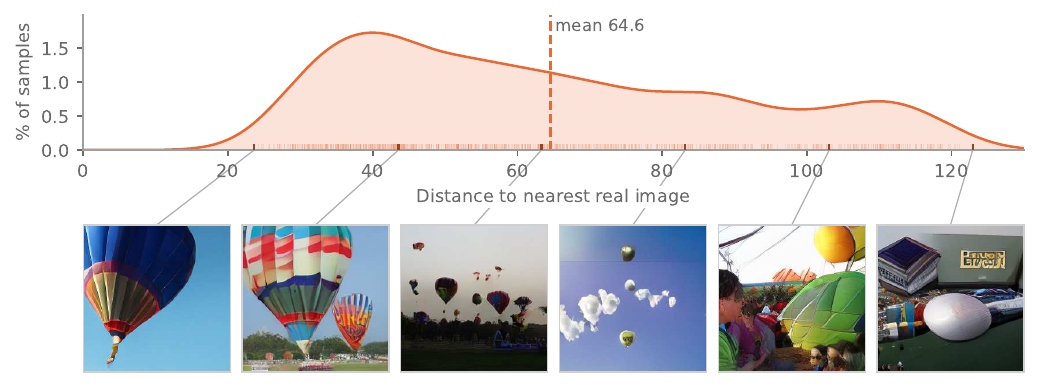}}
        \classsep
        \spreadrow{hummingbird}
            {\includegraphics[scale=\spreadscale]{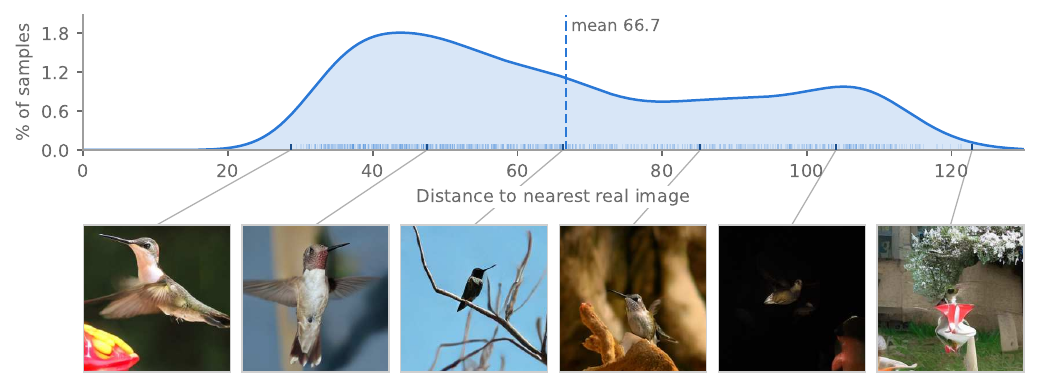}}
            {\includegraphics[scale=\spreadscale]{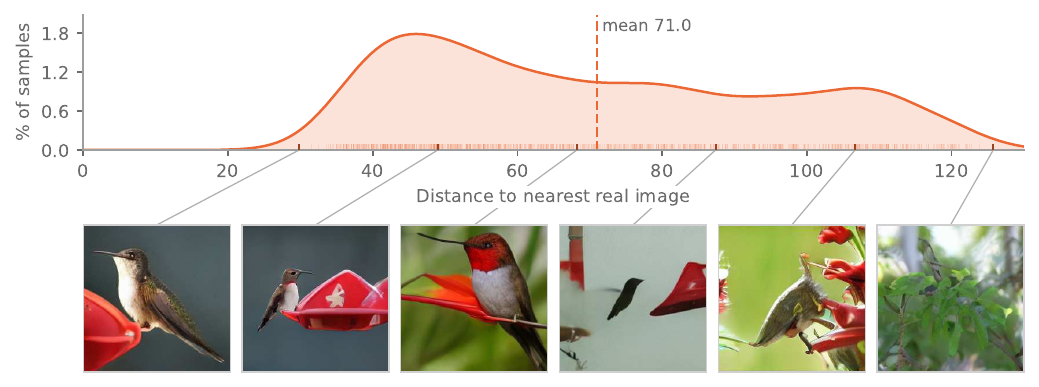}}
    \end{tabular}
    \caption{\textbf{Per-sample \ourmetricshort-Fidelity within a class.} Distribution of per-sample distances over 1{,}000 samples per class; the dashed line is the class mean. Six samples are shown at evenly spaced distances. Image quality decreases as distance grows.}
    \label{fig:interpret-class-spread}
\end{figure}

We can decompose the performance of a model by an individual class for ImageNet class-conditional generation. Consider Figure~\ref{fig:interpret-class-spread}, which shows the performance of two one-step models, iMeanFlow (iMF-XL/2) and Drifting-L, across different classes. For each class, we plot the distance of every generated sample to its nearest real image of the same class. This gives us per-sample Chamfer-Fidelity for that class. Note that as these distances increase, the generated samples become less realistic. We can observe that the performance of both models is not uniform across classes: they model some classes, such as cheeseburger, considerably better than others, such as hot-air balloon. This analysis also enables comparing models on the same class. For example, for the classes shown in Figure~\ref{fig:interpret-class-spread}, iMeanFlow performs better than Drifting-L. Figure~\ref{fig:interpret-class-spread-extra} shows the same analysis for additional classes.

\clearpage

\subsection{Comparing models using per-sample \ourmetricshort{}-Fidelity}
\label{sec:interpret-fidelity}

\begin{figure}[h]
    \centering
    \includegraphics[width=0.95\linewidth]{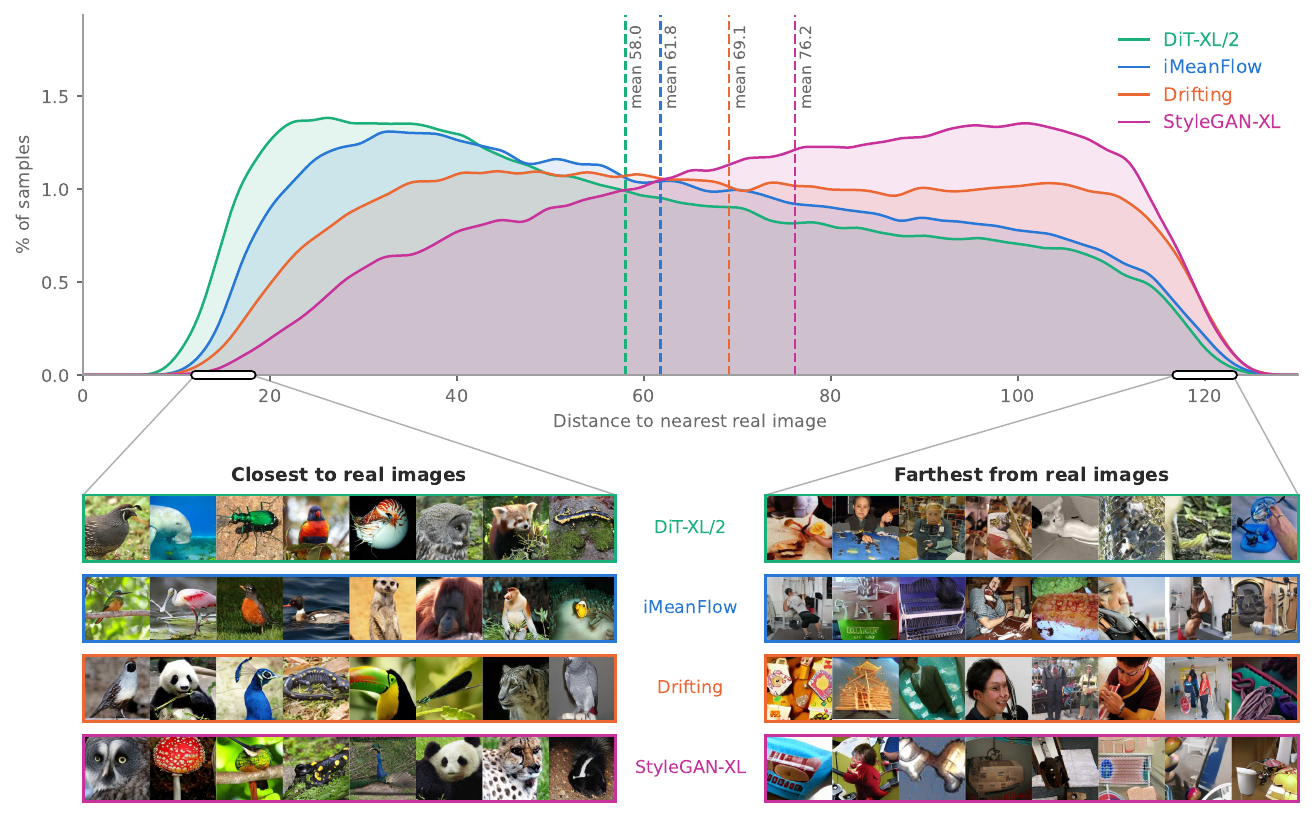}
    \caption{\textbf{Per-sample \ourmetricshort-Fidelity across models.} Dashed lines mark each model's mean (single seed). Bottom: samples closest to and farthest from the real set.}
    \label{fig:interpret-model-compare-v8}
\end{figure}

We can extend the previous analysis from a single class to all classes. Consider Figure~\ref{fig:interpret-model-compare-v8}, which shows the performance of four models: DiT-XL/2 (250 steps, cfg=1.5), iMeanFlow, Drifting-L, and StyleGAN-XL. For each model, we plot the distance of every generated sample to its nearest real image. The means of these distributions follow the same order as \ourmetricshort-Fidelity in Table~\ref{tab:100k-combined-cos1-k1}. 

We can also inspect the samples at both ends of each distribution. For every model, the samples closest to the real images are clean and well formed, while the farthest samples show cluttered compositions.

The distributions also differ in shape. For DiT-XL/2, most samples lie relatively close to their nearest real image, while the opposite is true for StyleGAN-XL. This difference is reflected in their respective \ourmetricshort{}-Fidelity scores, which are the means of the distributions. FID, in contrast, rates the two models as similar ($1.92$ against $2.09$).

\clearpage

\subsection{Comparing models using per-sample \ourmetricshort{}-Coverage}
\label{sec:interpret-coverage}

\begin{figure}[h]
    \centering
    \includegraphics[width=0.95\linewidth]{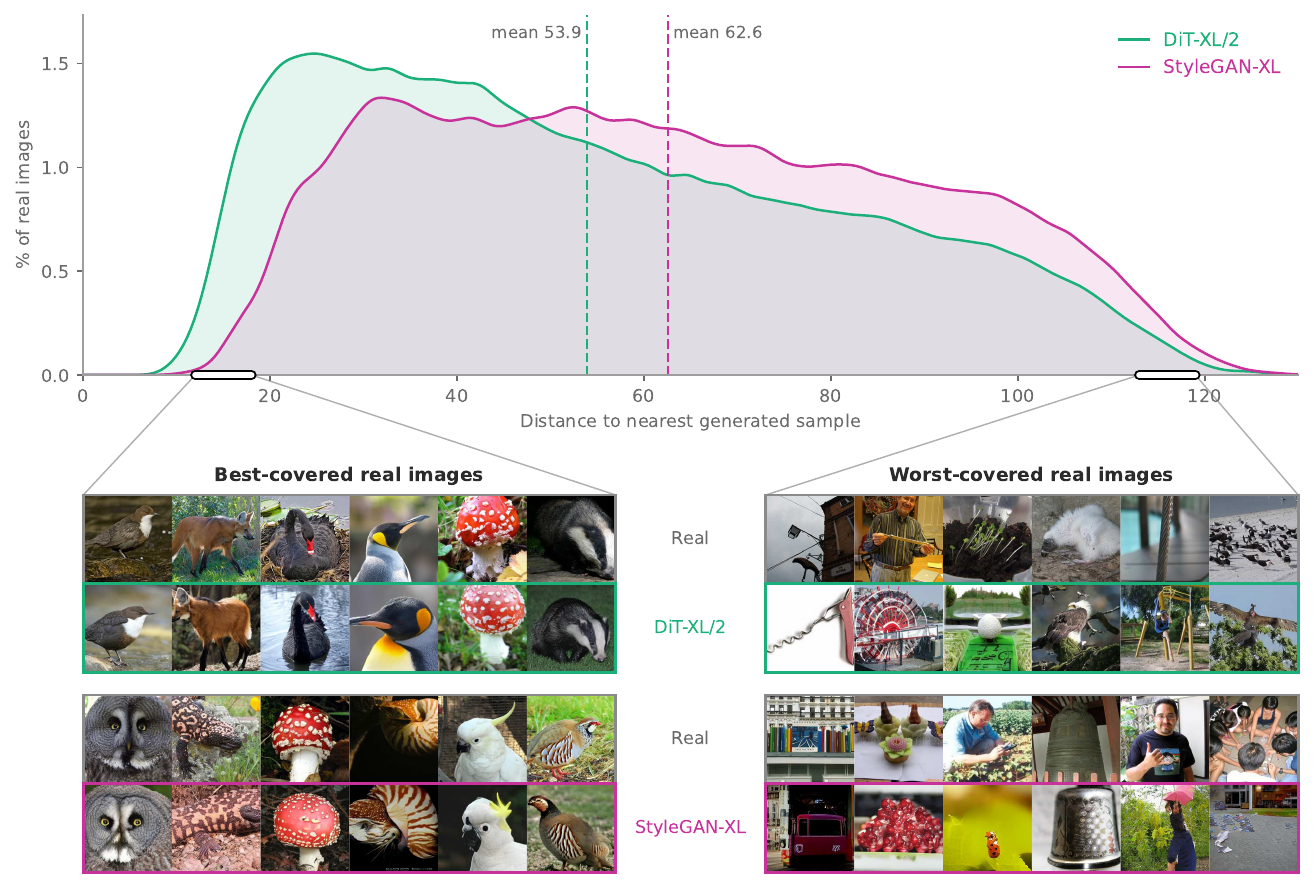}
    \caption{\textbf{Per-sample \ourmetricshort-Coverage.} Dashed lines mark each model's mean (single seed). Bottom: best- and worst-covered real images, each above its nearest generated sample.}
    \label{fig:interpret-coverage-compare}
\end{figure}

We can perform a similar analysis for coverage. Consider Figure~\ref{fig:interpret-coverage-compare}, which shows the performance of two models, DiT-XL/2 and StyleGAN-XL. For each model, we plot the distance of every real image to its nearest generated sample. DiT-XL/2 has a lower mean ($53.9$ against $62.6$), consistent with its better \ourmetricshort-Coverage in Table~\ref{tab:100k-combined-cos1-k1}. We can also inspect the real images at both ends of each distribution, along with their nearest generated sample. Real images with a small distance have a close match in the generated set, while real images with a large distance do not. Hence, per-sample \ourmetricshort-Coverage shows which real images a model fails to represent.

\clearpage

\subsection{Average per-sample \ourmetricshort{}-Coverage across models}
\label{sec:interpret-coverage-across}

\begin{figure}[h]
    \centering
    \includegraphics[width=0.95\linewidth]{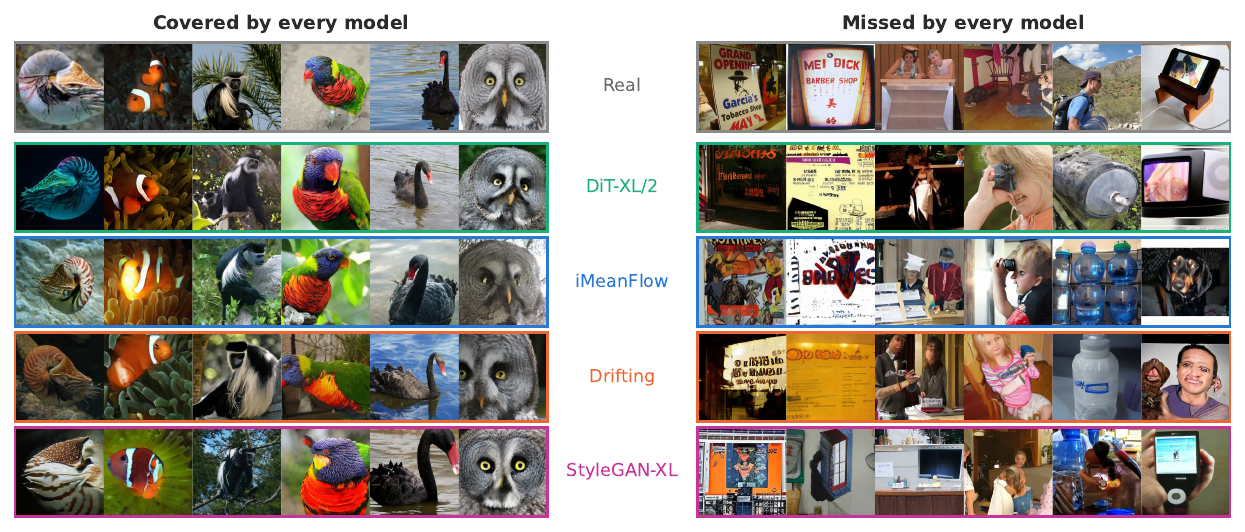}
    \caption{\textbf{Real images covered and missed by every model.} Top row: real image. Below: its nearest generated sample from each model.}
    \label{fig:interpret-coverage-across}
\end{figure}

We can also combine per-sample \ourmetricshort-Coverage across models. For each real image, we compute its per-sample \ourmetricshort-Coverage under each of the four models and take the average of these four scores. This gives a single score for every real image, which indicates how easy that image is to model on average across different generators. Consider Figure~\ref{fig:interpret-coverage-across}, which shows the real images with the lowest and highest scores, along with their nearest generated sample from each model. Images with the lowest scores are, on average, well covered by the four models, while images with the highest scores are poorly covered. Such analysis identifies the parts of the data distribution that remain difficult for these existing generative models.


\section{Discussions and Conclusion}

\textbf{Future Work} This work studies existing generative evaluation metrics from an empirical perspective and proposes \ourmetric{} as an alternative that avoids their limitations. We note that \ourmetric{} is encoder-agnostic and can be paired with feature spaces that better capture human perception, aesthetic quality, semantic alignment, or domain-specific similarity. Studying these encoder choices is an important direction left for future work.

\textbf{Conclusion} Widely used evaluation metrics often conflate fidelity and diversity into a single symmetric score. Existing metrics that report the two separately rely on inflated support estimates, which makes them unreliable. We propose \ourmetric{}, a disentangled metric based on cross-set nearest-neighbor, which provides interpretable signals for fidelity and coverage. Across our analyses, \ourmetric{} is less sensitive to the neighborhood parameter, responds more meaningfully to image corruption and sampling density, and remains computationally efficient.

\textbf{Acknowledgments}
This research was enabled in part by NSERC, the Canada Foundation for Innovation, the Canada CIFAR AI Chairs program, the BC DRI Group and the Digital Research Alliance of Canada.

\clearpage

\bibliography{references}

@misc{fid,
      title={GANs Trained by a Two Time-Scale Update Rule Converge to a Local Nash Equilibrium}, 
      author={Martin Heusel and Hubert Ramsauer and Thomas Unterthiner and Bernhard Nessler and Sepp Hochreiter},
      year={2018},
      eprint={1706.08500},
      archivePrefix={arXiv},
      primaryClass={cs.LG},
      url={https://arxiv.org/abs/1706.08500}, 
}

@misc{repa,
      title={Representation Alignment for Generation: Training Diffusion Transformers Is Easier Than You Think}, 
      author={Sihyun Yu and Sangkyung Kwak and Huiwon Jang and Jongheon Jeong and Jonathan Huang and Jinwoo Shin and Saining Xie},
      year={2025},
      eprint={2410.06940},
      archivePrefix={arXiv},
      primaryClass={cs.CV},
      url={https://arxiv.org/abs/2410.06940}, 
}

@misc{improved-pr,
      title={Improved Precision and Recall Metric for Assessing Generative Models}, 
      author={Tuomas Kynkäänniemi and Tero Karras and Samuli Laine and Jaakko Lehtinen and Timo Aila},
      year={2019},
      eprint={1904.06991},
      archivePrefix={arXiv},
      primaryClass={stat.ML},
      url={https://arxiv.org/abs/1904.06991}, 
}

@misc{density-and-coverage,
      title={Reliable Fidelity and Diversity Metrics for Generative Models}, 
      author={Muhammad Ferjad Naeem and Seong Joon Oh and Youngjung Uh and Yunjey Choi and Jaejun Yoo},
      year={2020},
      eprint={2002.09797},
      archivePrefix={arXiv},
      primaryClass={cs.CV},
      url={https://arxiv.org/abs/2002.09797}, 
}

@misc{assym-pr,
      title={Emergent Asymmetry of Precision and Recall for Measuring Fidelity and Diversity of Generative Models in High Dimensions}, 
      author={Mahyar Khayatkhoei and Wael AbdAlmageed},
      year={2023},
      eprint={2306.09618},
      archivePrefix={arXiv},
      primaryClass={cs.LG},
      url={https://arxiv.org/abs/2306.09618}, 
}

@misc{kid,
      title={Demystifying MMD GANs}, 
      author={Mikołaj Bińkowski and Danica J. Sutherland and Michael Arbel and Arthur Gretton},
      year={2021},
      eprint={1801.01401},
      archivePrefix={arXiv},
      primaryClass={stat.ML},
      url={https://arxiv.org/abs/1801.01401}, 
}

@misc{rethinking-fid,
      title={Rethinking FID: Towards a Better Evaluation Metric for Image Generation}, 
      author={Sadeep Jayasumana and Srikumar Ramalingam and Andreas Veit and Daniel Glasner and Ayan Chakrabarti and Sanjiv Kumar},
      year={2024},
      eprint={2401.09603},
      archivePrefix={arXiv},
      primaryClass={cs.CV},
      url={https://arxiv.org/abs/2401.09603}, 
}

@article{deng2026generative,
  title={Generative Modeling via Drifting},
  author={Deng, Mingyang and Li, He and Li, Tianhong and Du, Yilun and He, Kaiming},
  journal={arXiv preprint arXiv:2602.04770},
  year={2026}
}

@misc{priming-models,
      title={Are We Overthinking One-Step Generation?}, 
      author={Woobin Park and Jaeyeon Kim and Yilun Du and Sitan Chen and Bohyung Han},
      year={2026},
      eprint={2610.20331},
      archivePrefix={arXiv},
      primaryClass={cs.LG},
      url={https://arxiv.org/abs/2610.20331}, 
}

@misc{jit-diffusion,
      title={Back to Basics: Let Denoising Generative Models Denoise}, 
      author={Tianhong Li and Kaiming He},
      year={2026},
      eprint={2511.13720},
      archivePrefix={arXiv},
      primaryClass={cs.CV},
      url={https://arxiv.org/abs/2511.13720}, 
}

@misc{pixel-mean-flows,
      title={One-step Latent-free Image Generation with Pixel Mean Flows}, 
      author={Yiyang Lu and Susie Lu and Qiao Sun and Hanhong Zhao and Zhicheng Jiang and Xianbang Wang and Tianhong Li and Zhengyang Geng and Kaiming He},
      year={2026},
      eprint={2601.22158},
      archivePrefix={arXiv},
      primaryClass={cs.CV},
      url={https://arxiv.org/abs/2601.22158}, 
}

@misc{tiktok-32-tokens,
      title={An Image is Worth 32 Tokens for Reconstruction and Generation}, 
      author={Qihang Yu and Mark Weber and Xueqing Deng and Xiaohui Shen and Daniel Cremers and Liang-Chieh Chen},
      year={2024},
      eprint={2406.07550},
      archivePrefix={arXiv},
      primaryClass={cs.CV},
      url={https://arxiv.org/abs/2406.07550}, 
}

@misc{randar-xl,
      title={RandAR: Decoder-only Autoregressive Visual Generation in Random Orders}, 
      author={Ziqi Pang and Tianyuan Zhang and Fujun Luan and Yunze Man and Hao Tan and Kai Zhang and William T. Freeman and Yu-Xiong Wang},
      year={2025},
      eprint={2412.01827},
      archivePrefix={arXiv},
      primaryClass={cs.CV},
      url={https://arxiv.org/abs/2412.01827}, 
}

@misc{rar-xl,
      title={Randomized Autoregressive Visual Generation}, 
      author={Qihang Yu and Ju He and Xueqing Deng and Xiaohui Shen and Liang-Chieh Chen},
      year={2024},
      eprint={2411.00776},
      archivePrefix={arXiv},
      primaryClass={cs.CV},
      url={https://arxiv.org/abs/2411.00776}, 
}

@misc{open-magvit2,
      title={Open-MAGVIT2: An Open-Source Project Toward Democratizing Auto-regressive Visual Generation}, 
      author={Zhuoyan Luo and Fengyuan Shi and Yixiao Ge and Yujiu Yang and Limin Wang and Ying Shan},
      year={2025},
      eprint={2409.04410},
      archivePrefix={arXiv},
      primaryClass={cs.CV},
      url={https://arxiv.org/abs/2409.04410}, 
}

@article{sun2024autoregressive,
  title={Autoregressive Model Beats Diffusion: Llama for Scalable Image Generation},
  author={Sun, Peize and Jiang, Yi and Chen, Shoufa and Zhang, Shilong and Peng, Bingyue and Luo, Ping and Yuan, Zehuan},
  journal={arXiv preprint arXiv:2406.06525},
  year={2024}
}

@article{mmd-test,
  author  = {Arthur Gretton and Karsten M. Borgwardt and Malte J. Rasch and Bernhard Sch{{\"o}}lkopf and Alexander Smola},
  title   = {A Kernel Two-Sample Test},
  journal = {Journal of Machine Learning Research},
  year    = {2012},
  volume  = {13},
  number  = {25},
  pages   = {723-773},
  url     = {http://jmlr.org/papers/v13/gretton12a.html}
}

@misc{wacv-gaussian-mixtures,
      title={Evaluating generative networks using Gaussian mixtures of image features}, 
      author={Lorenzo Luzi and Carlos Ortiz Marrero and Nile Wynar and Richard G. Baraniuk and Michael J. Henry},
      year={2022},
      eprint={2110.05240},
      archivePrefix={arXiv},
      primaryClass={cs.CV},
      url={https://arxiv.org/abs/2110.05240}, 
}

@misc{faiss-2,
      title={The Faiss library},
      author={Matthijs Douze and Alexandr Guzhva and Chengqi Deng and Jeff Johnson and Gergely Szilvasy and Pierre-Emmanuel Mazar{\'e} and Maria Lomeli and Lucas Hosseini and Herv{\'e} J{\'e}gou},
      year={2025},
      eprint={2401.08281},
      archivePrefix={arXiv},
      primaryClass={cs.LG},
      url={https://arxiv.org/abs/2401.08281},
}

@misc{inception,
      title={Going Deeper with Convolutions}, 
      author={Christian Szegedy and Wei Liu and Yangqing Jia and Pierre Sermanet and Scott Reed and Dragomir Anguelov and Dumitru Erhan and Vincent Vanhoucke and Andrew Rabinovich},
      year={2014},
      eprint={1409.4842},
      archivePrefix={arXiv},
      primaryClass={cs.CV},
      url={https://arxiv.org/abs/1409.4842}, 
}

@misc{imagenet-1k,
      title={ImageNet Large Scale Visual Recognition Challenge}, 
      author={Olga Russakovsky and Jia Deng and Hao Su and Jonathan Krause and Sanjeev Satheesh and Sean Ma and Zhiheng Huang and Andrej Karpathy and Aditya Khosla and Michael Bernstein and Alexander C. Berg and Li Fei-Fei},
      year={2015},
      eprint={1409.0575},
      archivePrefix={arXiv},
      primaryClass={cs.CV},
      url={https://arxiv.org/abs/1409.0575}, 
}

@misc{dhariwal2021diffusionmodelsbeatgans,
      title={Diffusion Models Beat GANs on Image Synthesis}, 
      author={Prafulla Dhariwal and Alex Nichol},
      year={2021},
      eprint={2105.05233},
      archivePrefix={arXiv},
      primaryClass={cs.LG},
      url={https://arxiv.org/abs/2105.05233}, 
}

@inproceedings{
      zhou2025inductive,
      title={Inductive Moment Matching},
      author={Linqi Zhou and Stefano Ermon and Jiaming Song},
      booktitle={Forty-second International Conference on Machine Learning},
      year={2025}
}

@inproceedings{clean-fid,
  title={On Aliased Resizing and Surprising Subtleties in GAN Evaluation},
  author={Parmar, Gaurav and Zhang, Richard and Zhu, Jun-Yan},
  booktitle={CVPR},
  year={2022}
}

@misc{fid-infinite,
      title={Effectively Unbiased FID and Inception Score and where to find them}, 
      author={Min Jin Chong and David Forsyth},
      year={2020},
      eprint={1911.07023},
      archivePrefix={arXiv},
      primaryClass={cs.CV},
      url={https://arxiv.org/abs/1911.07023}, 
}

@inproceedings{dgm-eval,
  title={Exposing flaws of generative model evaluation metrics and their unfair treatment of diffusion models},
  author={Stein, George and Cresswell, Jesse and Hosseinzadeh, Rasa and Sui, Yi and Ross, Brendan and Villecroze, Valentin and Liu, Zhaoyan and Caterini, Anthony L and Taylor, Eric and Loaiza-Ganem, Gabriel},
  booktitle={Advances in Neural Information Processing Systems},
  volume={36},
  year={2023}
}

@inproceedings{chamfer-og,
      author = {Barrow, H. G. and Tenenbaum, J. M. and Bolles, R. C. and Wolf, H. C.},
      title = {Parametric correspondence and chamfer matching: two new techniques for image matching},
      year = {1977},
      publisher = {Morgan Kaufmann Publishers Inc.},
      address = {San Francisco, CA, USA},
      booktitle = {Proceedings of the 5th International Joint Conference on Artificial Intelligence - Volume 2},
      pages = {659–663},
      numpages = {5},
      location = {Cambridge, USA},
      series = {IJCAI'77}
}

@article{chamfer-2,
      author       = {Gunilla Borgefors},
      title        = {Hierarchical Chamfer Matching: {A} Parametric Edge Matching Algorithm},
      journal      = {{IEEE} Trans. Pattern Anal. Mach. Intell.},
      volume       = {10},
      number       = {6},
      pages        = {849--865},
      year         = {1988},
      url          = {https://doi.org/10.1109/34.9107},
      doi          = {10.1109/34.9107},
      bibsource    = {dblp computer science bibliography, https://dblp.org}
}

@misc{chamfer-3,
      title={A Point Set Generation Network for 3D Object Reconstruction from a Single Image}, 
      author={Haoqiang Fan and Hao Su and Leonidas Guibas},
      year={2016},
      eprint={1612.00603},
      archivePrefix={arXiv},
      primaryClass={cs.CV},
      url={https://arxiv.org/abs/1612.00603}, 
}

@misc{chamfer-4,
      title={Learning Representations and Generative Models for 3D Point Clouds}, 
      author={Panos Achlioptas and Olga Diamanti and Ioannis Mitliagkas and Leonidas Guibas},
      year={2018},
      eprint={1707.02392},
      archivePrefix={arXiv},
      primaryClass={cs.CV},
      url={https://arxiv.org/abs/1707.02392}, 
}

@misc{vae,
      title={Auto-Encoding Variational Bayes}, 
      author={Diederik P Kingma and Max Welling},
      year={2022},
      eprint={1312.6114},
      archivePrefix={arXiv},
      primaryClass={stat.ML},
      url={https://arxiv.org/abs/1312.6114}, 
}

@misc{gan,
      title={Generative Adversarial Networks}, 
      author={Ian J. Goodfellow and Jean Pouget-Abadie and Mehdi Mirza and Bing Xu and David Warde-Farley and Sherjil Ozair and Aaron Courville and Yoshua Bengio},
      year={2014},
      eprint={1406.2661},
      archivePrefix={arXiv},
      primaryClass={stat.ML},
      url={https://arxiv.org/abs/1406.2661}, 
}

@misc{dc-gan,
      title={Unsupervised Representation Learning with Deep Convolutional Generative Adversarial Networks}, 
      author={Alec Radford and Luke Metz and Soumith Chintala},
      year={2016},
      eprint={1511.06434},
      archivePrefix={arXiv},
      primaryClass={cs.LG},
      url={https://arxiv.org/abs/1511.06434}, 
}

@misc{ddpm,
      title={Denoising Diffusion Probabilistic Models}, 
      author={Jonathan Ho and Ajay Jain and Pieter Abbeel},
      year={2020},
      eprint={2006.11239},
      archivePrefix={arXiv},
      primaryClass={cs.LG},
      url={https://arxiv.org/abs/2006.11239}, 
}

@misc{flow-matching,
      title={Flow Matching for Generative Modeling}, 
      author={Yaron Lipman and Ricky T. Q. Chen and Heli Ben-Hamu and Maximilian Nickel and Matt Le},
      year={2023},
      eprint={2210.02747},
      archivePrefix={arXiv},
      primaryClass={cs.LG},
      url={https://arxiv.org/abs/2210.02747}, 
}

@misc{normalizing-flows,
      title={Variational Inference with Normalizing Flows}, 
      author={Danilo Jimenez Rezende and Shakir Mohamed},
      year={2016},
      eprint={1505.05770},
      archivePrefix={arXiv},
      primaryClass={stat.ML},
      url={https://arxiv.org/abs/1505.05770}, 
}

@misc{stylegan-xl,
      title={StyleGAN-XL: Scaling StyleGAN to Large Diverse Datasets}, 
      author={Axel Sauer and Katja Schwarz and Andreas Geiger},
      year={2022},
      eprint={2202.00273},
      archivePrefix={arXiv},
      primaryClass={cs.LG},
      url={https://arxiv.org/abs/2202.00273}, 
}

@misc{dinov2,
      title={DINOv2: Learning Robust Visual Features without Supervision}, 
      author={Maxime Oquab and Timothée Darcet and Théo Moutakanni and Huy Vo and Marc Szafraniec and Vasil Khalidov and Pierre Fernandez and Daniel Haziza and Francisco Massa and Alaaeldin El-Nouby and Mahmoud Assran and Nicolas Ballas and Wojciech Galuba and Russell Howes and Po-Yao Huang and Shang-Wen Li and Ishan Misra and Michael Rabbat and Vasu Sharma and Gabriel Synnaeve and Hu Xu and Hervé Jegou and Julien Mairal and Patrick Labatut and Armand Joulin and Piotr Bojanowski},
      year={2024},
      eprint={2304.07193},
      archivePrefix={arXiv},
      primaryClass={cs.CV},
      url={https://arxiv.org/abs/2304.07193}, 
}

@misc{clip,
      title={Learning Transferable Visual Models From Natural Language Supervision}, 
      author={Alec Radford and Jong Wook Kim and Chris Hallacy and Aditya Ramesh and Gabriel Goh and Sandhini Agarwal and Girish Sastry and Amanda Askell and Pamela Mishkin and Jack Clark and Gretchen Krueger and Ilya Sutskever},
      year={2021},
      eprint={2103.00020},
      archivePrefix={arXiv},
      primaryClass={cs.CV},
      url={https://arxiv.org/abs/2103.00020}, 
}

@article{geng2025improved,
  title={Improved mean flows: On the challenges of fastforward generative models},
  author={Geng, Zhengyang and Lu, Yiyang and Wu, Zongze and Shechtman, Eli and Kolter, J Zico and He, Kaiming},
  journal={arXiv preprint arXiv:2512.02012},
  year={2025}
}

@misc{yang2026representationfrechetlossvisual,
      title={Representation Fr\'echet Loss for Visual Generation}, 
      author={Jiawei Yang and Zhengyang Geng and Xuan Ju and Yonglong Tian and Yue Wang},
      year={2026},
      eprint={2604.28190},
      archivePrefix={arXiv},
      primaryClass={cs.CV},
      url={https://arxiv.org/abs/2604.28190}, 
}

@misc{stable-diffusion,
      title={High-Resolution Image Synthesis with Latent Diffusion Models}, 
      author={Robin Rombach and Andreas Blattmann and Dominik Lorenz and Patrick Esser and Björn Ommer},
      year={2022},
      eprint={2112.10752},
      archivePrefix={arXiv},
      primaryClass={cs.CV},
      url={https://arxiv.org/abs/2112.10752}, 
}

@misc{laion5b,
      title={LAION-5B: An open large-scale dataset for training next generation image-text models}, 
      author={Christoph Schuhmann and Romain Beaumont and Richard Vencu and Cade Gordon and Ross Wightman and Mehdi Cherti and Theo Coombes and Aarush Katta and Clayton Mullis and Mitchell Wortsman and Patrick Schramowski and Srivatsa Kundurthy and Katherine Crowson and Ludwig Schmidt and Robert Kaczmarczyk and Jenia Jitsev},
      year={2022},
      eprint={2210.08402},
      archivePrefix={arXiv},
      primaryClass={cs.CV},
      url={https://arxiv.org/abs/2210.08402}, 
}

@inproceedings{
brock2018large,
title={Large Scale {GAN} Training for High Fidelity Natural Image Synthesis},
author={Andrew Brock and Jeff Donahue and Karen Simonyan},
booktitle={International Conference on Learning Representations},
year={2019},
url={https://openreview.net/forum?id=B1xsqj09Fm},
}

@inproceedings{sauer2022stylegan,
  title={Stylegan-xl: Scaling stylegan to large diverse datasets},
  author={Sauer, Axel and Schwarz, Katja and Geiger, Andreas},
  booktitle={ACM SIGGRAPH 2022 conference proceedings},
  pages={1--10},
  year={2022}
}

@inproceedings{peebles2023scalable,
  title={Scalable diffusion models with transformers},
  author={Peebles, William and Xie, Saining},
  booktitle={Proceedings of the IEEE/CVF international conference on computer vision},
  pages={4195--4205},
  year={2023}
}

@inproceedings{ma2024sit,
  title={Sit: Exploring flow and diffusion-based generative models with scalable interpolant transformers},
  author={Ma, Nanye and Goldstein, Mark and Albergo, Michael S and Boffi, Nicholas M and Vanden-Eijnden, Eric and Xie, Saining},
  booktitle={European Conference on Computer Vision},
  pages={23--40},
  year={2024},
  organization={Springer}
}

@article{tian2024visual,
  title={Visual autoregressive modeling: Scalable image generation via next-scale prediction},
  author={Tian, Keyu and Jiang, Yi and Yuan, Zehuan and Peng, Bingyue and Wang, Liwei},
  journal={Advances in neural information processing systems},
  volume={37},
  pages={84839--84865},
  year={2024}
}

\clearpage
\appendix

\section{Limitations of FID}
\label{appendix:fid-limit}

Despite its widespread use, FID has several known limitations, including sensitivity to sample size \citep{fid-infinite}, implementation details \citep{clean-fid}, and the choice of feature encoder \citep{dgm-eval,yang2026representationfrechetlossvisual}. We focus on a more fundamental issue: FID approximates complex feature distributions with a single Gaussian. Given real and generated samples, FID embeds both sets using a feature extractor and approximates each resulting feature distribution by a single Gaussian. The score then compares only the fitted means and covariance matrices. Thus, FID does not compare the full feature distributions; it compares a two-moment summary of them. This creates a fundamental blind spot. Distinct distributions can have identical first and second moments while differing substantially in shape, support, modality, and tail behavior. This failure is not tied to a particular encoder: any metric (including $\text{FD}_{\text{DINOv2}}$ \citep{dgm-eval}) that reduces a feature distribution to a single mean and covariance matrix is oblivious to discrepancies beyond second-order structure.

To make this limitation explicit, consider two distributions in \(\mathbb{R}^d\): $P = \mathcal{N}(0,I_d) $ and $Q = \mathrm{Unif}([-\sqrt{3},\sqrt{3}]^d)$. These distributions are clearly different: \(P\) has unbounded support and Gaussian tails, while \(Q\) is supported on a compact hypercube. However, they have identical first and second moments. Under the Gaussian approximation used by FID, the moment-matched Gaussians for \(P\) and \(Q\) are identical, so their Fréchet distance under the Gaussian approximation is zero despite \(P \ne Q\). This is further supported by the results in Table~\ref{tab:pentagon}, where the two distinct sets achieve a perfect FD score.

\section{Additional Samples from BigGAN and IMM}
\label{appendix:dist-confusion-extra}

\begin{figure}[ht]
    \centering
    \newcommand{\dclabel}[1]{\raisebox{-.5\height}{\rotatebox{90}{\footnotesize #1}}}%
    \newlength{\dcxlabw}\settowidth{\dcxlabw}{\dclabel{king penguin}}%
    \newlength{\dcxgridw}\setlength{\dcxgridw}{\dimexpr(\textwidth-\dcxlabw-10pt)/2\relax}%
    \newcommand{\dcgrid}[1]{\raisebox{-.5\height}{\setkeys{Gin}{width=\dcxgridw}#1}}%
    \setlength{\tabcolsep}{0pt}%
    \begin{tabular}{@{}c@{\hspace{4pt}}c@{\hspace{6pt}}c@{}}
        & \small\textbf{BigGAN}: FID $6.67$ & \small\textbf{IMM (1 step)}: FID $8.11$ \\ \noalign{\vskip 3pt}
        \dclabel{tench}        & \dcgrid{\includegraphics{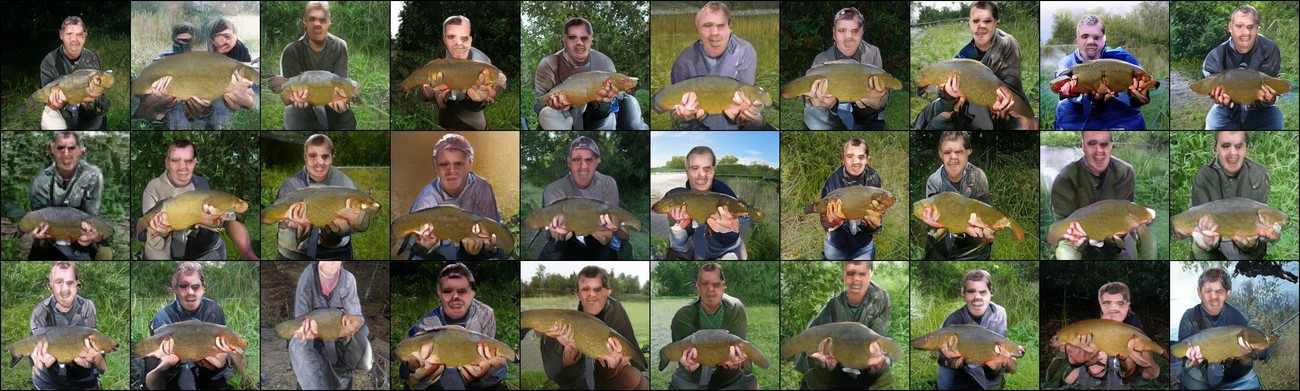}}   & \dcgrid{\includegraphics{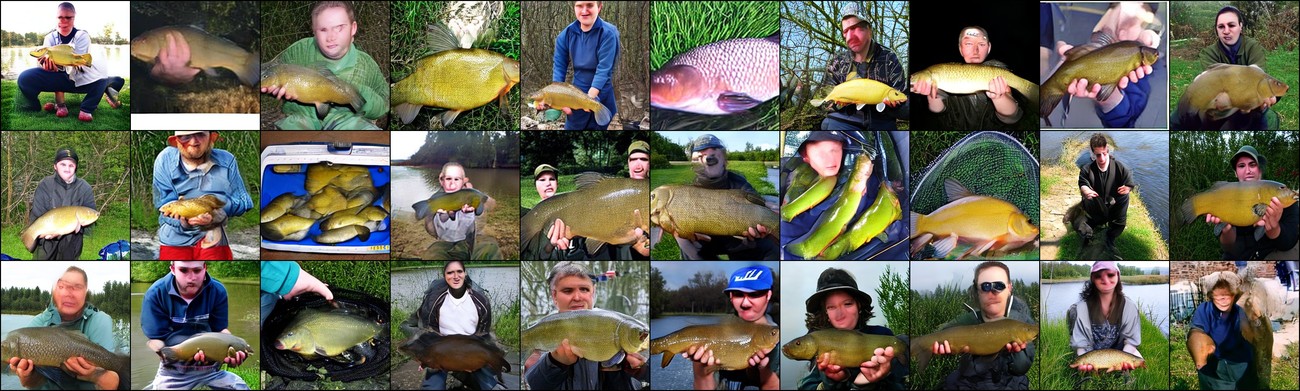}}   \\ \noalign{\vskip 3pt}
        \dclabel{macaw}        & \dcgrid{\includegraphics{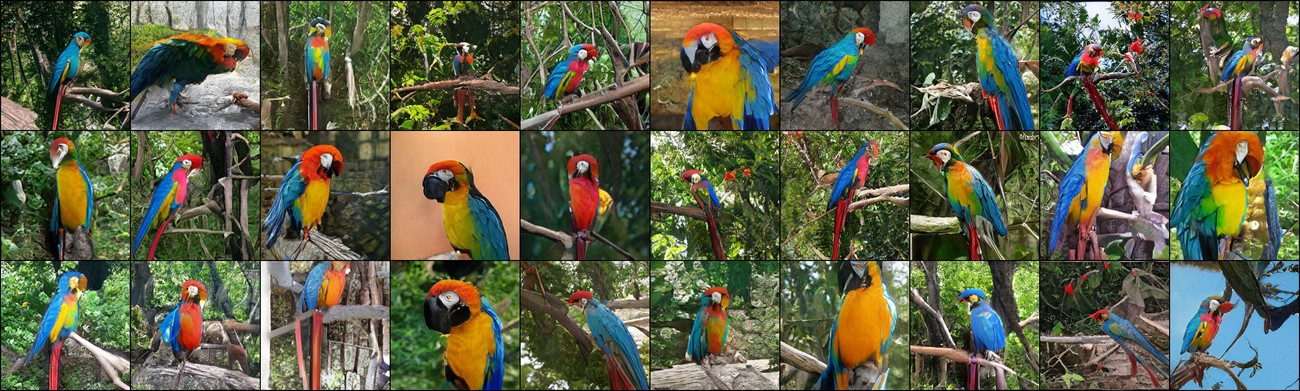}}  & \dcgrid{\includegraphics{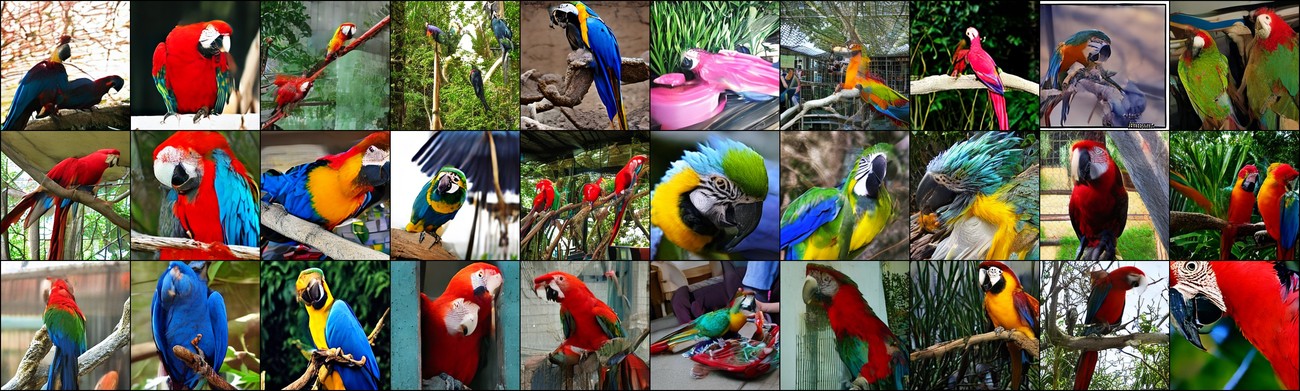}}  \\ \noalign{\vskip 3pt}
        \dclabel{king penguin} & \dcgrid{\includegraphics{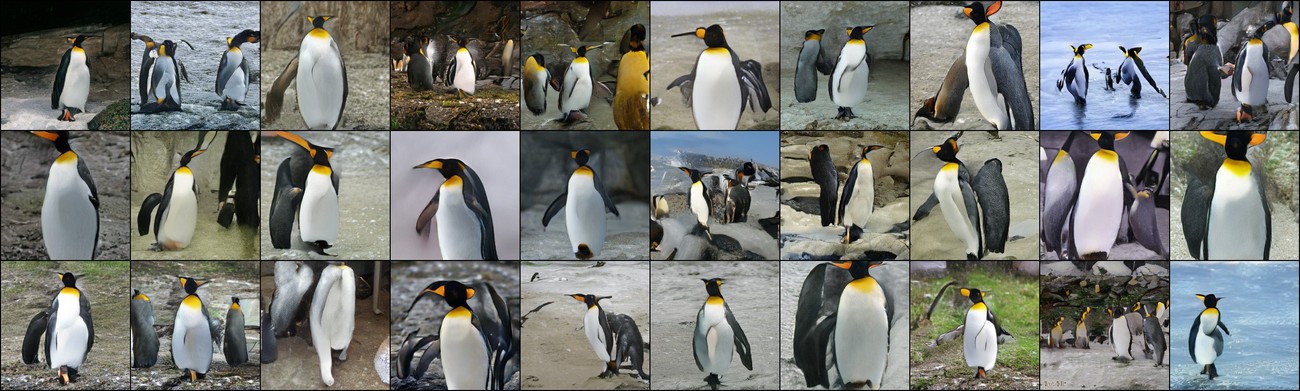}} & \dcgrid{\includegraphics{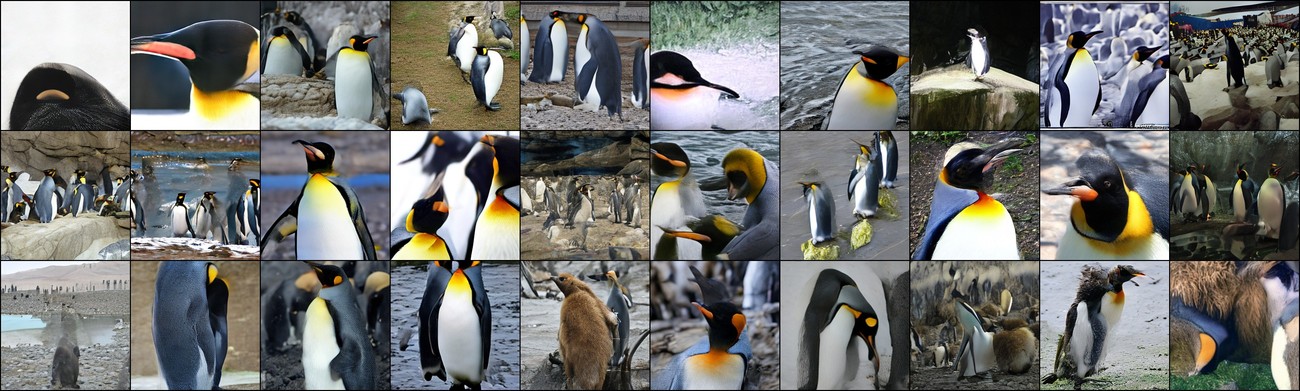}} \\ \noalign{\vskip 3pt}
        \dclabel{sports car}   & \dcgrid{\includegraphics{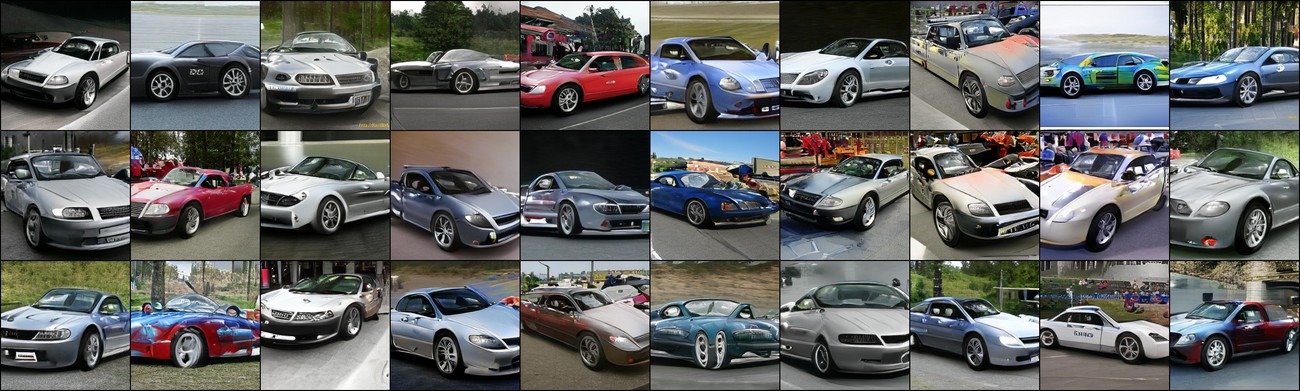}} & \dcgrid{\includegraphics{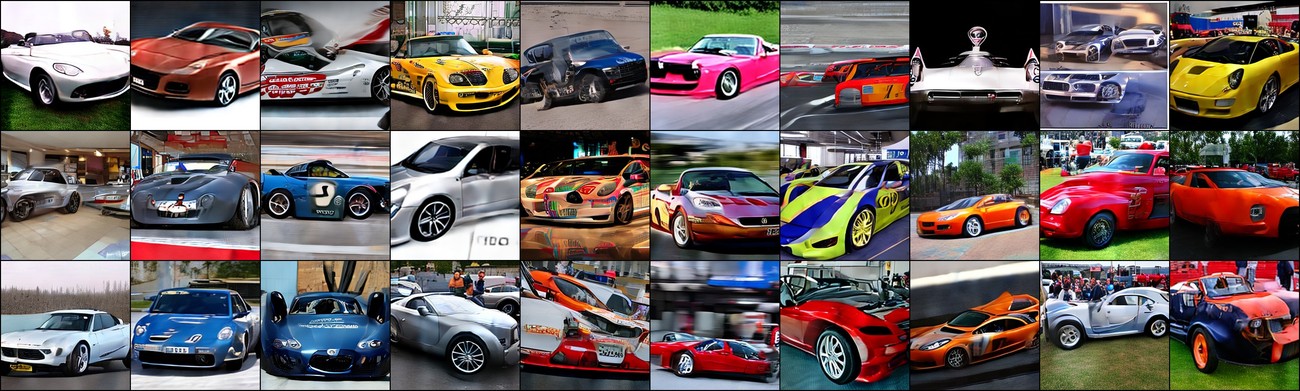}} \\ \noalign{\vskip 3pt}
        \dclabel{bell pepper}  & \dcgrid{\includegraphics{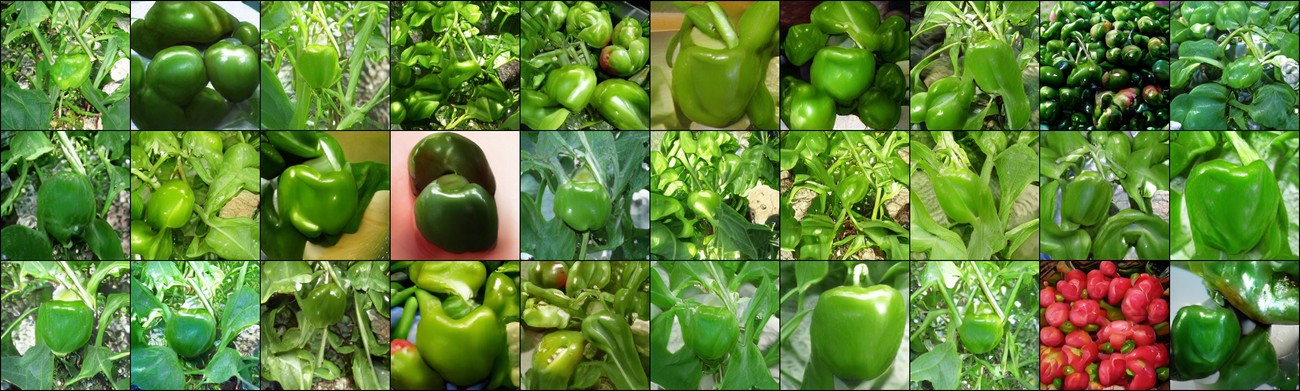}} & \dcgrid{\includegraphics{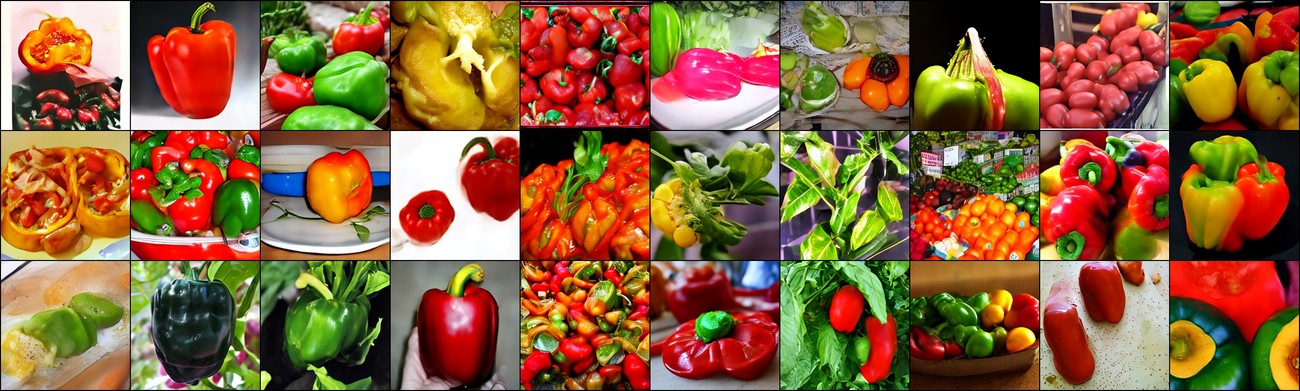}} \\
    \end{tabular}
    \caption{\textbf{Random samples for five additional ImageNet classes.} Same layout as Figure~\ref{fig:dist-confusion}.}
    \label{fig:dist-confusion-extra}
\end{figure}

Figure~\ref{fig:dist-confusion-extra} shows random samples from BigGAN and IMM (1 step) for five additional ImageNet classes, in the same layout as Figure~\ref{fig:dist-confusion}.

\section{Construction of the Pentagon Example}
\label{appendix:pentagon}

The reference set $X$ (\refmark) in Figure~\ref{fig:pentagon-sets} consists of the five vertices of a regular pentagon inscribed in the unit circle:
\[
X = \left\{ \left(\cos\theta_i,\ \sin\theta_i\right) \right\}_{i=1}^{5},
\qquad
\theta_i = \frac{\pi}{2} + \frac{2\pi i}{5}.
\]
The candidate set $Y$ (\candmark) consists of the origin and the five vertices of a second regular pentagon, rotated by $\frac{\pi}{5}$ and scaled by $\rho = \frac{\sqrt{5}}{2}$:
\[
Y = \{(0,0)\} \cup \left\{ \rho \left(\cos\left(\theta_i + \frac{\pi}{5}\right),\ \sin\left(\theta_i + \frac{\pi}{5}\right)\right) \right\}_{i=1}^{5}.
\]
Both sets have zero mean and the same sample covariance, $\frac{5}{8}I$.

\section{Evaluation on Distribution Shifts}
\label{appendix:dist-shift}

The main paper studies a controlled shift, \emph{sensitivity to sampling density}, where the candidate set is progressively enlarged while the reference set is held fixed and the behavior of each metric is measured (Section~\ref{sec:ball-limitations}). In this section, we evaluate different metrics on other controlled distribution shifts. All experiments follow the oracle setting from Table~\ref{tab:metrics}: the reference set is 10K images sampled uniformly from ImageNet, and the candidate set is 50K. For all the experiments below, we use $k=3$ and Inception-v3 as the feature encoder. This setup isolates the distributional shift, since both sets are drawn from real images and only the sampling of the candidate set is varied.

\subsection{Density Reweighting}

We partition the 1000 classes into a group of 100 and a group of 900, keeping both present in the candidate set so the support is unchanged. We progressively increase the mass of the 100-class group while correspondingly decreasing the mass of the 900-class group. This shifts the density while holding support fixed. Below, ``Head \%'' is the fraction of the candidate-set mass assigned to the 100-class group and ``Factor'' its upweighting relative to uniform ($10\%$ / $1\times$ is uniform).

\begin{table}[ht]
\centering
\caption{\textbf{Density reweighting.} The 1000 classes are split into a head group of 100 and a tail group of 900. Both remain present in the candidate set, so the support is unchanged and only the probability mass is shifted.}
\label{tab:shift-density}
\shadedrules
\begin{tabular}{cc ccc >{\columncolor{oursbg}}c >{\columncolor{oursbg}}c}
\toprule
\Tstrut\Bstrut & & & \multicolumn{2}{c}{\textbf{PR}} & \multicolumn{2}{>{\columncolor{oursbg}}c}{\textbf{Chamfer} \textit{(Ours)}} \\
\cmidrule(lr){4-5} \cmidrule(lr){6-7}
\textbf{Head \%}\Tstrut\Bstrut & \textbf{Factor} & \textbf{FID} \dn & {Prec.} \up & {Recall} \up & {Fid.} \dn & {Cov.} \dn \\
\midrule
10\%\Tstrut & $1\times$ (uniform) &  0.76 & 0.757 & 0.711 & 11.21 & 10.09 \\
20\% & $2\times$           &  1.32 & 0.760 & 0.714 & 11.20 & 10.10 \\
30\% & $3\times$           &  2.73 & 0.761 & 0.711 & 11.21 & 10.14 \\
40\% & $4\times$           &  4.89 & 0.762 & 0.712 & 11.20 & 10.20 \\
50\% & $5\times$           &  7.72 & 0.763 & 0.712 & 11.21 & 10.26 \\
60\% & $6\times$           & 11.28 & 0.764 & 0.713 & 11.21 & 10.35 \\
70\% & $7\times$           & 15.55 & 0.766 & 0.717 & 11.21 & 10.49 \\
80\% & $8\times$           & 20.62 & 0.766 & 0.723 & 11.22 & 10.69 \\
90\%\Bstrut & $9\times$           & 26.64 & 0.768 & 0.740 & 11.23 & 11.14 \\
\bottomrule
\end{tabular}
\end{table}

As the head group is upweighted, Recall fails to register the shift: it in fact moves in the wrong direction, increasing slightly ($0.711 \to 0.740$) even as the distribution becomes more concentrated. \ourmetricshort{}-Coverage, in contrast, responds monotonically ($10.09 \to 11.14$), correctly reflecting that the downweighted 900 classes become progressively less well covered. \ourmetricshort{}-Fidelity remains flat, as expected, since the reference set is unchanged.

\subsection{Sample Duplication}

We hold the sample set at 50K while reducing the number of unique images (50K $\to$ 10K), filling the shortfall with exact copies. Replication is applied within each class, so every class contributes uniformly.
\begin{table}[ht]
\centering
\caption{\textbf{Sample duplication.} The sample set is held at 50K while the number of unique images is reduced, with the shortfall filled by exact copies.}
\label{tab:shift-duplication}
\shadedrules
\begin{tabular}{cc ccc >{\columncolor{oursbg}}c >{\columncolor{oursbg}}c}
\toprule
\Tstrut\Bstrut & & & \multicolumn{2}{c}{\textbf{PR} } & \multicolumn{2}{>{\columncolor{oursbg}}c}{\textbf{Chamfer} \textit{(Ours)}} \\
\cmidrule(lr){4-5} \cmidrule(lr){6-7}
\textbf{Unique}\Tstrut\Bstrut & \textbf{50K / Unique} & \textbf{FID} \dn & {Prec.} \up & {Recall} \up & {Fid.} \dn & {Cov.} \dn \\
\midrule
50K\Tstrut & $\times 1.00$ & 0.76 & 0.759 & 0.713 & 11.21 & 10.09 \\
40K & $\times 1.25$ & 1.02 & 0.760 & 0.674 & 11.20 & 10.13 \\
30K & $\times 1.67$ & 1.29 & 0.758 & 0.594 & 11.21 & 10.17 \\
20K & $\times 2.50$ & 1.82 & 0.757 & 0.492 & 11.22 & 10.23 \\
10K\Bstrut & $\times 5.00$ & 3.47 & 0.756 & 0.000 & 11.26 & 10.53 \\
\bottomrule
\end{tabular}
\end{table}

\ourmetricshort{}-Coverage degrades smoothly and monotonically as uniqueness falls ($10.09 \to 10.53$), tracking the loss of distinct samples throughout the sweep. Recall also degrades, but fails discontinuously: at 10K unique images, every sample carries five copies, giving it a zero-radius neighborhood, so Recall reads exactly 0.

\subsection{Dropping Subsets of Classes}

We progressively restrict the sample set to fewer classes ($1000 \to 500 \to 250 \to 100$), while the reference set remains the full uniform 10K. As classes are removed from the sample set, those regions of the real distribution become uncovered.

\begin{table}[ht]
\centering
\caption{\textbf{Dropping subsets of classes.} The sample set is progressively restricted to fewer ImageNet classes while the reference set remains the full uniform 10K subset.}
\label{tab:shift-class-drop}
\shadedrules
\begin{tabular}{cc ccc >{\columncolor{oursbg}}c >{\columncolor{oursbg}}c}
\toprule
\Tstrut\Bstrut & & & \multicolumn{2}{c}{\textbf{PR}} & \multicolumn{2}{>{\columncolor{oursbg}}c}{\textbf{Chamfer} \textit{(Ours)}} \\
\cmidrule(lr){4-5} \cmidrule(lr){6-7}
\textbf{Classes}\Tstrut\Bstrut & \textbf{Imgs/class} & \textbf{FID} \dn & {Prec.} \up & {Recall} \up & {Fid.} \dn & {Cov.} \dn \\
\midrule
1000\Tstrut &  50 &  0.76 & 0.757 & 0.711 & 11.21 & 10.09 \\
 500 & 100 &  7.80 & 0.754 & 0.558 & 11.19 & 11.23 \\
 250 & 200 & 17.66 & 0.748 & 0.454 & 11.20 & 12.11 \\
 100\Bstrut & 500 & 35.46 & 0.752 & 0.362 & 11.15 & 12.99 \\
\bottomrule
\end{tabular}
\end{table}

For this experiment, all metrics respond correctly, including the ball-based ones. The result confirms our decomposition behaves as intended: \ourmetricshort{}-Coverage worsens monotonically ($10.09 \to 12.99$) as classes are removed, while \ourmetricshort{}-Fidelity stays flat ($11.21 \to 11.15$), since the remaining samples are still real and close to the reference.

\section{Ablation over Distance Function}
\label{appendix:abl-dist-func}

Equation~\ref{eq:kchamfer-one-sided} is stated with the $\ell_2$ norm, following the convention of the classical Chamfer distance. The construction is not restricted to this choice, and any distance function on the feature space yields a valid one-sided measure. In this section we ablate over $p$-norms ($\ell_1$, $\ell_2$, $\ell_3$, $\ell_4$) and cosine distance. We repeat the oracle setting of Table~\ref{tab:metrics}: both the sample set and the reference set are drawn from real training images, and we vary the sample-set size (10K / 30K / 50K) to observe the effect of sampling density. Because the absolute scale of the metric differs by orders of magnitude across these choices, we also report $\Delta$Cov., the relative decrease in \ourmetricshort{}-Coverage from 10K to 50K. This gives a scale-free summary of how strongly each variant responds to sampling density.

\begin{table}[ht]
\centering
\caption{\textbf{Ablation over the distance function used in \ourmetric{}.} Oracle setting on ImageNet-1K with Inception-v3 features and $k=1$.}
\label{tab:abl-dist-func}
\begin{tabular}{l *{6}{S[table-format=3.2]} c}
\toprule
& \multicolumn{2}{c}{\textbf{10K}} & \multicolumn{2}{c}{\textbf{30K}} & \multicolumn{2}{c}{\textbf{50K}} & \\
\cmidrule(lr){2-3} \cmidrule(lr){4-5} \cmidrule(lr){6-7}
\textbf{Distance} & {Fid. \dn} & {Cov. \dn} & {Fid. \dn} & {Cov. \dn} & {Fid. \dn} & {Cov. \dn} & \textbf{$\Delta$Cov.} \up \\
\midrule
$\ell_1$              & 315.10 & 312.02 & 314.38 & 293.09 & 314.22 & 284.05 & 8.97\% \\
$\ell_2$              & 10.646 & 10.538 & 10.621 &  9.884 & 10.618 &  9.570 & 9.18\% \\
$\ell_3$              &  4.005 &  3.961 &  3.996 &  3.709 &  3.995 &  3.587 & 9.44\% \\
$\ell_4$              &  2.642 &  2.611 &  2.636 &  2.439 &  2.635 &  2.357 & 9.74\% \\
Cosine     &  0.621 &  0.617 &  0.620 &  0.579 &  0.620 &  0.560 & 9.34\% \\
\bottomrule
\end{tabular}
\end{table}

The qualitative behavior is invariant across all choices, while only the absolute scale of the metric varies. In every case \ourmetricshort{}-Coverage decreases monotonically with sampling density, by a consistent 9--10\% from 10K to 50K, and \ourmetricshort{}-Fidelity remains stable because the reference set is fixed. We therefore recommend $\ell_2$ as the default.

\section{Ablation over Encoder}
\label{appendix:abl-encoder}

Equation~\ref{eq:kchamfer-one-sided} requires only a feature map $f$ and places no constraint on how that map is obtained. We report results using DINOv2-Giant (ViT-g/14) \citep{dinov2} in the main paper, following common practice, but the construction is deliberately encoder-agnostic. It is therefore reasonable to ask whether the behavior we report is a property of the metric or of a particular feature space.

In this section we ablate across encoders spanning feature dimensions from 384 to 2048: four DINOv2 \citep{dinov2} vision transformers and Inception-v3 \citep{inception}. We repeat the \emph{oracle setting} of Table~\ref{tab:metrics}. Because the absolute scale of the metric differs across feature spaces, we report $\Delta$Cov., the relative decrease in \ourmetricshort{}-Coverage from 10K to 50K, as a scale-free summary of how strongly each variant responds.

\begin{table}[ht]
\centering
\caption{\textbf{Ablation over the encoder used in \ourmetric{}.} Oracle setting on ImageNet-1K with $k=1$ and the $\ell_2$ norm. The reference set is fixed while the sample set grows from 10K to 50K.}
\label{tab:abl-encoder}
\begin{tabular}{l c *{6}{S[table-format=2.2]} c}
\toprule
& & \multicolumn{2}{c}{\textbf{10K}} & \multicolumn{2}{c}{\textbf{30K}} & \multicolumn{2}{c}{\textbf{50K}} & \\
\cmidrule(lr){3-4} \cmidrule(lr){5-6} \cmidrule(lr){7-8}
\textbf{Backbone} & \textbf{Dim.} & {Fid. \dn} & {Cov. \dn} & {Fid. \dn} & {Cov. \dn} & {Fid. \dn} & {Cov. \dn} & \textbf{$\Delta$Cov.} \up \\
\midrule
DINOv2 ViT-S/14  &  384 & 40.497 & 40.446 & 40.504 & 37.507 & 40.479 & 36.026 & 10.93\% \\
DINOv2 ViT-B/14  &  768 & 37.820 & 37.612 & 37.788 & 34.581 & 37.759 & 33.115 & 11.95\% \\
DINOv2 ViT-L/14  & 1024 & 33.969 & 33.589 & 33.923 & 30.596 & 33.894 & 29.213 & 13.03\% \\
DINOv2 ViT-g/14  & 1536 & 31.825 & 31.291 & 31.783 & 28.169 & 31.741 & 26.797 & 14.36\% \\
Inception-v3      & 2048 & 10.646 & 10.538 & 10.621 &  9.884 & 10.618 &  9.570 &  9.18\% \\
\bottomrule
\end{tabular}
\end{table}

The qualitative behavior is invariant across all encoders. In every case \ourmetricshort{}-Coverage decreases monotonically as the sample set grows denser, correctly reflecting improved coverage, while \ourmetricshort{}-Fidelity remains essentially constant because the reference set is fixed. The conclusions drawn from Table~\ref{tab:metrics} are valid, independent of the choice of feature extractor. Within the DINOv2 family, $\Delta$Cov.\ increases with model capacity. 

\section{Note on Earth Mover's Distance}
\label{appendix:emd}

\ourmetric{} is widely used in 3D vision, where it measures the discrepancy between two point clouds. Earth Mover's Distance, also called the Wasserstein distance, is the other standard measure from that literature, and the two are commonly reported together. Since we use one-sided \ourmetric{} to evaluate generative models, it is natural to ask how well this second classical point-cloud measure would serve the same purpose.

Earth Mover's Distance is symmetric, so $\mathrm{EMD}(P_n,Q_m) = \mathrm{EMD}(Q_m, P_n)$. It therefore returns a single scalar and cannot provide disentangled diagnostics. If the Earth Mover's Distance between the real and generated sets is high, we cannot tell whether this is caused by poor fidelity (meaning that samples look unrealistic) or by poor coverage (meaning that modes of the real distribution are missed). The measure we propose is asymmetric, since $d_{\text{Chamfer}}(P_n,Q_m) \neq d_{\text{Chamfer}}(Q_m,P_n)$ in general. This is precisely what allows us to report fidelity and coverage as two separate axes.

Earth Mover's Distance is also expensive to compute, requiring $O(N^3)$ time and $O(N^2)$ space. This makes it infeasible at the scale of our benchmark, where Table~\ref{tab:100k-combined-cos1-k1} uses 100K samples. By contrast, \ourmetric{} requires only two cross-set nearest-neighbor lookups, which is much cheaper computationally, as discussed in Appendix~\ref{appendix:comp-analysis}.

\section{Image Distortions on ImageNet}
\label{app:distortion-imagenet}

\begin{figure}[h]
    \centering
    \setlength{\tabcolsep}{2pt}
    \renewcommand{\arraystretch}{0.5}
    \begin{tabular}{@{}>{\centering\arraybackslash}p{0.045\textwidth} ccccccc@{}}
        \rotatebox{90}{\small\textbf{\;\;Original}} &
            \imnimg{\includegraphics{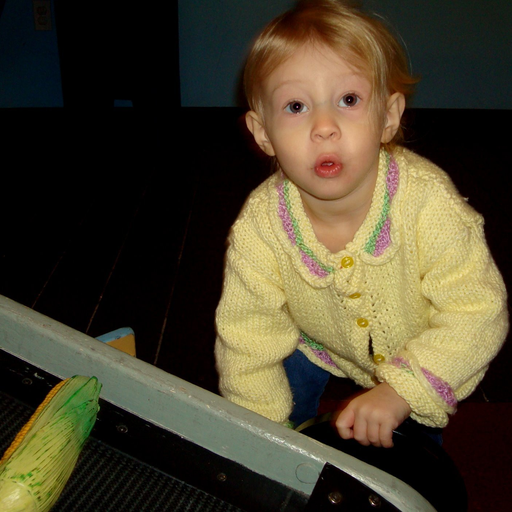}} & \imnimg{\includegraphics{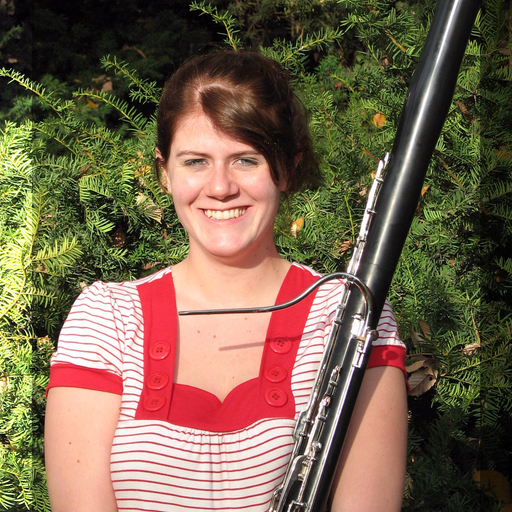}} & \imnimg{\includegraphics{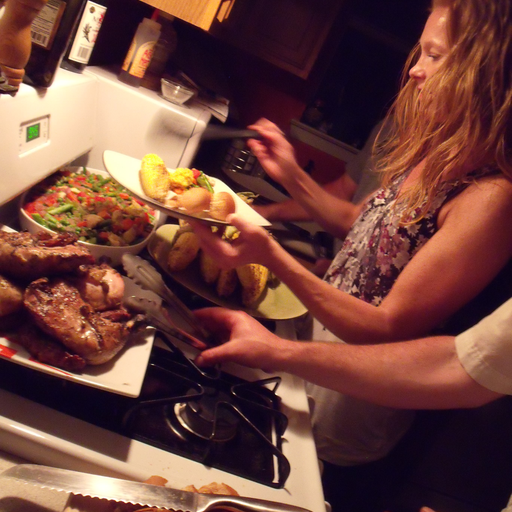}} & \imnimg{\includegraphics{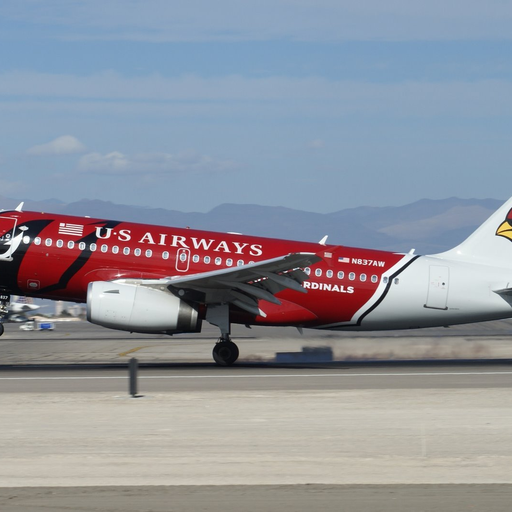}} & \imnimg{\includegraphics{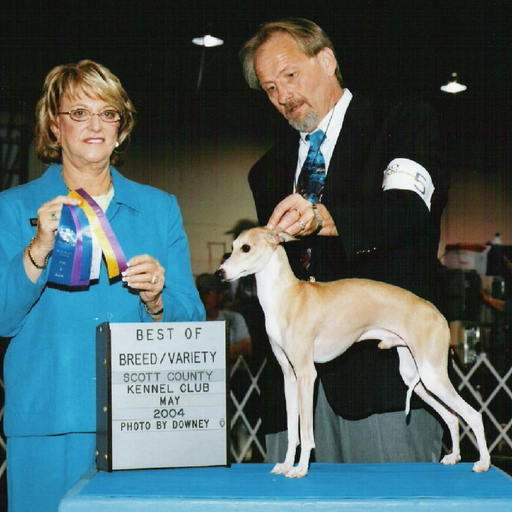}} & \imnimg{\includegraphics{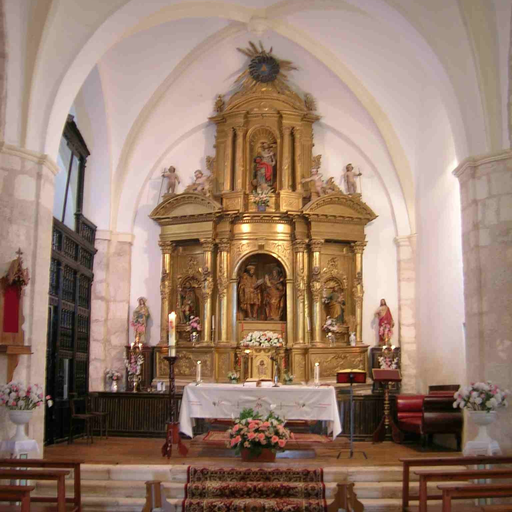}} & \imnimg{\includegraphics{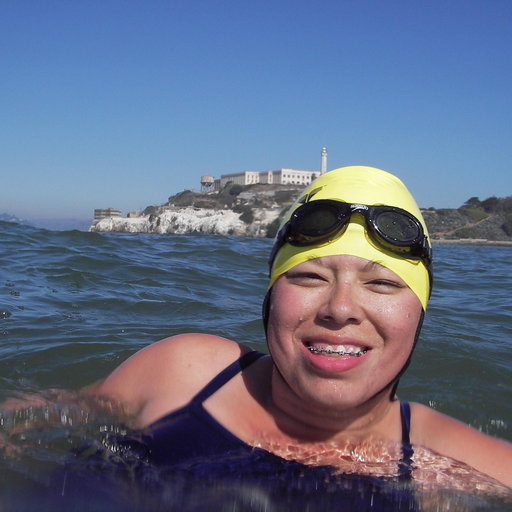}} \\[4pt]
        \rotatebox{90}{\small\textbf{\;\;Distorted}} &
            \imnimg{\includegraphics{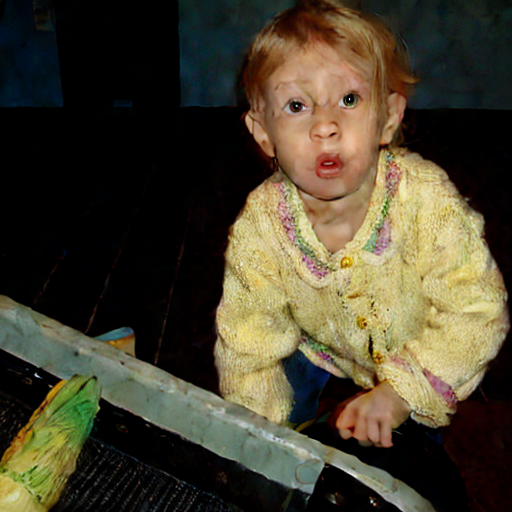}} & \imnimg{\includegraphics{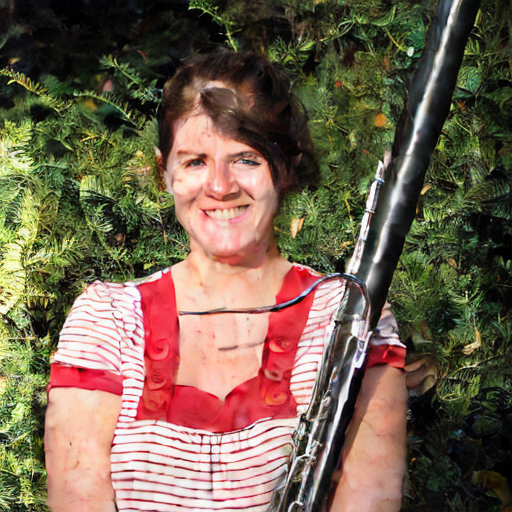}} & \imnimg{\includegraphics{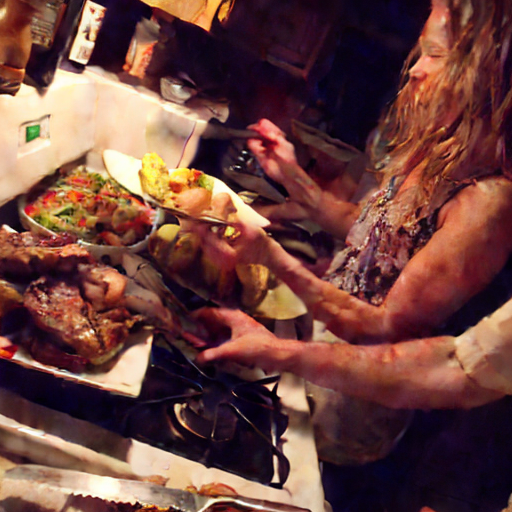}} & \imnimg{\includegraphics{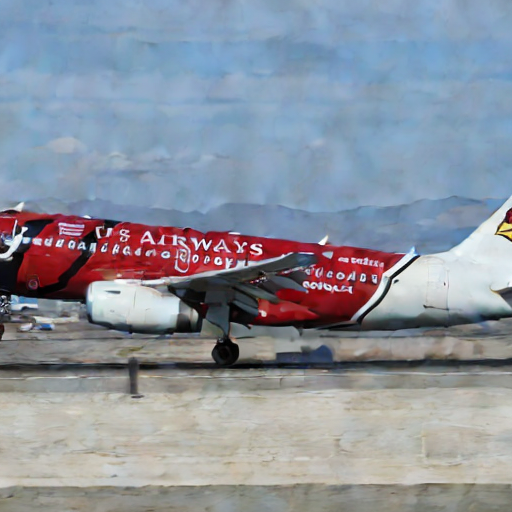}} & \imnimg{\includegraphics{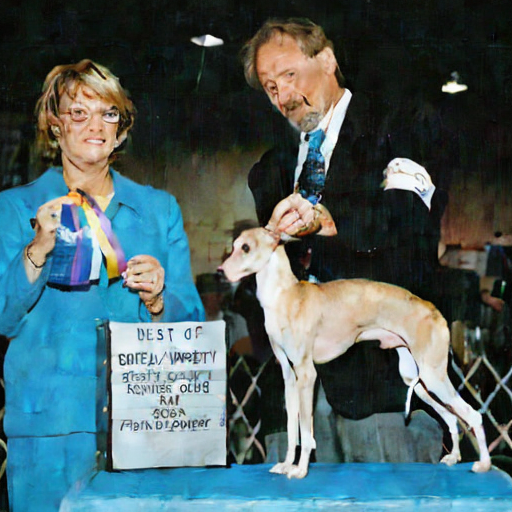}} & \imnimg{\includegraphics{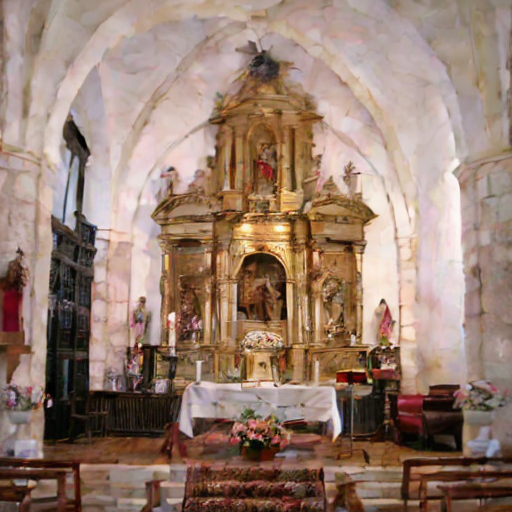}} & \imnimg{\includegraphics{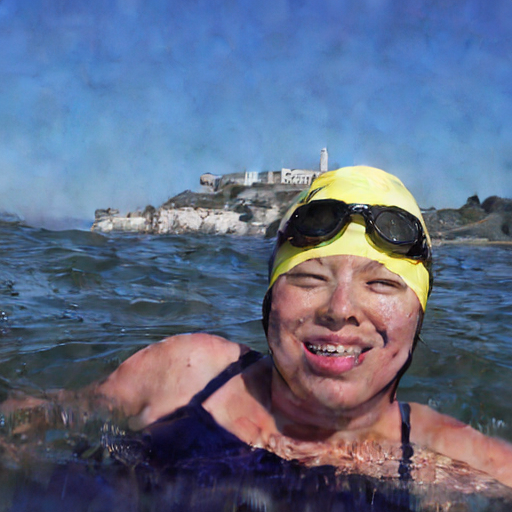}} \\
    \end{tabular}
    \caption{\textbf{Original and distorted image pairs.} Each column shows an original ImageNet image (top) and its distorted counterpart obtained by adding element-wise Gaussian noise with mean \(0\) and standard deviation \(0.225\) to the VAE latent representation before decoding.}
    \label{fig:noise-grid-imagenet}
\end{figure}

\textbf{Sensitivity to Image Distortions.}
We repeat the image distortion experiment of Section~\ref{sec:ball-limitations} on 5K randomly sampled ImageNet images, using Inception-v3 \citep{inception} features with $k=3$. Figure~\ref{fig:noise-grid-imagenet} shows representative original-distorted pairs, and Table~\ref{tab:noise-imagenet} reports the metrics. KNN-ball-based metrics again remain nearly unchanged as the noise level increases, while \ourmetricshort{}-Fidelity and \ourmetricshort{}-Coverage increase monotonically with the degradation.

\begin{table}[t]
\centering
\small
\caption{\textbf{Sensitivity to Image Distortions (ImageNet).} Metrics are computed using Inception-v3 \citep{inception} features for 5K ImageNet images and their distorted counterparts across increasing latent noise levels. Rows correspond to the standard deviation \(\sigma\) associated with the Gaussian noise added.}
\label{tab:noise-imagenet}
\shadedrules
\begin{tabular}{l S S S S S S >{\columncolor{oursbg}}S >{\columncolor{oursbg}}S}
\toprule
\Tstrut\Bstrut & \multicolumn{2}{c}{\textbf{PR}} & \multicolumn{2}{c}{\textbf{DC}} & \multicolumn{2}{c}{\textbf{Symmetric}} & \multicolumn{2}{>{\columncolor{oursbg}}c}{\textbf{Chamfer} \textit{(Ours)}} \\
\cmidrule(lr){2-3} \cmidrule(lr){4-5} \cmidrule(lr){6-7} \cmidrule(lr){8-9}
\(\sigma\)\Tstrut\Bstrut & {Prec.\up} & {Recall\up} & {Density\up} & {Cov.\up} & {Prec.\up} & {Recall\up} & {Fid.\dn} & {Cov.\dn} \\
\midrule
0.10\Tstrut   & 1.000 & 1.000 & 1.133 & 1.000 & 1.000 & 1.000 &  8.787 &  8.737 \\
0.125  & 0.999 & 1.000 & 1.111 & 1.000 & 0.999 & 1.000 &  9.052 &  8.977 \\
0.15   & 0.999 & 1.000 & 1.078 & 0.999 & 0.999 & 0.999 &  9.371 &  9.264 \\
0.175  & 0.993 & 0.999 & 1.040 & 0.997 & 0.993 & 0.997 &  9.735 &  9.585 \\
0.20   & 0.982 & 0.999 & 0.993 & 0.991 & 0.982 & 0.991 & 10.136 &  9.932 \\
0.225\Bstrut  & 0.960 & 0.996 & 0.926 & 0.979 & 0.960 & 0.979 & 10.568 & 10.300 \\
\bottomrule
\end{tabular}
\end{table}

\clearpage
\section{Additional Qualitative Analysis}
\label{app:interpret-extra}

\subsection{Per-sample \ourmetricshort{}-Fidelity for a class}

\begin{figure}[h]
    \centering
    \definecolor{imfblue}{HTML}{1F5FAF}
    \definecolor{driftorange}{HTML}{C4501F}
    \ifdefined\spreadnat\else\newlength{\spreadnat}\fi
    \settowidth{\spreadnat}{\includegraphics{figures/interpret/class_spread_1k/cheeseburger_imeanflow.pdf}}%
    \makeatletter
    \edef\spreadscale{\strip@pt\dimexpr 1pt*\dimexpr 0.47\linewidth\relax/\spreadnat\relax}%
    \makeatother
    \newcommand{\spreadpanel}[1]{\raisebox{-.5\height}{#1}}%
    \newcommand{\spreadrow}[3]{%
        \raisebox{-.5\height}{\rotatebox{90}{\small #1}} &
        \spreadpanel{#2} &
        \spreadpanel{#3} \\}%
    \newcommand{\classsep}{\arrayrulecolor{black!18}\midrule}
    \setlength{\tabcolsep}{3pt}%
    \setlength{\aboverulesep}{5pt}%
    \setlength{\belowrulesep}{5pt}%
    \setlength{\lightrulewidth}{0.4pt}%
    \begin{tabular}{@{}c r r@{}}
        & \multicolumn{1}{c}{\small\color{imfblue} iMeanFlow} & \multicolumn{1}{c@{}}{\small\color{driftorange} Drifting-L} \\[3pt]
        \spreadrow{tiger}
            {\includegraphics[scale=\spreadscale]{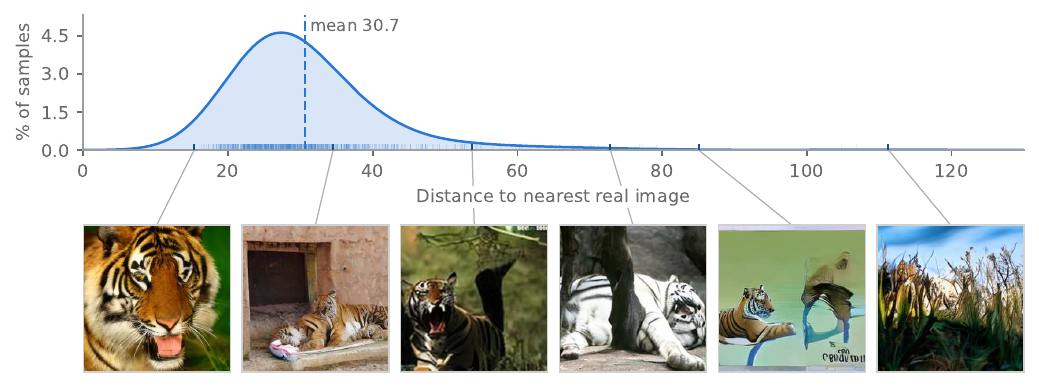}}
            {\includegraphics[scale=\spreadscale]{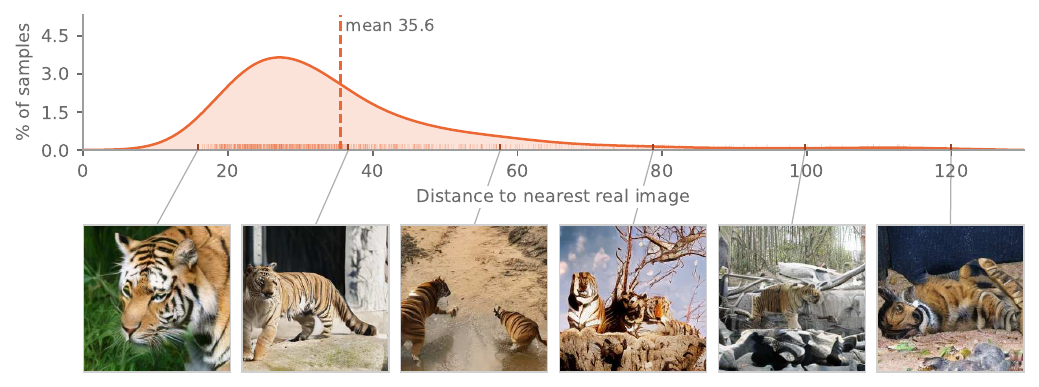}}
        \classsep
        \spreadrow{goldfish}
            {\includegraphics[scale=\spreadscale]{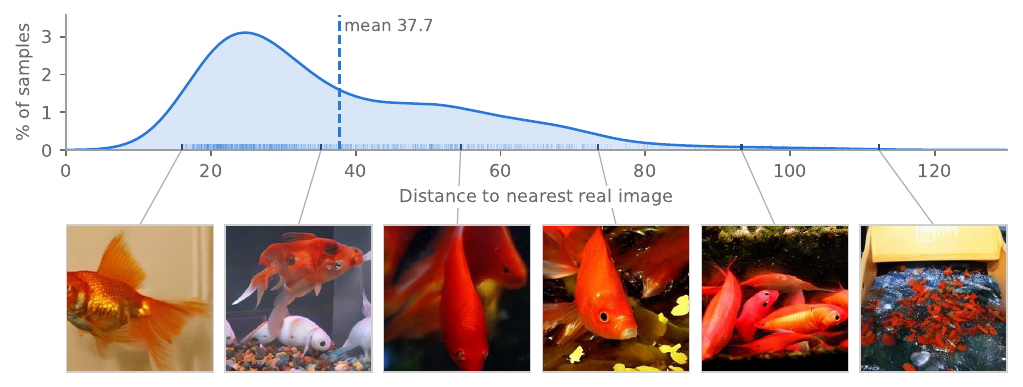}}
            {\includegraphics[scale=\spreadscale]{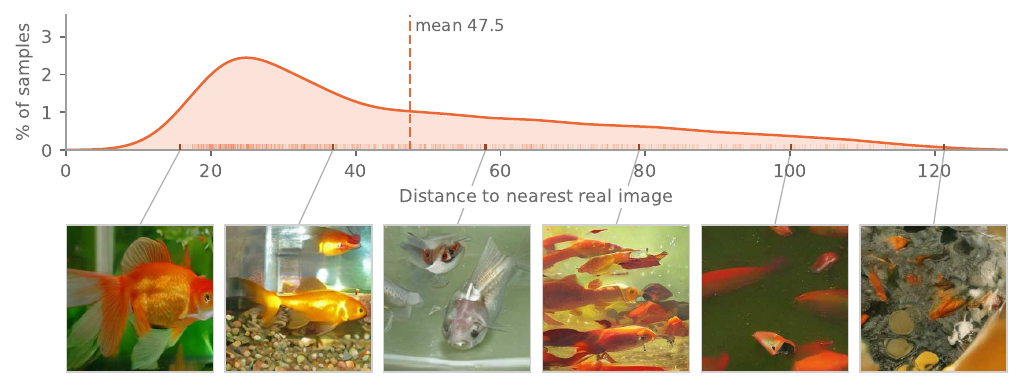}}
        \classsep
        \spreadrow{blue jay}
            {\includegraphics[scale=\spreadscale]{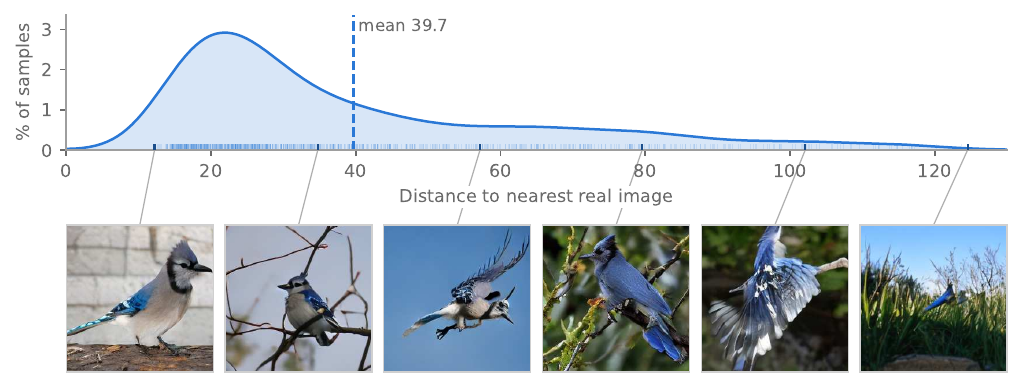}}
            {\includegraphics[scale=\spreadscale]{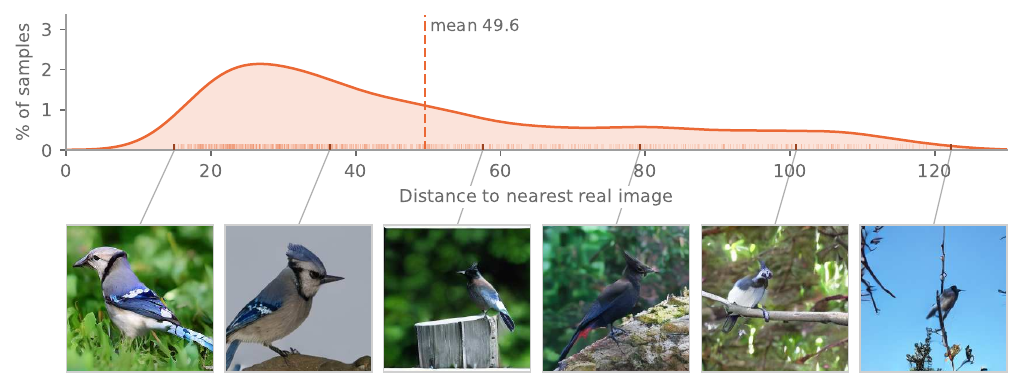}}
        \classsep
        \spreadrow{koala}
            {\includegraphics[scale=\spreadscale]{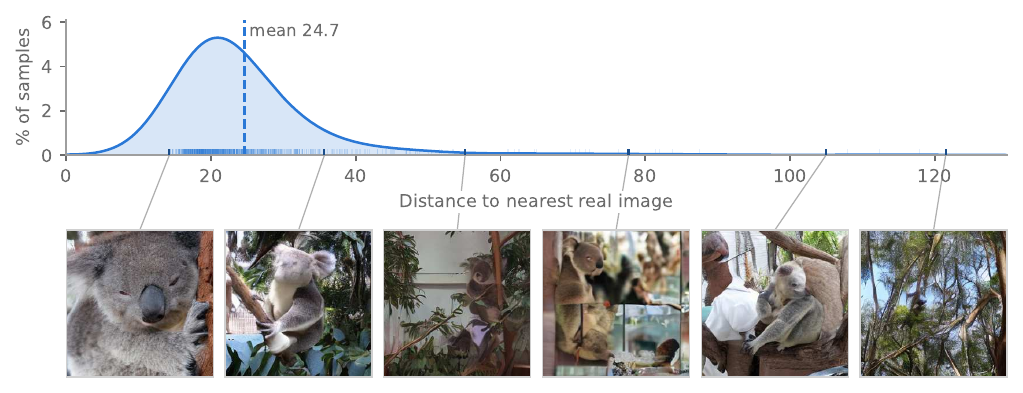}}
            {\includegraphics[scale=\spreadscale]{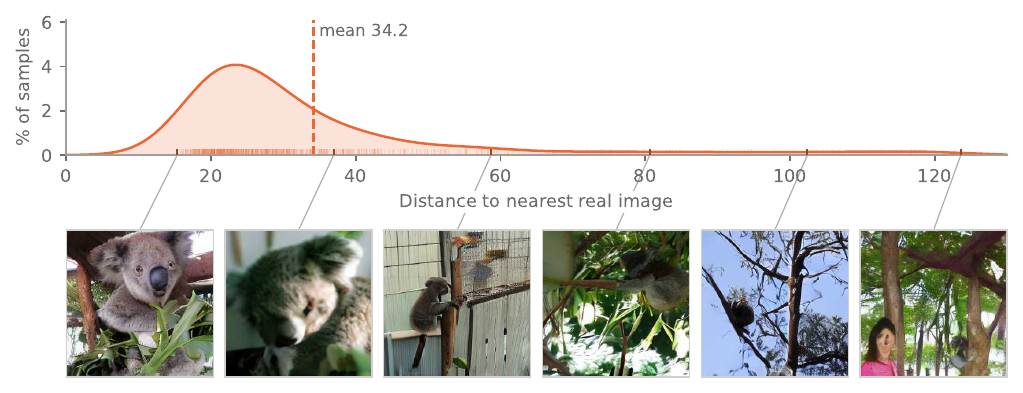}}
    \end{tabular}
    \caption{\textbf{Additional classes.} Same setup as Figure~\ref{fig:interpret-class-spread}.}
    \label{fig:interpret-class-spread-extra}
\end{figure}

\section{Benchmark Evaluation Protocol}
\label{appendix:gen-zoo}

For Table~\ref{tab:100k-combined-cos1-k1}, in order to compute Fréchet metrics (FID and $\text{FD}_{\text{DINOv2}}$), we generate 100K samples from the models and evaluate them against statistics computed from the full ImageNet training set. 
For the other metrics, namely Precision, Recall, \ourmetricshort-Fidelity and \ourmetricshort-Coverage, we evaluate on 100K generated samples against 100K reference samples, which are randomly drawn from the ImageNet training set. We repeat each evaluation with 3 random seeds, each drawing a new set of 100K generated samples, and report the mean and standard deviation.
We embed images with DINOv2 ViT-g/14 \citep{dinov2}, using the CLS token of the final layer. Real images are center-cropped to $256\times256$ following the ADM \citep{dhariwal2021diffusionmodelsbeatgans} pre-processing pipeline. For every model, we use the official checkpoint and the default sampling settings (like guidance scale) released by its authors.

\section{Discussion of Computational Complexity}
\label{appendix:comp-analysis}

For reference and generated sets of sizes $N$ and $M$ with $d$-dimensional features, \ourmetric{} reduces to two nearest-neighbor searches between the sets, one in each direction. We run these as exact searches with FAISS \citep{faiss-2} on the GPU. An exact search takes $O(NMd)$ time, but FAISS computes it as dense matrix multiplications, which GPUs execute at high throughput. FID estimates a $d \times d$ covariance matrix for each set, which takes $O((N+M)d^2)$ time, and then computes a matrix square root, which takes $O(d^3)$ time. The square root does not depend on the number of samples, which is consistent with FID's runtime in Table~\ref{tab:time} changing little between 10K and 50K generated samples (8.36\,s and 9.67\,s). In every setting of Table~\ref{tab:time}, \ourmetric{} is the fastest of the three metrics. At 100K samples per set, it takes 4.20\,s, compared with 25.95\,s for FID.

\section{Additional Benchmarking Results}
\label{appendix:additional-table}

\begin{table}[ht]
\centering
\caption{\textbf{Evaluation of modern generative models on ImageNet (unnormalized features, $k=1$).} Precision, Recall, \ourmetricshort{}-Fidelity and \ourmetricshort{}-Coverage use unnormalized DINOv2-Giant (ViT-g/14) features and $k=1$ nearest neighbor; \ourmetricshort{} values are not rescaled. Values are mean $\pm$ std over 3 runs, each drawing a new set of 100K generated samples. A std shown as $0.00$ is below 0.005.}
\label{tab:TODO-euc-k1}
\resizebox{\ifdim\width>\textwidth\textwidth\else\width\fi}{!}{%
\sisetup{round-mode=places, round-precision=2}%
\shadedrules%
\newcommand{\famrule}{\arrayrulecolor{black!20}\specialrule{0.4pt}{0pt}{0pt}\arrayrulecolor{black}}
\newcommand{\meanstd}[2]{\mbox{\tablenum{#1}}\ensuremath{{}_{\pm\num{#2}}}}%
\newcommand{\meanph}[1]{\mbox{\tablenum{#1}}\ensuremath{{}_{\pm\textcolor{red}{?}}}}
\begin{tabular}{l
  S[table-format=4.0, round-mode=none]
  S[table-format=3.0, round-mode=none]
  >{\sisetup{table-format=3.2}}c
  >{\sisetup{table-format=1.2}}c
  >{\sisetup{table-format=1.2}}c
  >{\columncolor{oursbg}\sisetup{table-format=2.2}}c
  >{\columncolor{oursbg}\sisetup{table-format=2.2}}c
  >{\sisetup{table-format=1.2}}c}
\toprule
\Tstrut\Bstrut & & & \multicolumn{5}{c}{\textbf{DINOv2-Giant (ViT-g/14)}} & \multicolumn{1}{c}{\textbf{Inception-v3}} \\
\cmidrule(lr){4-8} \cmidrule(lr){9-9}
\textbf{Model}\Tstrut\Bstrut
  & {\textbf{Params (M)}}
  & {\textbf{NFE} \dn}
  & {\textbf{$\text{FD}_{\text{DINOv2}}$} \dn}
  & {\textbf{Precision} \up}
  & {\textbf{Recall} \up}
  & {\textbf{Chamfer-Fid.} \dn}
  & {\textbf{Chamfer-Cov.} \dn}
  & {\textbf{FID} \dn} \\
\midrule
\multicolumn{3}{l}{\textit{GANs}}\Tstrut & & & & & & \\
BigGAN~\citep{brock2018large}                & 112  & 1   & \meanstd{743.29}{0.85} & \meanstd{0.2318}{0.0001} & \meanstd{0.0497}{0.0030} & \meanstd{43.297}{0.048} & \meanstd{38.558}{0.072} & \meanstd{6.665}{0.047} \\
StyleGAN-XL~\citep{sauer2022stylegan}\Bstrut 
                                             & 166  & 1   & \meanstd{317.998}{0.81} & \meanstd{0.1068}{0.0008} & \meanstd{0.1732}{0.0023} & \meanstd{41.507}{0.032} & \meanstd{34.496}{0.02} & \meanstd{2.087}{0.056} \\
\midrule
\multicolumn{3}{l}{\textit{Diffusion and flow matching}}\Tstrut & & & & & & \\
DiT-XL/2 (50 steps)~\citep{peebles2023scalable} & 675  & 50  & \meanstd{168.70}{1.03} & \meanstd{0.2299}{0.0009} & \meanstd{0.3039}{0.0023} & \meanstd{33.904}{0.028} & \meanstd{30.226}{0.020} & \meanstd{3.363}{0.057} \\
DiT-XL/2 (125 steps)                         & 675  & 125 & \meanstd{120.10}{0.38} & \meanstd{0.2600}{0.0009} & \meanstd{0.3302}{0.0024} & \meanstd{32.207}{0.037} & \meanstd{29.841}{0.007} & \meanstd{2.068}{0.045} \\
DiT-XL/2 (250 steps)\Bstrut                  & 675  & 250 & \meanstd{108.55}{0.86} & \meanstd{0.2691}{0.0003} & \meanstd{0.3425}{0.0023} & \meanstd{31.757}{0.014} & \meanstd{29.740}{0.001} & \meanstd{1.918}{0.011} \\
\famrule
SiT-XL/2~\citep{ma2024sit}\Tstrut            & 675  & 250 & \meanstd{151.16}{0.72} & \meanstd{0.2159}{0.0013} & \meanstd{0.3509}{0.0019} & \meanstd{34.450}{0.073} & \meanstd{30.233}{0.023} & \meanstd{1.742}{0.012} \\
SiT-XL/2 + REPA \cite{repa} 
                                             & 675  & 250 & \meanstd{75.59}{0.50} & \meanstd{0.3016}{0.0011} & \meanstd{0.4050}{0.0028} & \meanstd{30.654}{0.023} & \meanstd{28.837}{0.014} & \meanstd{1.026}{0.009} \\
JiT-H/16 \cite{jit-diffusion} \Bstrut 
                                             & 953  & 99  & \meanstd{79.93}{0.21} & \meanstd{0.2831}{0.0011} & \meanstd{0.3901}{0.0009} & \meanstd{31.039}{0.022} & \meanstd{29.061}{0.006} & \meanstd{1.653}{0.010} \\
\midrule
\multicolumn{3}{l}{\textit{Few-step}}\Tstrut & & & & & & \\
IMM (1 step)~\citep{zhou2025inductive}       & 675  & 1   & \meanstd{384.37}{1.77} & \meanstd{0.1233}{0.0006} & \meanstd{0.3665}{0.0018} & \meanstd{43.399}{0.039} & \meanstd{33.061}{0.032} & \meanstd{8.107}{0.079} \\
IMM (2 steps)                                & 675  & 2   & \meanstd{230.07}{0.61} & \meanstd{0.1506}{0.0012} & \meanstd{0.4386}{0.0005} & \meanstd{39.192}{0.033} & \meanstd{31.422}{0.018} & \meanstd{3.674}{0.052} \\
IMM (4 steps)                                & 675  & 4   & \meanstd{163.24}{0.86} & \meanstd{0.1795}{0.0003} & \meanstd{0.4369}{0.0027} & \meanstd{36.565}{0.061} & \meanstd{30.700}{0.025} & \meanstd{2.190}{0.014} \\
IMM (8 steps)\Bstrut                         & 675  & 8   & \meanstd{128.80}{0.31} & \meanstd{0.2050}{0.0012} & \meanstd{0.4304}{0.0036} & \meanstd{34.793}{0.014} & \meanstd{30.294}{0.030} & \meanstd{1.693}{0.010} \\
\famrule
iMF-B/2~\citep{geng2025improved}\Tstrut      & 89   & 1   & \meanstd{295.71}{0.79} & \meanstd{0.1514}{0.0008} & \meanstd{0.1548}{0.0014} & \meanstd{37.717}{0.040} & \meanstd{32.509}{0.028} & \meanstd{3.169}{0.019} \\
iMF-M/2                                      & 174  & 1   & \meanstd{207.64}{1.13} & \meanstd{0.1775}{0.0009} & \meanstd{0.2325}{0.0028} & \meanstd{35.814}{0.061} & \meanstd{31.221}{0.031} & \meanstd{2.094}{0.018} \\
iMF-L/2                                      & 409  & 1   & \meanstd{153.50}{0.18} & \meanstd{0.2039}{0.0004} & \meanstd{0.2983}{0.0007} & \meanstd{34.237}{0.026} & \meanstd{30.366}{0.017} & \meanstd{1.582}{0.019} \\
iMF-XL/2\Bstrut                              & 610  & 1   & \meanstd{138.65}{0.41} & \meanstd{0.2158}{0.0009} & \meanstd{0.3242}{0.0030} & \meanstd{33.722}{0.051} & \meanstd{30.044}{0.016} & \meanstd{1.477}{0.011} \\
\famrule
pMF-H/16~\citep{pixel-mean-flows}\Tstrut\Bstrut
                                             & 956  & 1   & \meanstd{111.19}{0.57} & \meanstd{0.2448}{0.0021} & \meanstd{0.3478}{0.0014} & \meanstd{32.866}{0.028} & \meanstd{29.688}{0.022} & \meanstd{1.969}{0.015} \\
\famrule
Drifting-B~\citep{deng2026generative}\Tstrut    & 133  & 1   & \meanstd{248.38}{0.32} & \meanstd{0.1401}{0.0012} & \meanstd{0.2965}{0.0021} & \meanstd{38.588}{0.007} & \meanstd{32.024}{0.014} & \meanstd{1.436}{0.004} \\
Drifting-L\Bstrut                               & 463  & 1   & \meanstd{213.14}{1.14} & \meanstd{0.1529}{0.0005} & \meanstd{0.3018}{0.0038} & \meanstd{37.537}{0.060} & \meanstd{31.447}{0.017} & \meanstd{1.238}{0.005} \\
\famrule
Priming-L~\citep{priming-models}\Bstrut\Tstrut & 405  & 1   & \meanstd{88.12}{0.18} & \meanstd{0.2031}{0.0012} & \meanstd{0.4094}{0.0035} & \meanstd{33.382}{0.016} & \meanstd{29.886}{0.005} & \meanstd{0.925}{0.006} \\
\midrule
\multicolumn{3}{l}{\textit{Autoregressive and masked}}\Tstrut & & & & & & \\
VAR-d20~\citep{tian2024visual}               & 600  & 10  & \meanstd{147.80}{0.22} & \meanstd{0.2217}{0.0005} & \meanstd{0.2931}{0.0018} & \meanstd{33.304}{0.036} & \meanstd{30.372}{0.016} & \meanstd{2.389}{0.002} \\
VAR-d24                                      & 1000 & 10  & \meanstd{105.44}{0.14} & \meanstd{0.2676}{0.0004} & \meanstd{0.3504}{0.0014} & \meanstd{31.703}{0.021} & \meanstd{29.458}{0.007} & \meanstd{1.848}{0.004} \\
VAR-d30\Bstrut                               & 2000 & 10  & \meanstd{104.07}{0.20} & \meanstd{0.2742}{0.0004} & \meanstd{0.3810}{0.0022} & \meanstd{31.942}{0.027} & \meanstd{29.141}{0.001} & \meanstd{1.631}{0.011} \\
\famrule
LlamaGen-L~\citep{sun2024autoregressive}\Tstrut & 343  & 256 & \meanstd{269.79}{0.63} & \meanstd{0.2150}{0.0017} & \meanstd{0.2166}{0.0008} & \meanstd{34.926}{0.057} & \meanstd{31.943}{0.011} & \meanstd{4.131}{0.008} \\
Open-MAGVIT2-XL~\citep{open-magvit2}         & 1500 & 256 & \meanstd{145.70}{0.25} & \meanstd{0.2179}{0.0018} & \meanstd{0.3461}{0.0009} & \meanstd{34.007}{0.006} & \meanstd{30.313}{0.016} & \meanstd{2.093}{0.014} \\
RAR-XL~\citep{rar-xl}                        & 955  & 256 & \meanstd{263.36}{0.11} & \meanstd{0.1923}{0.0014} & \meanstd{0.2394}{0.0008} & \meanstd{36.622}{0.005} & \meanstd{31.298}{0.008} & \meanstd{1.194}{0.002} \\
RandAR-XL~\citep{randar-xl}                  & 775  & 88  & \meanstd{243.24}{0.59} & \meanstd{0.2285}{0.0002} & \meanstd{0.2668}{0.0021} & \meanstd{34.652}{0.026} & \meanstd{31.029}{0.012} & \meanstd{1.900}{0.013} \\
TiTok-L32~\citep{tiktok-32-tokens}\Bstrut    & 177  & 8   & \meanstd{377.30}{0.67} & \meanstd{0.2303}{0.0018} & \meanstd{0.1820}{0.0019} & \meanstd{40.604}{0.022} & \meanstd{31.857}{0.026} & \meanstd{2.497}{0.029} \\
\midrule
\multicolumn{3}{l}{\textit{FD-loss variants \cite{yang2026representationfrechetlossvisual}}}\Tstrut & & & & & & \\
pMF-H + FD-loss (Inception)                  & 956  & 1   & \meanstd{68.44}{0.13} & \meanstd{0.2634}{0.0017} & \meanstd{0.4187}{0.0013} & \meanstd{31.634}{0.018} & \meanstd{29.233}{0.008} & \meanstd{0.393}{0.004} \\
pMF-H + FD-loss (SIM)                        & 956  & 1   & \meanstd{35.88}{0.05} & \meanstd{0.3528}{0.0009} & \meanstd{0.3770}{0.0009} & \meanstd{28.718}{0.052} & \meanstd{28.631}{0.002} & \meanstd{0.433}{0.001} \\
JiT-H + FD-loss (Inception)                  & 953  & 1   & \meanstd{124.22}{0.17} & \meanstd{0.2232}{0.0020} & \meanstd{0.3204}{0.0006} & \meanstd{33.787}{0.020} & \meanstd{30.032}{0.006} & \meanstd{0.394}{0.003} \\
JiT-H + FD-loss (SIM)\Bstrut                 & 953  & 1   & \meanstd{51.90}{0.19} & \meanstd{0.3293}{0.0022} & \meanstd{0.2845}{0.0003} & \meanstd{29.109}{0.031} & \meanstd{28.994}{0.007} & \meanstd{0.433}{0.001} \\
\bottomrule
\end{tabular}%
        }
\end{table}

\begin{table}[ht]
\centering
\caption{\textbf{Evaluation of modern generative models on ImageNet (unnormalized features, $k=3$).} Precision, Recall, \ourmetricshort{}-Fidelity and \ourmetricshort{}-Coverage use unnormalized DINOv2-Giant (ViT-g/14) features and $k=3$ nearest neighbors; \ourmetricshort{} values are not rescaled. Values are mean $\pm$ std over 3 runs, each drawing a new set of 100K generated samples. A std shown as $0.00$ is below 0.005.}
\label{tab:TODO-euc-k3}
\resizebox{\ifdim\width>\textwidth\textwidth\else\width\fi}{!}{%
\sisetup{round-mode=places, round-precision=2}%
\shadedrules%
\newcommand{\famrule}{\arrayrulecolor{black!20}\specialrule{0.4pt}{0pt}{0pt}\arrayrulecolor{black}}
\newcommand{\meanstd}[2]{\mbox{\tablenum{#1}}\ensuremath{{}_{\pm\num{#2}}}}%
\newcommand{\meanph}[1]{\mbox{\tablenum{#1}}\ensuremath{{}_{\pm\textcolor{red}{?}}}}
\begin{tabular}{l
  S[table-format=4.0, round-mode=none]
  S[table-format=3.0, round-mode=none]
  >{\sisetup{table-format=3.2}}c
  >{\sisetup{table-format=1.2}}c
  >{\sisetup{table-format=1.2}}c
  >{\columncolor{oursbg}\sisetup{table-format=2.2}}c
  >{\columncolor{oursbg}\sisetup{table-format=2.2}}c
  >{\sisetup{table-format=1.2}}c}
\toprule
\Tstrut\Bstrut & & & \multicolumn{5}{c}{\textbf{DINOv2-Giant (ViT-g/14)}} & \multicolumn{1}{c}{\textbf{Inception-v3}} \\
\cmidrule(lr){4-8} \cmidrule(lr){9-9}
\textbf{Model}\Tstrut\Bstrut
  & {\textbf{Params (M)}}
  & {\textbf{NFE} \dn}
  & {\textbf{$\text{FD}_{\text{DINOv2}}$} \dn}
  & {\textbf{Precision} \up}
  & {\textbf{Recall} \up}
  & {\textbf{Chamfer-Fid.} \dn}
  & {\textbf{Chamfer-Cov.} \dn}
  & {\textbf{FID} \dn} \\
\midrule
\multicolumn{3}{l}{\textit{GANs}}\Tstrut & & & & & & \\
BigGAN~\citep{brock2018large}                & 112  & 1   & \meanstd{743.29}{0.85} & \meanstd{0.4635}{0.0011} & \meanstd{0.1400}{0.0051} & \meanstd{44.418}{0.048} & \meanstd{40.005}{0.062} & \meanstd{6.665}{0.047} \\
StyleGAN-XL~\citep{sauer2022stylegan}\Bstrut 
                                             & 166  & 1   & \meanstd{317.998}{0.81} & \meanstd{0.2858}{0.0008} & \meanstd{0.4102}{0.0023} & \meanstd{42.313}{0.032} & \meanstd{35.781}{0.02} & \meanstd{2.087}{0.056} \\
\midrule
\multicolumn{3}{l}{\textit{Diffusion and flow matching}}\Tstrut & & & & & & \\
DiT-XL/2 (50 steps)~\citep{peebles2023scalable} & 675  & 50  & \meanstd{168.70}{1.03} & \meanstd{0.4893}{0.0016} & \meanstd{0.5491}{0.0017} & \meanstd{34.787}{0.024} & \meanstd{31.445}{0.019} & \meanstd{3.363}{0.057} \\
DiT-XL/2 (125 steps)                         & 675  & 125 & \meanstd{120.10}{0.38} & \meanstd{0.5298}{0.0018} & \meanstd{0.5826}{0.0014} & \meanstd{33.109}{0.037} & \meanstd{31.036}{0.004} & \meanstd{2.068}{0.045} \\
DiT-XL/2 (250 steps)\Bstrut                  & 675  & 250 & \meanstd{108.55}{0.86} & \meanstd{0.5435}{0.0021} & \meanstd{0.5927}{0.0032} & \meanstd{32.668}{0.015} & \meanstd{30.920}{0.010} & \meanstd{1.918}{0.011} \\
\famrule
SiT-XL/2~\citep{ma2024sit}\Tstrut            & 675  & 250 & \meanstd{151.16}{0.72} & \meanstd{0.4704}{0.0014} & \meanstd{0.6062}{0.0006} & \meanstd{35.352}{0.073} & \meanstd{31.473}{0.008} & \meanstd{1.742}{0.012} \\
SiT-XL/2 + REPA \cite{repa} 
                                             & 675  & 250 & \meanstd{75.59}{0.50} & \meanstd{0.5866}{0.0017} & \meanstd{0.6628}{0.0015} & \meanstd{31.640}{0.020} & \meanstd{30.071}{0.012} & \meanstd{1.026}{0.009} \\
JiT-H/16 \cite{jit-diffusion} \Bstrut 
                                             & 953  & 99  & \meanstd{79.93}{0.21} & \meanstd{0.5670}{0.0004} & \meanstd{0.6492}{0.0018} & \meanstd{31.995}{0.024} & \meanstd{30.290}{0.006} & \meanstd{1.653}{0.010} \\
\midrule
\multicolumn{3}{l}{\textit{Few-step}}\Tstrut & & & & & & \\
IMM (1 step)~\citep{zhou2025inductive}       & 675  & 1   & \meanstd{384.37}{1.77} & \meanstd{0.3106}{0.0006} & \meanstd{0.6290}{0.0009} & \meanstd{44.283}{0.039} & \meanstd{34.579}{0.027} & \meanstd{8.107}{0.079} \\
IMM (2 steps)                                & 675  & 2   & \meanstd{230.07}{0.61} & \meanstd{0.3654}{0.0014} & \meanstd{0.6883}{0.0005} & \meanstd{40.080}{0.029} & \meanstd{32.804}{0.018} & \meanstd{3.674}{0.052} \\
IMM (4 steps)                                & 675  & 4   & \meanstd{163.24}{0.86} & \meanstd{0.4158}{0.0007} & \meanstd{0.6881}{0.0010} & \meanstd{37.469}{0.061} & \meanstd{32.011}{0.026} & \meanstd{2.190}{0.014} \\
IMM (8 steps)\Bstrut                         & 675  & 8   & \meanstd{128.80}{0.31} & \meanstd{0.4563}{0.0005} & \meanstd{0.6826}{0.0026} & \meanstd{35.707}{0.010} & \meanstd{31.557}{0.027} & \meanstd{1.693}{0.010} \\
\famrule
iMF-B/2~\citep{geng2025improved}\Tstrut      & 89   & 1   & \meanstd{295.71}{0.79} & \meanstd{0.3644}{0.0013} & \meanstd{0.3610}{0.0029} & \meanstd{38.515}{0.040} & \meanstd{33.741}{0.021} & \meanstd{3.169}{0.019} \\
iMF-M/2                                      & 174  & 1   & \meanstd{207.64}{1.13} & \meanstd{0.4121}{0.0014} & \meanstd{0.4699}{0.0011} & \meanstd{36.641}{0.061} & \meanstd{32.437}{0.033} & \meanstd{2.094}{0.018} \\
iMF-L/2                                      & 409  & 1   & \meanstd{153.50}{0.18} & \meanstd{0.4554}{0.0010} & \meanstd{0.5531}{0.0008} & \meanstd{35.105}{0.027} & \meanstd{31.587}{0.017} & \meanstd{1.582}{0.019} \\
iMF-XL/2\Bstrut                              & 610  & 1   & \meanstd{138.65}{0.41} & \meanstd{0.4722}{0.0010} & \meanstd{0.5810}{0.0016} & \meanstd{34.617}{0.051} & \meanstd{31.276}{0.015} & \meanstd{1.477}{0.011} \\
\famrule
pMF-H/16~\citep{pixel-mean-flows}\Tstrut\Bstrut
                                             & 956  & 1   & \meanstd{111.19}{0.57} & \meanstd{0.5082}{0.0008} & \meanstd{0.6073}{0.0016} & \meanstd{33.767}{0.029} & \meanstd{30.920}{0.021} & \meanstd{1.969}{0.015} \\
\famrule
Drifting-B~\citep{deng2026generative}\Tstrut    & 133  & 1   & \meanstd{248.38}{0.32} & \meanstd{0.3496}{0.0016} & \meanstd{0.5609}{0.0013} & \meanstd{39.436}{0.010} & \meanstd{33.308}{0.005} & \meanstd{1.436}{0.004} \\
Drifting-L\Bstrut                               & 463  & 1   & \meanstd{213.14}{1.14} & \meanstd{0.3728}{0.0005} & \meanstd{0.5715}{0.0025} & \meanstd{38.398}{0.061} & \meanstd{32.696}{0.009} & \meanstd{1.238}{0.005} \\
\famrule
Priming-L~\citep{priming-models}\Bstrut\Tstrut & 405  & 1   & \meanstd{88.12}{0.18} & \meanstd{0.4558}{0.0015} & \meanstd{0.6770}{0.0038} & \meanstd{34.279}{0.017} & \meanstd{31.094}{0.003} & \meanstd{0.925}{0.006} \\
\midrule
\multicolumn{3}{l}{\textit{Autoregressive and masked}}\Tstrut & & & & & & \\
VAR-d20~\citep{tian2024visual}               & 600  & 10  & \meanstd{147.80}{0.22} & \meanstd{0.4816}{0.0010} & \meanstd{0.5386}{0.0011} & \meanstd{34.173}{0.037} & \meanstd{31.569}{0.009} & \meanstd{2.389}{0.002} \\
VAR-d24                                      & 1000 & 10  & \meanstd{105.44}{0.14} & \meanstd{0.5443}{0.0018} & \meanstd{0.6048}{0.0013} & \meanstd{32.631}{0.015} & \meanstd{30.686}{0.006} & \meanstd{1.848}{0.004} \\
VAR-d30\Bstrut                               & 2000 & 10  & \meanstd{104.07}{0.20} & \meanstd{0.5479}{0.0009} & \meanstd{0.6399}{0.0001} & \meanstd{32.932}{0.027} & \meanstd{30.437}{0.002} & \meanstd{1.631}{0.011} \\
\famrule
LlamaGen-L~\citep{sun2024autoregressive}\Tstrut & 343  & 256 & \meanstd{269.79}{0.63} & \meanstd{0.4628}{0.0019} & \meanstd{0.4298}{0.0028} & \meanstd{35.910}{0.058} & \meanstd{33.174}{0.006} & \meanstd{4.131}{0.008} \\
Open-MAGVIT2-XL~\citep{open-magvit2}         & 1500 & 256 & \meanstd{145.70}{0.25} & \meanstd{0.4724}{0.0006} & \meanstd{0.5953}{0.0014} & \meanstd{34.911}{0.005} & \meanstd{31.561}{0.009} & \meanstd{2.093}{0.014} \\
RAR-XL~\citep{rar-xl}                        & 955  & 256 & \meanstd{263.36}{0.11} & \meanstd{0.4338}{0.0009} & \meanstd{0.4985}{0.0011} & \meanstd{37.577}{0.005} & \meanstd{32.586}{0.009} & \meanstd{1.194}{0.002} \\
RandAR-XL~\citep{randar-xl}                  & 775  & 88  & \meanstd{243.24}{0.59} & \meanstd{0.4842}{0.0018} & \meanstd{0.4992}{0.0032} & \meanstd{35.634}{0.025} & \meanstd{32.283}{0.013} & \meanstd{1.900}{0.013} \\
TiTok-L32~\citep{tiktok-32-tokens}\Bstrut    & 177  & 8   & \meanstd{377.30}{0.67} & \meanstd{0.4782}{0.0024} & \meanstd{0.4245}{0.0022} & \meanstd{41.614}{0.018} & \meanstd{33.212}{0.031} & \meanstd{2.497}{0.029} \\
\midrule
\multicolumn{3}{l}{\textit{FD-loss variants \cite{yang2026representationfrechetlossvisual}}}\Tstrut & & & & & & \\
pMF-H + FD-loss (Inception)                  & 956  & 1   & \meanstd{68.44}{0.13} & \meanstd{0.5395}{0.0021} & \meanstd{0.6798}{0.0021} & \meanstd{32.571}{0.016} & \meanstd{30.440}{0.001} & \meanstd{0.393}{0.004} \\
pMF-H + FD-loss (SIM)                        & 956  & 1   & \meanstd{35.88}{0.05} & \meanstd{0.6435}{0.0023} & \meanstd{0.6473}{0.0017} & \meanstd{29.699}{0.051} & \meanstd{29.782}{0.006} & \meanstd{0.433}{0.001} \\
JiT-H + FD-loss (Inception)                  & 953  & 1   & \meanstd{124.22}{0.17} & \meanstd{0.4871}{0.0013} & \meanstd{0.5963}{0.0009} & \meanstd{34.715}{0.018} & \meanstd{31.261}{0.007} & \meanstd{0.394}{0.003} \\
JiT-H + FD-loss (SIM)\Bstrut                 & 953  & 1   & \meanstd{51.90}{0.19} & \meanstd{0.6221}{0.0030} & \meanstd{0.5612}{0.0014} & \meanstd{30.094}{0.031} & \meanstd{30.108}{0.006} & \meanstd{0.433}{0.001} \\
\bottomrule
\end{tabular}%
        }
\end{table}

\begin{table}[ht]
\centering
\caption{\textbf{Evaluation of modern generative models on ImageNet (normalized features, $k=3$).} Precision, Recall, \ourmetricshort{}-Fidelity and \ourmetricshort{}-Coverage use $\ell_2$-normalized DINOv2-Giant (ViT-g/14) features and $k=3$ nearest neighbors; \ourmetricshort{} values are multiplied by 100. Table~\ref{tab:100k-combined-cos1-k1} shows the same setting with $k=1$. Values are mean $\pm$ std over 3 runs, each drawing a new set of 100K generated samples. A std shown as $0.00$ is below 0.005.}
\label{tab:TODO-cos-k3}
\resizebox{\ifdim\width>\textwidth\textwidth\else\width\fi}{!}{%
\sisetup{round-mode=places, round-precision=2}%
\shadedrules%
\newcommand{\famrule}{\arrayrulecolor{black!20}\specialrule{0.4pt}{0pt}{0pt}\arrayrulecolor{black}}
\newcommand{\meanstd}[2]{\mbox{\tablenum{#1}}\ensuremath{{}_{\pm\num{#2}}}}%
\newcommand{\meanph}[1]{\mbox{\tablenum{#1}}\ensuremath{{}_{\pm\textcolor{red}{?}}}}
\begin{tabular}{l
  S[table-format=4.0, round-mode=none]
  S[table-format=3.0, round-mode=none]
  >{\sisetup{table-format=3.2}}c
  >{\sisetup{table-format=1.2}}c
  >{\sisetup{table-format=1.2}}c
  >{\columncolor{oursbg}\sisetup{table-format=2.2}}c
  >{\columncolor{oursbg}\sisetup{table-format=2.2}}c
  >{\sisetup{table-format=1.2}}c}
\toprule
\Tstrut\Bstrut & & & \multicolumn{5}{c}{\textbf{DINOv2-Giant (ViT-g/14)}} & \multicolumn{1}{c}{\textbf{Inception-v3}} \\
\cmidrule(lr){4-8} \cmidrule(lr){9-9}
\textbf{Model}\Tstrut\Bstrut
  & {\textbf{Params (M)}}
  & {\textbf{NFE} \dn}
  & {\textbf{$\text{FD}_{\text{DINOv2}}$} \dn}
  & {\textbf{Precision} \up}
  & {\textbf{Recall} \up}
  & {\textbf{Chamfer-Fid.} \dn}
  & {\textbf{Chamfer-Cov.} \dn}
  & {\textbf{FID} \dn} \\
\midrule
\multicolumn{3}{l}{\textit{GANs}}\Tstrut & & & & & & \\
BigGAN~\citep{brock2018large}                & 112  & 1   & \meanstd{743.29}{0.85} & \meanstd{0.2909}{0.0005} & \meanstd{0.1653}{0.0048} & \meanstd{83.830}{0.091} & \meanstd{73.411}{0.118} & \meanstd{6.665}{0.047} \\
StyleGAN-XL~\citep{sauer2022stylegan}\Bstrut 
                                             & 166  & 1   & \meanstd{317.998}{0.81} & \meanstd{0.2394}{0.0008} & \meanstd{0.4252}{0.0023} & \meanstd{77.624}{0.095} & \meanstd{64.910}{0.120} & \meanstd{2.087}{0.056} \\
\midrule
\multicolumn{3}{l}{\textit{Diffusion and flow matching}}\Tstrut & & & & & & \\
DiT-XL/2 (50 steps)~\citep{peebles2023scalable} & 675  & 50  & \meanstd{168.70}{1.03} & \meanstd{0.4521}{0.0017} & \meanstd{0.5629}{0.0016} & \meanstd{63.795}{0.049} & \meanstd{57.038}{0.036} & \meanstd{3.363}{0.057} \\
DiT-XL/2 (125 steps)                         & 675  & 125 & \meanstd{120.10}{0.38} & \meanstd{0.5038}{0.0015} & \meanstd{0.5965}{0.0012} & \meanstd{60.502}{0.066} & \meanstd{56.266}{0.006} & \meanstd{2.068}{0.045} \\
DiT-XL/2 (250 steps)\Bstrut                  & 675  & 250 & \meanstd{108.55}{0.86} & \meanstd{0.5195}{0.0015} & \meanstd{0.6056}{0.0028} & \meanstd{59.639}{0.024} & \meanstd{56.046}{0.017} & \meanstd{1.918}{0.011} \\
\famrule
SiT-XL/2~\citep{ma2024sit}\Tstrut            & 675  & 250 & \meanstd{151.16}{0.72} & \meanstd{0.4389}{0.0012} & \meanstd{0.6195}{0.0010} & \meanstd{64.728}{0.138} & \meanstd{57.044}{0.017} & \meanstd{1.742}{0.012} \\
SiT-XL/2 + REPA \cite{repa} 
                                             & 675  & 250 & \meanstd{75.59}{0.50} & \meanstd{0.5702}{0.0021} & \meanstd{0.6726}{0.0017} & \meanstd{57.622}{0.032} & \meanstd{54.481}{0.022} & \meanstd{1.026}{0.009} \\
JiT-H/16 \cite{jit-diffusion} \Bstrut 
                                             & 953  & 99  & \meanstd{79.93}{0.21} & \meanstd{0.5502}{0.0007} & \meanstd{0.6602}{0.0009} & \meanstd{58.302}{0.048} & \meanstd{54.877}{0.014} & \meanstd{1.653}{0.010} \\
\midrule
\multicolumn{3}{l}{\textit{Few-step}}\Tstrut & & & & & & \\
IMM (1 step)~\citep{zhou2025inductive}       & 675  & 1   & \meanstd{384.37}{1.77} & \meanstd{0.2398}{0.0018} & \meanstd{0.6529}{0.0013} & \meanstd{82.098}{0.088} & \meanstd{62.774}{0.048} & \meanstd{8.107}{0.079} \\
IMM (2 steps)                                & 675  & 2   & \meanstd{230.07}{0.61} & \meanstd{0.3223}{0.0014} & \meanstd{0.7067}{0.0016} & \meanstd{73.701}{0.057} & \meanstd{59.465}{0.038} & \meanstd{3.674}{0.052} \\
IMM (4 steps)                                & 675  & 4   & \meanstd{163.24}{0.86} & \meanstd{0.3842}{0.0009} & \meanstd{0.7042}{0.0007} & \meanstd{68.645}{0.120} & \meanstd{58.008}{0.045} & \meanstd{2.190}{0.014} \\
IMM (8 steps)\Bstrut                         & 675  & 8   & \meanstd{128.80}{0.31} & \meanstd{0.4325}{0.0004} & \meanstd{0.6966}{0.0027} & \meanstd{65.267}{0.022} & \meanstd{57.174}{0.047} & \meanstd{1.693}{0.010} \\
\famrule
iMF-B/2~\citep{geng2025improved}\Tstrut      & 89   & 1   & \meanstd{295.71}{0.79} & \meanstd{0.3071}{0.0016} & \meanstd{0.3765}{0.0022} & \meanstd{71.064}{0.075} & \meanstd{61.295}{0.038} & \meanstd{3.169}{0.019} \\
iMF-M/2                                      & 174  & 1   & \meanstd{207.64}{1.13} & \meanstd{0.3690}{0.0013} & \meanstd{0.4859}{0.0015} & \meanstd{67.328}{0.114} & \meanstd{58.861}{0.058} & \meanstd{2.094}{0.018} \\
iMF-L/2                                      & 409  & 1   & \meanstd{153.50}{0.18} & \meanstd{0.4230}{0.0012} & \meanstd{0.5669}{0.0006} & \meanstd{64.256}{0.045} & \meanstd{57.280}{0.029} & \meanstd{1.582}{0.019} \\
iMF-XL/2\Bstrut                              & 610  & 1   & \meanstd{138.65}{0.41} & \meanstd{0.4441}{0.0002} & \meanstd{0.5938}{0.0019} & \meanstd{63.286}{0.097} & \meanstd{56.700}{0.027} & \meanstd{1.477}{0.011} \\
\famrule
pMF-H/16~\citep{pixel-mean-flows}\Tstrut\Bstrut
                                             & 956  & 1   & \meanstd{111.19}{0.57} & \meanstd{0.4956}{0.0007} & \meanstd{0.6134}{0.0017} & \meanstd{61.515}{0.055} & \meanstd{55.995}{0.041} & \meanstd{1.969}{0.015} \\
\famrule
Drifting-B~\citep{deng2026generative}\Tstrut    & 133  & 1   & \meanstd{248.38}{0.32} & \meanstd{0.2973}{0.0004} & \meanstd{0.5800}{0.0007} & \meanstd{72.505}{0.024} & \meanstd{60.456}{0.010} & \meanstd{1.436}{0.004} \\
Drifting-L\Bstrut                               & 463  & 1   & \meanstd{213.14}{1.14} & \meanstd{0.3277}{0.0004} & \meanstd{0.5894}{0.0020} & \meanstd{70.511}{0.117} & \meanstd{59.320}{0.017} & \meanstd{1.238}{0.005} \\
\famrule
Priming-L~\citep{priming-models}\Bstrut\Tstrut & 405  & 1   & \meanstd{88.12}{0.18} & \meanstd{0.4503}{0.0016} & \meanstd{0.6821}{0.0040} & \meanstd{62.277}{0.034} & \meanstd{56.275}{0.008} & \meanstd{0.925}{0.006} \\
\midrule
\multicolumn{3}{l}{\textit{Autoregressive and masked}}\Tstrut & & & & & & \\
VAR-d20~\citep{tian2024visual}               & 600  & 10  & \meanstd{147.80}{0.22} & \meanstd{0.4591}{0.0006} & \meanstd{0.5486}{0.0016} & \meanstd{62.452}{0.069} & \meanstd{57.213}{0.018} & \meanstd{2.389}{0.002} \\
VAR-d24                                      & 1000 & 10  & \meanstd{105.44}{0.14} & \meanstd{0.5275}{0.0017} & \meanstd{0.6132}{0.0023} & \meanstd{59.500}{0.029} & \meanstd{55.588}{0.010} & \meanstd{1.848}{0.004} \\
VAR-d30\Bstrut                               & 2000 & 10  & \meanstd{104.07}{0.20} & \meanstd{0.5306}{0.0018} & \meanstd{0.6475}{0.0001} & \meanstd{60.075}{0.053} & \meanstd{55.118}{0.004} & \meanstd{1.631}{0.011} \\
\famrule
LlamaGen-L~\citep{sun2024autoregressive}\Tstrut & 343  & 256 & \meanstd{269.79}{0.63} & \meanstd{0.4053}{0.0017} & \meanstd{0.4508}{0.0031} & \meanstd{65.988}{0.110} & \meanstd{60.315}{0.006} & \meanstd{4.131}{0.008} \\
Open-MAGVIT2-XL~\citep{open-magvit2}         & 1500 & 256 & \meanstd{145.70}{0.25} & \meanstd{0.4488}{0.0008} & \meanstd{0.6060}{0.0009} & \meanstd{63.751}{0.010} & \meanstd{57.199}{0.017} & \meanstd{2.093}{0.014} \\
RAR-XL~\citep{rar-xl}                        & 955  & 256 & \meanstd{263.36}{0.11} & \meanstd{0.3609}{0.0004} & \meanstd{0.5225}{0.0021} & \meanstd{69.316}{0.003} & \meanstd{59.267}{0.019} & \meanstd{1.194}{0.002} \\
RandAR-XL~\citep{randar-xl}                  & 775  & 88  & \meanstd{243.24}{0.59} & \meanstd{0.4184}{0.0020} & \meanstd{0.5239}{0.0030} & \meanstd{65.572}{0.052} & \meanstd{58.715}{0.025} & \meanstd{1.900}{0.013} \\
TiTok-L32~\citep{tiktok-32-tokens}\Bstrut    & 177  & 8   & \meanstd{377.30}{0.67} & \meanstd{0.3589}{0.0021} & \meanstd{0.4493}{0.0028} & \meanstd{77.668}{0.034} & \meanstd{60.428}{0.059} & \meanstd{2.497}{0.029} \\
\midrule
\multicolumn{3}{l}{\textit{FD-loss variants \cite{yang2026representationfrechetlossvisual}}}\Tstrut & & & & & & \\
pMF-H + FD-loss (Inception)                  & 956  & 1   & \meanstd{68.44}{0.13} & \meanstd{0.5292}{0.0020} & \meanstd{0.6869}{0.0019} & \meanstd{59.209}{0.031} & \meanstd{55.125}{0.004} & \meanstd{0.393}{0.004} \\
pMF-H + FD-loss (SIM)                        & 956  & 1   & \meanstd{35.88}{0.05} & \meanstd{0.6421}{0.0018} & \meanstd{0.6509}{0.0026} & \meanstd{53.785}{0.097} & \meanstd{53.906}{0.009} & \meanstd{0.433}{0.001} \\
JiT-H + FD-loss (Inception)                  & 953  & 1   & \meanstd{124.22}{0.17} & \meanstd{0.4607}{0.0018} & \meanstd{0.6100}{0.0015} & \meanstd{63.389}{0.035} & \meanstd{56.685}{0.015} & \meanstd{0.394}{0.003} \\
JiT-H + FD-loss (SIM)\Bstrut                 & 953  & 1   & \meanstd{51.90}{0.19} & \meanstd{0.6184}{0.0029} & \meanstd{0.5665}{0.0022} & \meanstd{54.530}{0.061} & \meanstd{54.519}{0.012} & \meanstd{0.433}{0.001} \\
\bottomrule
\end{tabular}%
        }
\end{table}

\clearpage

\end{document}